\documentclass[preprints,article,submit,oneauthor,pdftex]{Definitions/mdpi}
\pdfoutput=1
\firstpage{1} 
\pubvolume{1}
\issuenum{1}
\articlenumber{0}
\pubyear{2026}
\copyrightyear{2026}
\datereceived{ } 
\daterevised{ } 
\dateaccepted{ } 
\datepublished{ } 

\usepackage{amsmath,amssymb,amsfonts,amsthm,bm,bbm,xurl}
\usepackage{nicefrac,microtype}
\usepackage{float}
\graphicspath{{./Figures/}}

\usepackage{algorithm,algorithmic}

\def\bi{\begin{itemize}}
\def\ei{\end{itemize}}
\def\bn{\begin{enumerate}[label={(\arabic*)}]}
\def\en{\end{enumerate}}
\def\bd{\begin{description}}
\def\ed{\end{description}}
\newcommand{\EE}{\mathbb{E}}

\newcommand{\RR}{\mathbb{R}}
\newcommand{\NN}{\mathbb{N}}
\newcommand{\rmd}{\text{d}}
\newcommand{\rme}{\text{e}}
\def\wh#1{\widehat{#1}}
\def\wt#1{\widetilde{#1}}
\def\mkakko#1{\left(#1\right)}
\def\ckakko#1{\left\{#1\right\}}

\newcommand{\xx}{\mathbf{x}}
\newcommand{\yy}{\mathbf{y}}
\newcommand{\zz}{\mathbf{z}}
\newcommand{\nn}{\mathbf{n}}
\newcommand{\SSS}{\mathbb{S}}
\newcommand{\W}{\text{\sf W}}
\newcommand{\Z}{\mathcal{Z}}
\newcommand{\Q}{\mathcal{Q}}
\newcommand{\OOO}{\mathrm{O}}
\newcommand{\D}{\mathcal{D}}
\newcommand{\model}{\mathcal{F}}
\DeclareMathOperator{\SW}{\text{\sf SW}}

\DeclareMathOperator{\ED}{\text{\sf ED}}
\DeclareMathOperator{\CRPS}{\text{\sf CRPS}}
\DeclareMathOperator{\ES}{\text{\sf ES}}
\newcommand{\LLL}{\mathcal{L}}
\newcommand{\al}[1]{%
  \begin{align}
    #1
  \end{align}
}
\def\hako#1{\begin{tabular}{l}#1\end{tabular}}

\allowdisplaybreaks

\Title{Multivariate Uncertainty Quantification with Tomographic Quantile Forests}

\Author{Takuya Kanazawa $^{1}$\orcidA{}}

\AuthorNames{Takuya Kanazawa}

\address[1]{%
$^{1}$ \quad Department of Business Administration, 
Kobe Gakuin University, 1-1-3 Minatojima, Chuo-ku, Kobe 650-8586, Japan; tkanazawa@ba.kobegakuin.ac.jp}

\corres{Correspondence: tkanazawa@ba.kobegakuin.ac.jp}

\abstract{Quantifying predictive uncertainty is essential for safe and trustworthy real-world AI deployment. Yet, fully nonparametric estimation of conditional distributions remains challenging for multivariate targets. We propose Tomographic Quantile Forests (TQF), a nonparametric, uncertainty-aware, tree-based regression model for multivariate targets. TQF learns conditional quantiles of directional projections $\nn^{\top}\yy$ as functions of the input $\xx$ and the direction $\nn$. At inference, it aggregates quantiles across many directions and reconstructs the multivariate conditional distribution by minimizing the sliced Wasserstein distance via an efficient alternating scheme with convex subproblems. Unlike classical directional-quantile approaches that typically produce only convex quantile regions and require training separate models for different directions, TQF covers all directions with a single model to reconstruct the full conditional distribution itself, naturally overcoming any convexity restrictions. We evaluate TQF on synthetic and real-world datasets, and release the source code on GitHub.}


\begin{document}

\tableofcontents

\section{Introduction}

In machine learning, probabilistic prediction aims to estimate an output distribution rather than a single point estimate, thereby enabling the quantification of predictive uncertainty \cite{Gneiting2014,Caldeira2020,Hullermeier2021,Abdar2021,Bjerregaard2021,He2026,Gawlikowski2023,Tyralis2024,Klein2024,Wang2025}. Such approaches are essential in high-stakes domains, including medical treatment, autonomous driving, and financial risk assessment, where understanding decision risk is critical. Uncertainty estimation is also central to Bayesian experimental design, in which one seeks to maximize information gain from new observations under cost constraints \cite{Garnett2023book}. Deep neural networks (NNs) have driven substantial progress in many of these areas, largely due to their ability to handle high-dimensional data effectively. However, in the setting of tabular data, whether NNs consistently outperform traditional machine learning models remains actively debated \cite{Gorishniy2021,Shwartz2022,Grinsztajn2022,McElfresh2023,Zabergja2024,Somvanshi2024,Ren2025}. In practice, gradient boosting machines \cite{Chen2016,Ke2017} often perform comparably to, or even better than, deep learning approaches while being more computationally efficient. This motivates the development of uncertainty quantification methods based on non-NN models.

\begin{figure}[htb]
	\centering
	\includegraphics[width=.75\textwidth]{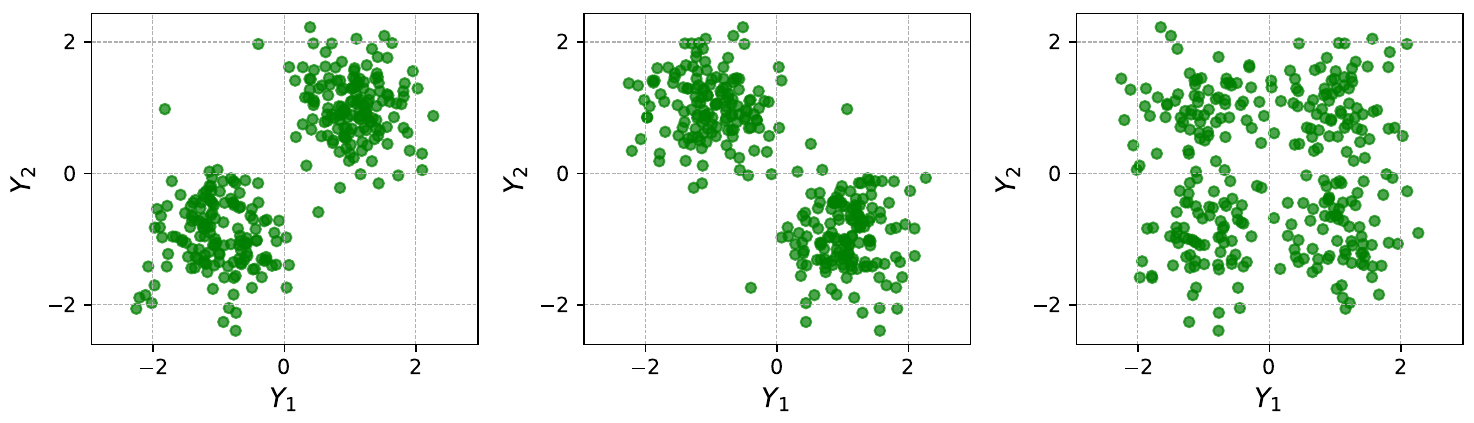}
	\caption{Toy datasets in $\RR^2$ containing 300 points with \emph{identical} marginal distributions.}\label{fg:toyex}
\end{figure}
For regression with a univariate target $y\in\RR$, predictive uncertainty is commonly summarized using confidence intervals or quantiles, or modeled using parametric distributions. In contrast, uncertainty quantification for a multivariate target $\yy\in\RR^d$ is considerably more challenging. A naïve approach is to model each component independently, but the resulting predictive distribution fails to capture dependencies among components; in general, $p(\yy|\xx)\neq\prod_{i=1}^d p(y_i|\xx)$. A didactic example is shown in Figure~\ref{fg:toyex}: the three datasets have clearly different joint distributions $p(y_1,y_2)$, yet their marginal distributions $p(y_1)$ and $p(y_2)$ coincide exactly. Ignoring such dependence can severely undermine reliability in downstream decision-making. Although conventional models assuming a Gaussian predictive distribution extend readily to multivariate targets, they inherently fail to capture multi-modality and heavy tails. More flexible parametric models such as Gaussian mixtures are often difficult to fit reliably and may converge to poor local optima. Recent work has therefore focused on NN-based approaches that address these limitations \cite{Bouchacourt2016,RussellR22,pmlr-v151-kan22a,DBLP:conf/ijcci/Kanazawa022,Vedula2023,DBLP:journals/jmlr/FeldmanBR23,Braun2025}. By comparison, multivariate uncertainty quantification with non-NN models has received relatively less attention.

To bridge this gap, we introduce a non-NN method for learning nonparametric, potentially multimodal predictive distributions in tabular-data settings. Our approach, a tomographic multivariate uncertainty quantification method termed Tomographic Quantile Forests (TQF), first produces probabilistic predictions for one-dimensional projections of the target, $p(\nn^{\top}\yy|\xx)$, over directions $\nn\in\SSS^{d-1}:=\{\xx\in \RR^d \mid \| \xx \|_2 = 1 \}$. It then aggregates these directional predictions to construct a multivariate predictive distribution $p(\yy|\xx)$, rather than estimating geometric quantile regions. TQF is motivated by the Cramér-Wold theorem \cite{CramerWold1936}, which states that a probability measure on $\RR^d$ is uniquely determined by its one-dimensional projections. While TQF can in principle use any probabilistic model for univariate targets as a backbone, we find it particularly effective to build on a variant of Quantile Regression Forests \cite{Meinshausen2006}, given their strong performance and favorable computational cost on large tabular datasets. We further show that TQF approximately minimizes the sliced 1-Wasserstein distance between the true conditional distribution and the predicted distribution. Finally, we validate TQF empirically on multiple benchmark problems and observe favorable performance. The source code for this paper is publicly available at GitHub.%
\footnote{\url{https://github.com/TaTKSM/TQF}}

The remainder of this paper is organized as follows. Section~\ref{sc:rwk} reviews prior work on uncertainty quantification in machine learning. Section~\ref{sc:qrfpp} introduces a modified forest approach, QRF++, and evaluates it numerically. Section~\ref{sc:mainpart} presents the proposed method, TQF, and its core component, the Quantile-Matching Empirical Measure (QMEM) algorithm, which reconstructs a probability distribution from directional quantile data. Section~\ref{sc:qmem} specifically evaluates the standalone performance of the QMEM algorithm. Section~\ref{sc:tqfn} reports numerical results for TQF on synthetic and real-world datasets to empirically validate performance. Finally, Section~\ref{sc:concl} presents conclusions. Appendix~\ref{sc:prel} summarizes technical background on distributional distance metrics, the Radon transform, and proper scoring rules.

\section{Related Work}\label{sc:rwk}

Table~\ref{tb:549gh} summarizes prior work on probabilistic prediction models, with an emphasis on non-neural approaches. We briefly review these methods in the following subsections. Readers interested in a more comprehensive overview of uncertainty quantification are referred to the survey articles in Refs.~\cite{Gneiting2014,Hullermeier2021,Abdar2021,He2026,Gawlikowski2023,Tyralis2024,Klein2024,Wang2025}.

\subsection{Univariate target variable}\label{sc:3x3sd5}

Quantile regression \cite{Koenker2005book} and expectile regression \cite{Newey1987} have been among the most widely used methods in statistics for evaluation of predictive errors and uncertainty. Since no parametric form of noise is assumed, they can in principle estimate heteroscedastic and multimodal predictive distributions. It is noteworthy that expectiles are the only risk measure that is both coherent and elicitable \cite{Artzner1999,Gneiting2011}, making them particularly well-suited for financial risk management. As linear regression \cite{Koenker2005book,Newey1987} can only estimate linear dependence on predictors, various nonlinear generalizations have been explored. For example, quantile regression NNs \cite{Taylor2000,Cannon2011} adopt artificial NNs, whereas Quantile Regression Forests (QRF) \cite{Meinshausen2006} train an ensemble of decision trees. In contrast to the original random forests \cite{Breiman2001} that record only the mean of the observations in each leaf, QRF keeps the value of all observations for subsequent quantile estimation. Generalized random forests \cite{Athey2019} is another generalization of random forests, which utilizes a tailored splitting criterion of trees and treats forests as a type of adaptive nearest neighbor estimator, demonstrating strong empirical performance. Gradient boosting decision trees such as XGBoost \cite{Chen2016} and LightGBM \cite{Ke2017} can also minimize quantile losses, but one model per one quantile is necessary, which works fine for confidence interval prediction but incurs high computational costs when the entire predictive distribution is to be estimated.

While classical Kernel Density Estimation (KDE) and density-ratio methods (e.g., Ref.~\cite{Sugiyama2010}) offer theoretically sound approaches for conditional density estimation, we do not treat them as direct baselines in our setting. Traditional KDE evaluating $P(Y|X)=P(X, Y)/P(X)$ suffers from zero-denominator instabilities and meaninglessly inflated distance metrics in the presence of irrelevant features. Although adapting bandwidths component-wise can mitigate this, it incurs prohibitive computational costs. Tree-based methods inherently bypass these limitations through adaptive spatial partitioning, making them more robust for high-dimensional inputs \cite{Hastiebook}.

\setlength{\tabcolsep}{1.8pt}
\begin{table}[H]
	\centering
	\begin{adjustwidth}{-0.05\extralength}{0cm}
	\caption{An inexhaustive list of statistical and machine-learning models for distributional prediction with univariate or multivariate target variables. The proposed method is listed in the bottom row. The column ``Nonparametric'' indicates whether a method can predict arbitrary (i.e., nonparametric) probability distributions; it does \emph{not} refer to whether the input–output relationship is modeled nonparametrically.}\label{tb:549gh}
	\hspace*{-2mm}
	\begin{tabular}{lrrrrrr}\toprule[0.8mm]
		& \hako{Admits\\$\mathsf{dim}\,\yy \! > \! 1$}
		& \hako{Nonpara-\\metric}
		& \hako{Use\\NNs}
		& \hako{Use\\decision\\trees}
		& \hako{Limited to\\convex\\predictive\\regions}
		& \hako{Curse of dimen-\\sionality when\\$\mathsf{dim}\,\xx$ is high}
		\\\midrule[0.5mm]
		Rasmussen \& Williams 2005 \cite{Rasmussen2005book} & Yes & No & No & No & Yes & No 
		\\\midrule
		Meinshausen 2006 \cite{Meinshausen2006} & No & Yes & No & Yes & No & No
		\\\midrule
		Sugiyama et al.~2010 \cite{Sugiyama2010} & Yes & Yes & No & No & No & Yes
		\\\midrule
		Hallin et al.~2010 \cite{Hallin2010} & Yes & Yes & No & No & Yes & No
		\\\midrule
		Paindaveine \& Šiman 2011 \cite{Paindaveine2011} & Yes & Yes & No & No & Yes & No		
		\\\midrule
		Kong \& Mizera 2012 \cite{Kong2012} & Yes & Yes & No & No & Yes & No		
		\\\midrule
		Bouchacourt et al.~2016 \cite{Bouchacourt2016} & Yes & Yes & Yes & No & No & No 
		\\\midrule
		Schlosser et al.~2019 \cite{Schlosser2019} & No & No & No & Yes & No & No
		\\\midrule
		Athey et al.~2019 \cite{Athey2019} & No & Yes & No & Yes & No & No
		\\\midrule
		\!\begin{tabular}{l}
			Duan et al.~2020 \cite{ngboost2020} and\\
			O’Malley et al.~2021 \cite{OMally2021}
		\end{tabular} & Yes & No & No & Yes & No & No
		\\\midrule
		Hothorn \& Zeileis 2017, 2021 \cite{Hothorn2017,Hothorn2021} & No & No & No & Yes & No & No
		\\\midrule
		Du et al.~2021 \cite{Du2021} & Yes & Yes & No & Yes & No & No
		\\\midrule
		\!\begin{tabular}{l}
			März 2019 \cite{Marz2019}, \\
			März \& Kneib 2022 \cite{Marz2022a},\\
			and  März 2022 \cite{Marz2022b}
		\end{tabular} 
		& Yes & Yes & No & Yes & No & No
		\\\midrule
		Russell \& Reale \cite{RussellR22} & Yes & No & Yes & No & Yes & No
		\\\midrule
		Kanazawa \& Gupta 2022 \cite{DBLP:conf/ijcci/Kanazawa022} & Yes & Yes & Yes & No & No & No
		\\\midrule
		Kan et al.~2022 \cite{pmlr-v151-kan22a} & Yes & Yes & Yes & No & No & No
		\\\midrule
		Vedula et al.~2023 \cite{Vedula2023} &Yes &Yes &Yes &No & No & No
		\\\midrule
		\!\begin{tabular}{l}
			Cevid et al.~2022 \cite{Cevid2022} and
			\\
			Näf et al.~2023 \cite{Naf2023}
		\end{tabular} 
		& Yes & Yes & No & Yes & No & No
		\\\midrule
		Feldman et al.~2023 \cite{DBLP:journals/jmlr/FeldmanBR23} & Yes & Yes & Yes & No & No & No
		\\\midrule
		Chen \& Müller 2023 \cite{Chen2023} & Yes & Yes & No & No & No & Yes
		\\\midrule
		Matsubara 2024 \cite{Matsubara2024} & No & Yes & No & Yes & No & No
		\\\midrule
		Barrio et al.~2024 \cite{Barrio2024} & Yes & Yes & No & No & No & Yes
		\\\midrule
		{\bf This work} & Yes & Yes & No & Yes & No & No
		\\\bottomrule[0.8mm]
	\end{tabular}
	\end{adjustwidth}
\end{table}

There are also models that exploit the Wasserstein distance, which stems from the optimal transport theory and has gained popularity in the machine learning community \cite{Villani2003book,Villani2009book,COTFNT,
Montesuma2025,Peyre2025}. Wasserstein random forests \cite{Du2021} employs a novel splitting criterion of trees based on the Wasserstein distance. Wasserstein gradient boosting \cite{Matsubara2024} trains weak learners successively via Wasserstein gradients and nonparametrically solves regression problems with distribution-valued responses.

Besides these nonparametric density estimation methods, there is another class of methods that aim to fit a parametric family of probability distributions to the target. A classical and most established example is the Gaussian Processes (GP) \cite{Rasmussen2005book}. Gradient boosting-based models such as NGBoost \cite{ngboost2020} and XGBoostLSS \cite{Marz2019} allow fitting various parametric (e.g., Gaussian, Laplace, Lognormal, and Gaussian mixture) distributions. Distributional regression forests \cite{Schlosser2019} learn parameters of a zero-censored Gaussian distribution from data, with a focus on meteorological applications. 
Transformation forests \cite{Hothorn2017,Hothorn2021} posit a monotone transformation that maps the target to a known base distribution and fit its parameters locally with trees using likelihood-based splits, yielding a smooth predictive CDF.

Recent years have witnessed remarkable progress in uncertainty quantification techniques for deep learning, such as Bayesian NNs, Monte Carlo dropout and deep ensembles, as reviewed in  Refs.~\cite{Caldeira2020,Hullermeier2021,Abdar2021,He2026,Gawlikowski2023,Tyralis2024,Wang2025}. A fundamental insight emerging from these lines of work is that uncertainty in machine learning generally comprises both \emph{aleatoric uncertainty} and \emph{epistemic uncertainty} \cite{Hullermeier2021}. The former is related to intrinsic stochasticity of the data generating process, which can be precisely modeled with quantile regression or parametric distribution fitting, whereas the latter represents our lack of knowledge and can be reduced by gathering more data; the latter uncertainty is high for out-of-distribution samples and hence its quantification is crucial in the task of anomaly (or outlier) detection. Whether NNs are used or not, it is generally difficult to capture both types of uncertainty within a single model. Tree-based methods, including gradient boosting and random forests, are not well suited for measuring epistemic uncertainty because trees are poor extrapolators by construction. Our work in this paper also uses trees, and we assume that the main application of our model should be in the area of aleatoric uncertainty quantification.

\subsection{Multivariate target variable}

Multioutput regression models have a long history of research \cite{Borchani2015,Waegeman2019,Xu2020}.
It is notoriously difficult to extend the concept of quantiles to more than one dimensions \cite{Serfling2002,Carlier2016}, hampering nonparametric distributional prediction for multidimensional outputs. Directional quantile regression (DQR) \cite{Hallin2010,Paindaveine2011,Kong2012} is an approach that computes directional half-spaces and takes their intersections to construct quantile regions in arbitrary dimensions. However, DQR by construction only yields convex quantile regions and cannot describe general nonconvex features. More recent work \cite{Hallin2021,Barrio2024} developed the theory of center-outward quantile regions, which proposes a novel multivariate quantile concept and overcomes DQR's limitation of convexity. However, its current implementation \cite{Barrio2024} is based on classic kernel weighting and is vulnerable to the curse of dimensionality when the predictor resides in high dimensions. Unlike these prior methods that primarily target the geometric boundaries of quantile regions, our focus is on estimating the full conditional distribution $p(\yy|\xx)$.

There are several lines of work in deep learning that addresses the challenge of multivariate probabilistic predictions. Refs.~\cite{pmlr-v151-kan22a} and \cite{Vedula2023} propose to formulate multivariate quantile functions as the gradient of partially input-convex NNs. Ref.~\cite{DBLP:journals/jmlr/FeldmanBR23} enhances DQR by representation learning via a conditional variational autoencoder, so that arbitrary nonconvex probability regions can be predicted. Ref.~\cite{RussellR22} integrates NNs with a Kalman filter for applications to a visual tracking problem. Refs.~\cite{Bouchacourt2016} and \cite{DBLP:conf/ijcci/Kanazawa022} employ implicit generative NNs, which receive a noise vector as additional input and generate dispersed predictions. There are more references in the domain of probabilistic multivariate time-series forecasting, e.g., \cite{pmlr-v139-rasul21a}.

Compared to deep learning, tree-based approaches to multivariate probabilistic predictions have been relatively less explored. Refs.~\cite{OMally2021,Marz2022b} extend gradient boosting so that parameters of multivariate distributions can be learned. Wasserstein random forests \cite{Du2021} offers a natural extension of random forests to multivariate targets. Distributional Random Forest (DRF) \cite{Cevid2022,Naf2023} grows an ensemble of decision trees with a distribution-aware splitting criterion based on the Maximal Mean Discrepancy metric \cite{Gretton2006,Gretton2012}. This is in marked contrast to QRF \cite{Meinshausen2006}, which uses the plain CART splitting criterion of the original random forests \cite{Breiman2001}. DRF dynamically decides weights for each training data point and estimates the predictive distribution by an empirical measure determined by training points and their weights. 

Fréchet regression \cite{Petersen2019} is a general framework for regressing targets that lie in non-Euclidean metric spaces. Ref.~\cite{Chen2023} extended this framework to multivariate distributions by using the sliced Wasserstein distance \cite{Rabin2011,Bonneel2015}, which provides a computationally efficient surrogate for the \emph{bona fide} Wasserstein distance. Specifically, Ref.~\cite{Chen2023} proposed four variants: GSWW, GSAW, LSWW, and LSAW. The \emph{global} variants (GSWW and GSAW) generalize linear regression to metric spaces, whereas the \emph{local} variants (LSWW and LSAW) use kernel smoothing to perform nonparametric local linear regression.

We highlight two technical differences between Ref.~\cite{Chen2023} and the present work. First, Ref.~\cite{Chen2023} provides strong theoretical guarantees, but its linear modeling assumptions may limit flexibility and can exacerbate the curse of dimensionality in high-dimensional settings. In contrast, our approach uses fully nonparametric, tree-based models, which can alleviate these limitations in practice. Second, GSWW and LSWW in Ref.~\cite{Chen2023} apply Fréchet regression independently to each Radon slice, whereas our method trains a single model that jointly covers all Radon slices.

\section{QRF++: A Simplified DRF Implementation}\label{sc:qrfpp}

\subsection{Definition}

This section introduces a slight modification of the original QRF model \cite{Meinshausen2006} with output-space augmentation. Let $\{(\xx_i,y_i)\}_{i=1}^{N}$ be training data with a univariate response $y_i\in\RR$. QRF builds an ensemble of CART trees using bootstrapped data and the squared-error splitting criterion; at prediction time, QRF estimates the conditional CDF $F(y | \xx)$ by aggregating training points that fall into the same leaves as the query $\xx$. As emphasized by Refs.~\cite{Athey2019,Cevid2022}, such univariate targets make splits primarily sensitive to changes in the conditional mean and they may overlook other distributional changes (e.g., variance, tail or multi-modality shifts).

While more sophisticated forest models \cite{Athey2019,Cevid2022} overcome such a limitation, we here propose a minimalist's remedy that requires no algorithmic changes and can be implemented with any off-the-shelf QRF package: augment the learning target from $y\in\RR$ to 
\al{
	\mkakko{
		y, \cos \frac{y}{w_1}, \sin \frac{y}{w_1}, 
		\cdots, \cos \frac{y}{w_T}, \sin \frac{y}{w_T}
	} \in \RR^{2T + 1}\,,
	\label{eq:comy}
}
followed by a per-component normalization so that all $2T+1$ components have comparable variance. We then train a multioutput forest using the average sum of squares across multiple outputs as the splitting criterion \cite{Death2002,Kocev2007}. The parameters $w_t>0$ determine the inverse frequency of random Fourier features. At prediction time we simply take the first component of the model's output and ignore the rest. We refer to this method as QRF++.

The rationale behind QRF++ is as follows. Consider a trivial example where $p(y | x)$ changes abruptly at $x=0$, from $\mathcal{N}(0,1^2)$ for $x<0$ to $\frac{1}{2}\delta_{-\pi}+\frac{1}{2}\delta_{+\pi}$ for $x>0$. Since both sides have $\EE[y \,|\, x]=0$, a univariate CART split need not occur at $x=0$. In contrast, $\EE[\cos y \,|\, x]$ jumps from $\rme^{-1/2}\simeq 0.606$ to $-1$ at $x=0$, creating a strong signal. More generally, one can uniquely determine the full distribution once the characteristic function $\EE[\rme^{i \alpha y} \,|\, x]$ is obtained for every $\alpha \in \RR$. Sampling a few frequencies as an approximation leads to \eqref{eq:comy}. 

Conceptually, QRF++ is a variant of DRF \cite{Cevid2022}. DRF selects splits by maximizing MMD in a chosen RKHS. If one chooses a linear kernel over the feature space \eqref{eq:comy}, the resulting DRF would match QRF++. Eq.~\eqref{eq:comy} can also be viewed as random Fourier features stemming from some stationary kernel through Bochner's theorem. For mathematical details, we refer the reader to Ref.~\cite[Theorem~1]{Cevid2022}. Thus QRF++ is in essence a compact lightweight DRF implementation.%
\footnote{To be precise, QRF++ does not follow the honesty principle \cite{Denil2014,Wager2018} in building trees, unlike DRF. Moreover, QRF++ uses bootstrapping, while DRF uses subsampling without replacement.}

An important caveat is that \eqref{eq:comy} does not define a \emph{characteristic kernel}: it is not possible to distinguish two \emph{arbitrary} distributions based on \eqref{eq:comy} as long as $T<\infty$. Nevertheless, as our experiments in the next subsection show, the minimal modification \eqref{eq:comy} substantially improves QRF's sensitivity to various distributional changes that are overlooked by vanilla QRF.

\subsection{Evaluation of QRF++}\label{sc:yfdlqs}

We compare QRF and QRF++ on four benchmark datasets. In all the datasets, the predictor $\xx\in\RR^{40}$ is sampled from a uniform distribution over $[-1,1]^{40}$. $39$ features are just noise, and only one feature is associated with the target $y$. Characteristics of the datasets (a)--(d) are summarized below.
\bd
	\item[(a)] $y|x \sim \mathcal{N}\mkakko{0,1^2}$ for $x<0$ and $\mathcal{N}\mkakko{0,3^2}$ for $x\geq 0$. The mean is constant but the variance jumps at $x=0$. There are 1,500 samples.
	\item[(b)] $y|x \sim \mathcal{N}\mkakko{0,1^2}$ for $x<0$ and $\frac{1}{2}\mkakko{
		\mathcal{N}\mkakko{\mu,\sigma^2} + \mathcal{N}\mkakko{-\mu,\sigma^2}
	}$ for $x\geq 0$, where $\mu=0.95$ and $\sigma =\sqrt{1-\mu^2}$. The mean and variance are $0$ and $1$ for all $x$. The distribution is unimodal for $x<0$ and bimodal for $x\geq 0$. There are 3,000 samples.
	\item[(c)] $y|x \sim \mathrm{Uniform}\mkakko{[-(x+2.5)^2, (x+2.5)^2]}$. This dataset exhibits a funnel-like shape when plotted in the $xy$-plane. There are 3,500 samples.
	\item[(d)] $y|x \sim 1 - \mathrm{Exp}(1)$ for $x<0$ and $\mathrm{Exp(1)} - 1$ for $x\geq 0$, where $\mathrm{Exp}(\lambda)$ is the exponential distribution with the rate parameter $\lambda$. The mean and variance are $0$ and $1$ for all $x$. There are 1,700 samples.
\ed
While there are a few publicly available QRF implementations \cite{scikitgarden,quantreg}, we use the Cython-optimized Python package \texttt{quantile-forest} \cite{Johnson2024}. We fix the number of trees to 100 and prevent overfitting by setting \verb|min_samples_leaf| $=30$. As for \eqref{eq:comy}, we set $T=3$ and let $(w_1,w_2,w_3)=(w_{\rm med}/2, w_{\rm med}, 2w_{\rm med})$, where $w_{\rm med}$ is the median pairwise distance of $\{y_i\}$, motivated by the median heuristic \cite{Gretton2012,DBLP:conf/nips/GrettonSSSBPF12,Bischoff2024review}.

\begin{figure}[tb]
\begin{adjustwidth}{0\extralength}{0cm}
	\centering
	\includegraphics[width=.23\textwidth]{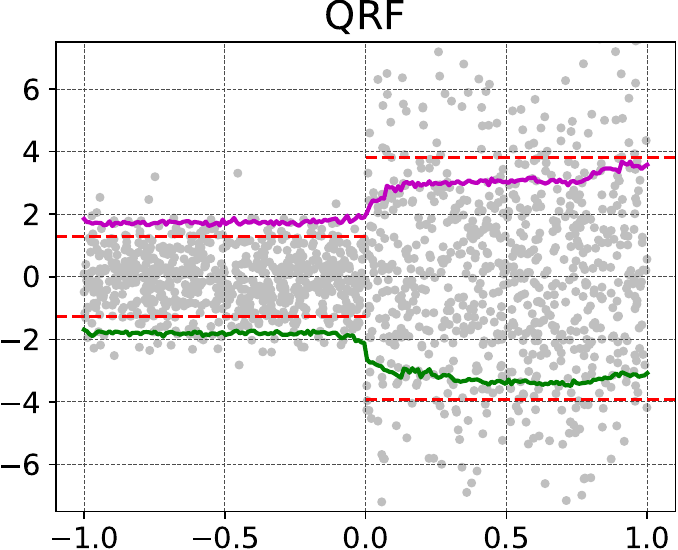}~
	\includegraphics[width=.23\textwidth]{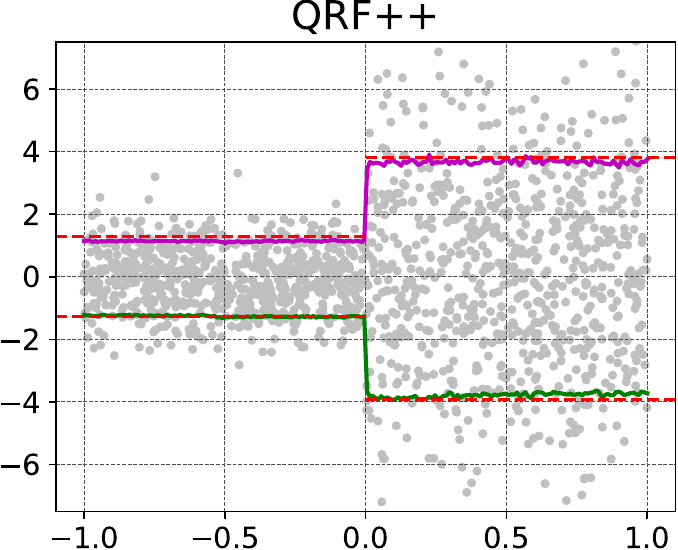}
	\quad~ 
	\includegraphics[width=.23\textwidth]{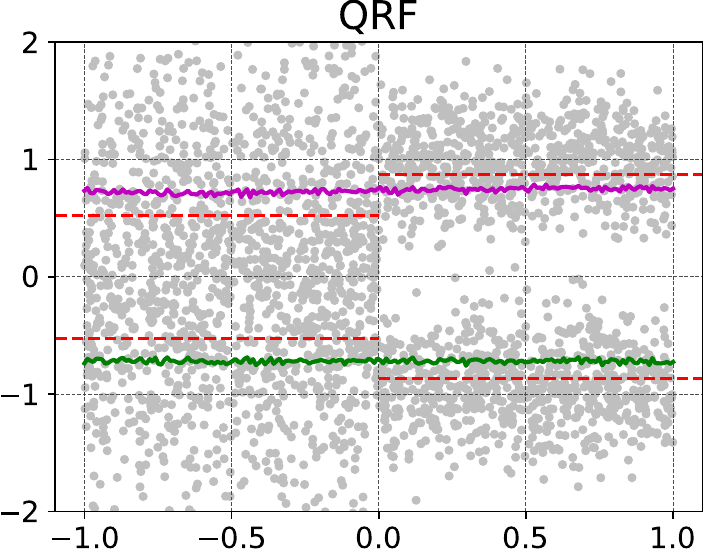}~
	\includegraphics[width=.23\textwidth]{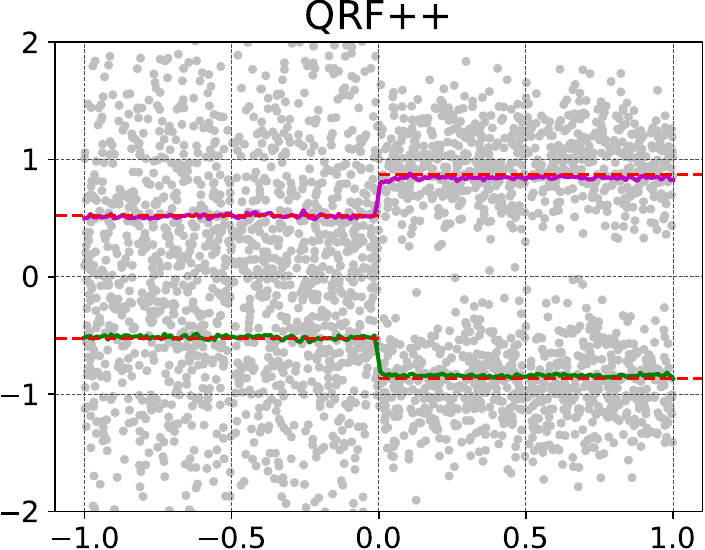}
	\put(-438, 83){\large (a)}
	\put(-217, 83){\large (b)}
	\vspace{5mm}\\
	\includegraphics[width=.23\textwidth]{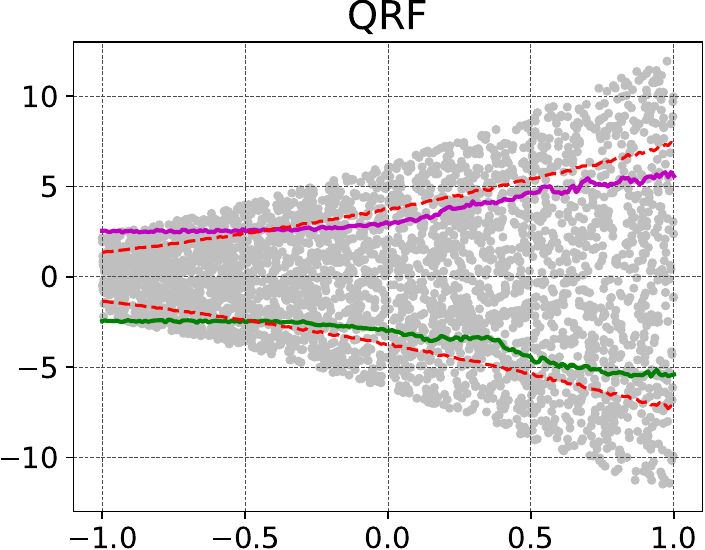}
	\includegraphics[width=.23\textwidth]{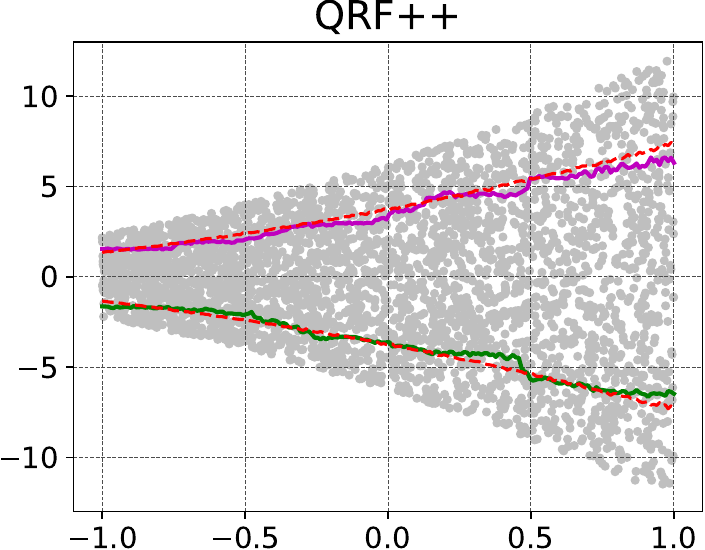}~
	\quad~
	\includegraphics[width=.23\textwidth]{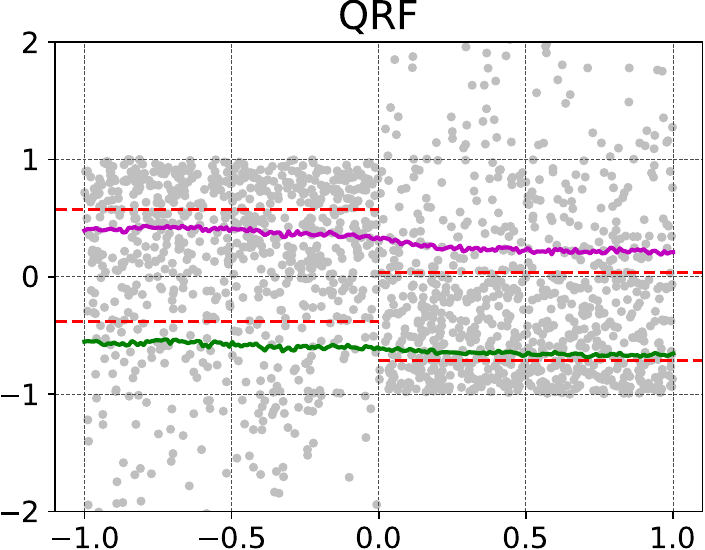}~
	\includegraphics[width=.23\textwidth]{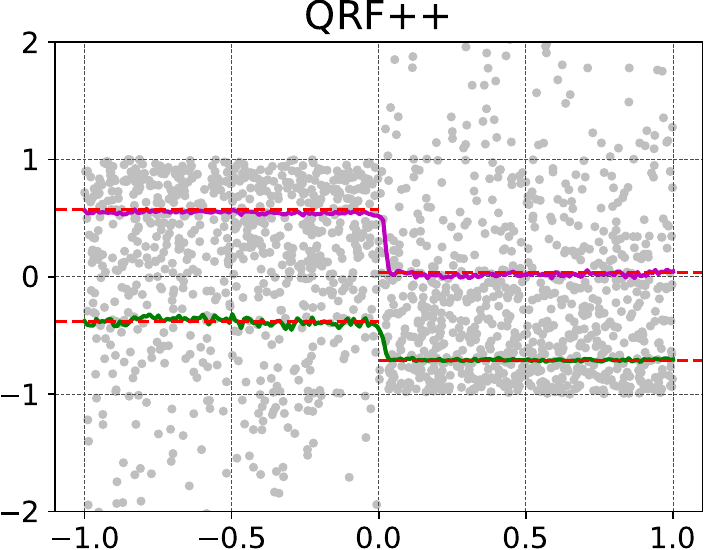}
	\put(-438, 83){\large (c)}
	\put(-217, 83){\large (d)}
	\vspace{\baselineskip}
	\caption{Illustration of the predicted quantiles from QRF (left panels) and QRF++ (right panels) on four synthetic datasets. 
The plotted quantile levels are (a) 0.10 and 0.90, (b) 0.30 and 0.70, (c) 0.20 and 0.80, and (d) 0.25 and 0.65. 
Red dashed lines show the ground truth; solid lines (magenta and green) show the model predictions. Gray points indicate the observed samples.}\label{fg:qrfpp_8}
\end{adjustwidth}
\end{figure}

Numerical results are shown in Figure~\ref{fg:qrfpp_8}. Across experiments (a)–(d), the vanilla QRF produces poor approximations to the true quantiles, consistent with the fact that a constant zero mean prevents the standard CART splitting criterion from capturing distributional variation. By contrast, QRF++ delivers accurate predictions in all cases. In particular, in (b) and (d) QRF++ sharply recovers the jumps in the quantiles, even though the mean and variance remain unchanged.

We examined the average tree depth and training time for both models. 
On average, QRF++ trees were 18\% shallower than those of QRF, and QRF++ trained 14\% faster. 
The likely explanation is that the output-augmented splits in QRF++ identify stronger partitions earlier, yielding more balanced trees. Although multioutput targets add a small constant overhead to each split evaluation, the reduction in per-node sample counts dominates, resulting in the observed speedup.

It is natural to ask which components of the augmented target most strongly shape the trees. 
We quantify this via a component-wise decomposition of the impurity decrease at each split under the multi-output CART criterion (summed squared error). 
Let a node be split into left/right children with sample sets $\mathcal{I}_P,\mathcal{I}_L,\mathcal{I}_R$ and let $Y^{(m)}$ denote the $m$-th target coordinate. 
Writing $\overline Y^{(m)}_{\#}$ for the (weighted) mean on each set, the reduction in sum of squared errors attributable to coordinate $m$ at that split is
\al{
  \Delta^{(m)} \;=\; 
  \sum_{i\in\mathcal{I}_P}\!\bigl(Y^{(m)}_i
  - \overline Y^{(m)}_{P}
  \bigr)^2 \;-\;
  \sum_{i\in\mathcal{I}_L}\!\bigl(Y^{(m)}_i
  - \overline Y^{(m)}_{L}\bigr)^2
  \;-\;
  \sum_{i\in\mathcal{I}_R}\!\bigl(Y^{(m)}_i
  - \overline Y^{(m)}_{R}\bigr)^2 .
}
The node’s total gain is $\Delta=\sum_m \Delta^{(m)}$. 
For a given tree $t$, we define the (raw) \emph{target importance} of coordinate $m$ as the sum over all internal nodes,
\al{
	G^{(m)}_t \;=\; \sum_{\text{nodes }v\in t} \Delta^{(m)}_v,
}
and normalize to obtain
\al{
	I^{(m)}_t \;=\; \frac{G^{(m)}_t}{\sum_{m'} G^{(m')}_t}
	\,, 
	\qquad \sum_m I^{(m)}_t = 1.
}
Averaging over all trees in the forest yields the forest-level importance 
$\overline I^{(m)}=\frac{1}{T}\sum_{t=1}^{T} I^{(m)}_t$, which sums to one across $m$ and reflects the relative contribution of each target coordinate to split selection during training.
This notion is directly analogous to the \emph{feature importance} (defined for input variables), except that here the roles of inputs and outputs are reversed. It is recommended to standardize each target coordinate before training so that importances are not dominated by scale. 

\begin{figure}[t]
	\centering
	\includegraphics[width=.88\textwidth]{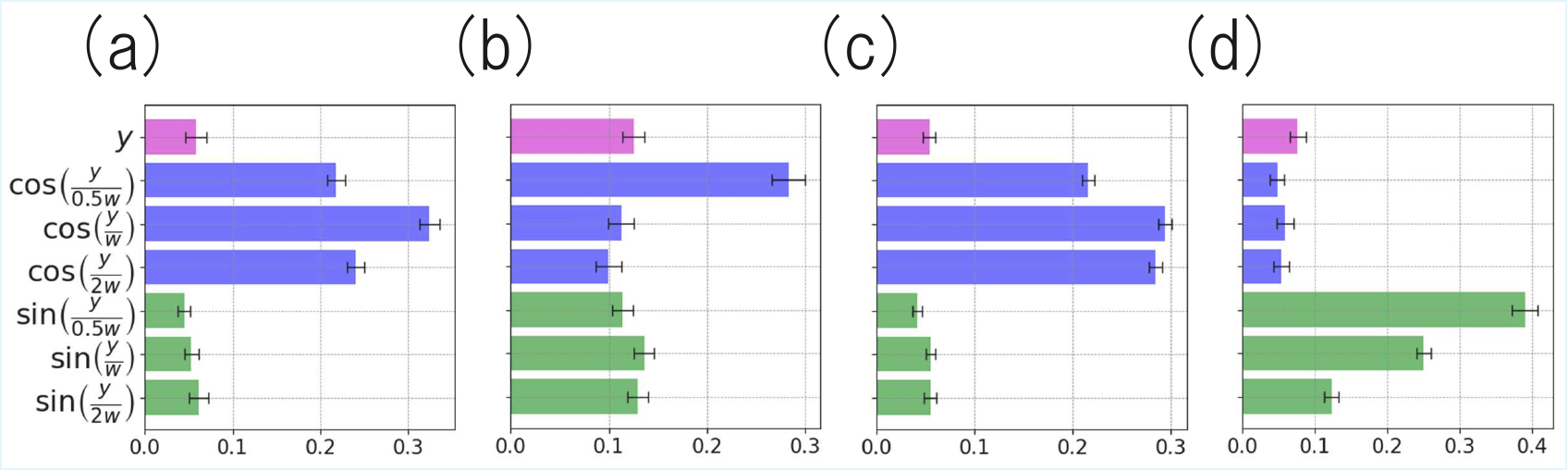}
	\caption{Target importance of QRF++ on the synthetic datasets; error bars show one standard deviation across 100 trees. The importances are normalized to sum to 1. Comments on each panel are given in the main text.}\label{fg:ti}
\end{figure}

The resulting target importances for the four synthetic datasets are shown in Figure~\ref{fg:ti}. Across all cases, the raw target $y$ carries an importance of about $0.1$ or less, indicating that the $2T$ auxiliary components dominate split decisions. In panels (a), (b), and (c), the importances of the $\sin$ components are small, which is consistent with the symmetry of the true conditional distributions around $y=0$. In panel (d), by contrast, the $\sin$ components receive large importances, as expected from the distributional flip $p(y\mid x>0)=p(-y\mid x<0)$ that creates an odd (sign-reversing) signal.

\subsection{Prediction Interval Coverage and Width}

Prediction Intervals (PIs) provide an intuitive approach to addressing the limitations of point predictions and are widely adopted in real-world forecasting tasks. Given their practical importance, it is crucial to verify that our proposed model, QRF++, generates reliable PIs. To evaluate this capability, we employ the Prediction Interval Coverage Probability (PICP) and Mean Prediction Interval Width (MPIW), two standard metrics widely recognized in the literature \cite{Khosravi2010,Pearce2018,Gawlikowski2023,Sluijterman2024,Wang2025,Wibbeke2025}. Suppose that a model generates a PI $[\wh{y}_i^{\rm L}, \wh{y}_i^{\rm U}]$ with bounds $\wh{y}_i^{\rm L} < \wh{y}_i^{\rm U}$ at a nominal confidence level $\lambda \in [0, 1]$ for each observation $(x_i, y_i)$ in a dataset of size $N$. These metrics are defined as follows:
\al{
	\text{PICP}(\lambda) & = \frac{1}{N}\sum_{i=1}^{N}
	\mathbbm{1}_{\{ \wh{y}_i^{\rm L} \leq y_i \leq \wh{y}_i^{\rm U} \}}\,,
	\\
	\text{MPIW}(\lambda) & = \frac{1}{N}\sum_{i=1}^{N}
	(\wh{y}_i^{\rm U}-\wh{y}_i^{\rm L})\,.
}
Intuitively, a well-calibrated prediction should yield an empirical coverage that closely matches the nominal level, meaning $\text{PICP}(\lambda) \approx \lambda$ for any specified $\lambda$. Furthermore, MPIW quantifies the average width of the PIs, serving as a measure of predictive sharpness. Since artificially widening the intervals trivially guarantees high coverage at the cost of informational value, and excessively narrow intervals fail to achieve the target coverage, a high-quality model must strike an optimal balance—minimizing the MPIW while strictly maintaining the target PICP.

In QRF++, a PI at a confidence level of $\lambda$ is constructed directly from the lower $(1-\lambda)/2$ and upper $(1+\lambda)/2$ quantiles of the predictive distribution. For instance, setting $\lambda=0.8$ yields a quantile range of $[0.1, 0.9]$.

\setlength{\tabcolsep}{5pt}
\begin{table}[tb]
	\caption{Datasets with univariate target, used to evaluate PI in this study. The \texttt{\bf concrete} dataset contained multiple duplicate instances and their removal reduces the dataset size to 1005.}
	\label{tb:data_pi}
	\centering
	\begin{tabular}{lrrr}
		\toprule
		Dataset & \# Instances & \# Input Features & Reference
		\\\midrule
		\texttt{\bf concrete} & 1030 & 8 & \cite{concrete_compressive_strength_165}
		\\
		\texttt{\bf diabetes} & 442 & 10 & \cite{Efron2004}
		\\
		\texttt{\bf airfoil} & 1503 & 5 & \cite{airfoil_self-noise_291}
		\\\bottomrule
	\end{tabular}
\end{table}

Table~\ref{tb:data_pi} summarizes the three real-world datasets used in our evaluation. While the \texttt{\bf concrete} and \texttt{\bf airfoil} datasets are relatively easy to predict ($\mathrm{R}^2\gtrsim 0.9$), the \texttt{\bf diabetes} dataset presents a more challenging task ($\mathrm{R}^2 \lesssim 0.5$). Note that our evaluation emphasizes PI quality over point forecast accuracy. As baselines, we evaluate both GP and standard QRF; however, since QRF and QRF++ perform similarly on these datasets, we report only the results of GP and QRF++ for brevity. Finally, all data were standardized prior to training to ensure numerical stability.

\paragraph{\bf Hyperparameters}

For the GP model, we utilized a Matérn kernel ($\nu=3/2$) with automatic relevance determination to adjust feature-specific length scales. In QRF++, the forest size was set to 300 trees for the \texttt{\bf concrete} and \texttt{\bf airfoil} datasets, and 100 trees for the \texttt{\bf diabetes} dataset. We evaluated the models using repeated random sub-sampling, partitioning the data into 80\% training, 10\% validation, and 10\% test sets over 20 independent trials, and reported the averaged results. While QRF++ regularization was managed via the \verb|max_leaf_nodes| parameter, the model interestingly achieved its best $\mathrm{R}^2$ performance on these datasets without any regularization.

\begin{figure}[tbh]
\begin{adjustwidth}{0\extralength}{0cm}
	\centering
	\includegraphics[width=.24\textwidth]{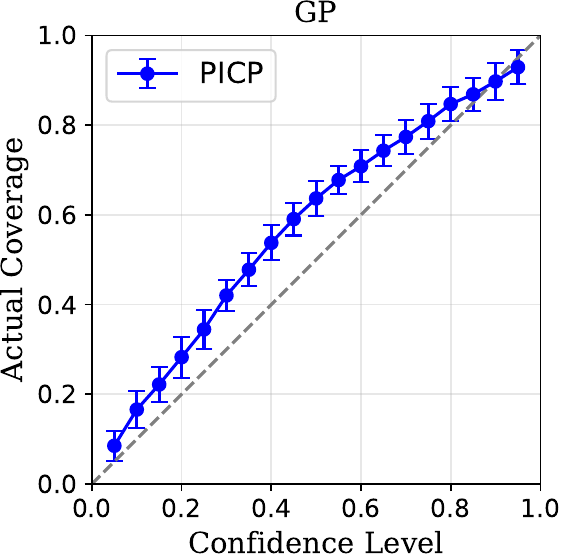}
	\includegraphics[width=.24\textwidth]{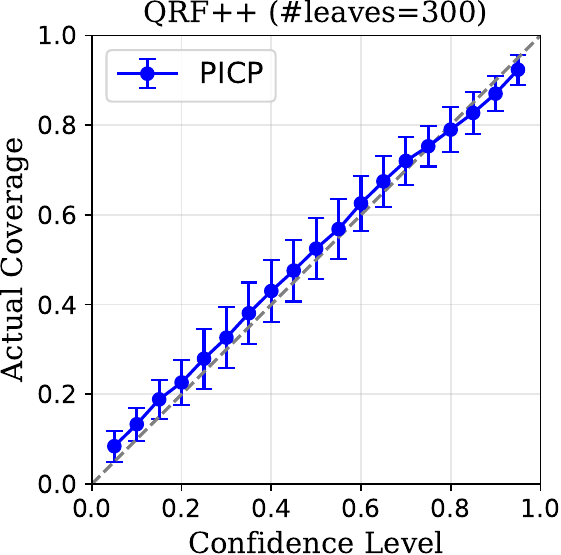}
	\includegraphics[width=.24\textwidth]{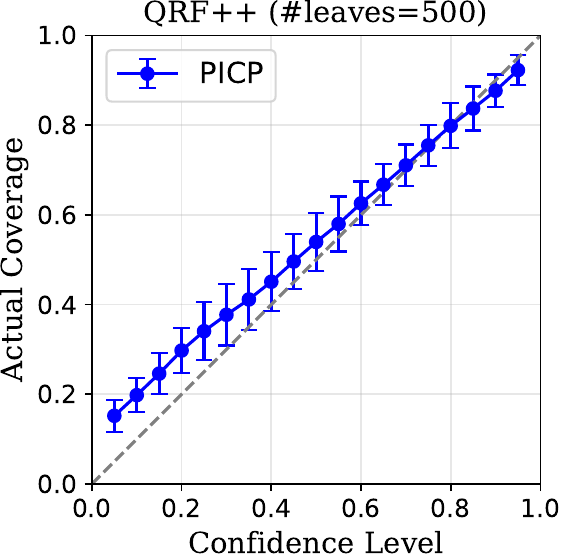}
	\includegraphics[width=.24\textwidth]{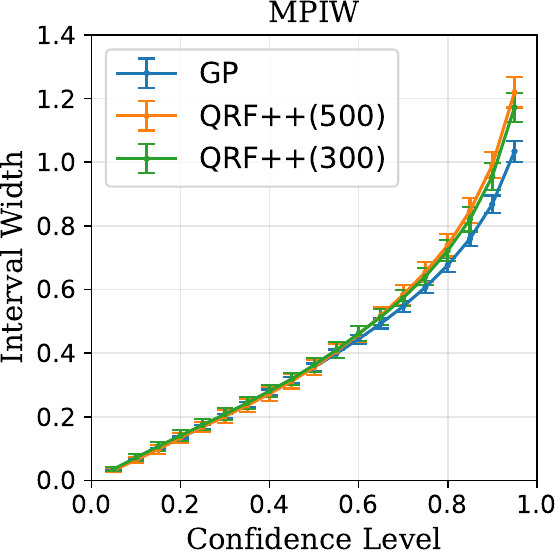}
	\caption{PICP and MPIW plots for the \texttt{\bf concrete} dataset. The gray dashed lines in the left three panels denote the theoretical coverage. Error bars represent one standard deviation across 20 runs.}
	\label{fg:concrete_picp}
\end{adjustwidth}
\end{figure}

\paragraph{\bf Results}

Figure~\ref{fg:concrete_picp} displays the results for the \texttt{\bf concrete} dataset. For the GP model (leftmost panel), the actual coverage is notably larger than the nominal confidence level, implying that the model is under-confident. For QRF++, we observe that when the regularization is strong (\verb|max_leaf_nodes| $=300$), the PICP curve closely matches the oracle line, indicating good calibration. Conversely, under weak regularization (\verb|max_leaf_nodes| $=500$), the PICP deviates from the oracle line at low confidence levels. This suggests that pursuing the best $R^2$ score may hinder achieving optimal calibration. Finally, the MPIW curves (rightmost panel) for the three models closely overlap.

\begin{figure}[tbh]
	\centering
	\includegraphics[width=.4\textwidth]{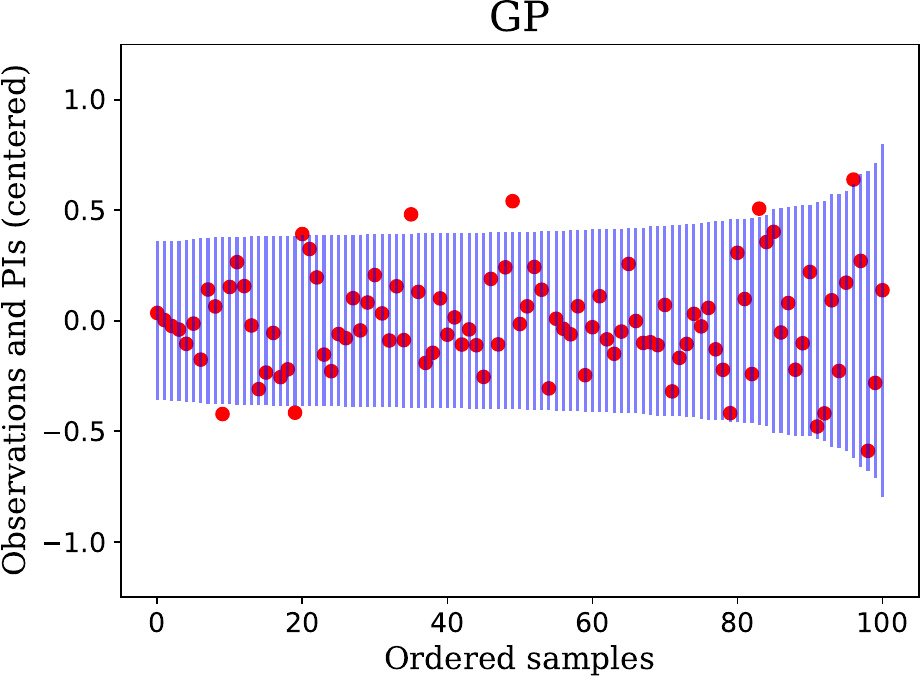}
	\qquad
	\includegraphics[width=.4\textwidth]{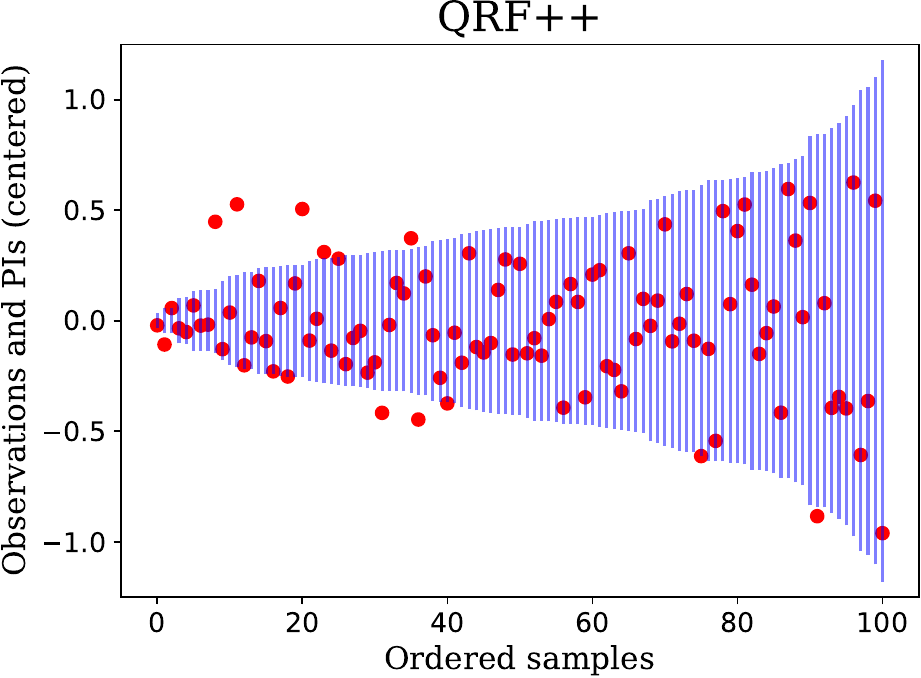}
	\caption{Test samples from the \texttt{\bf concrete} dataset, ordered by the width of their 90\% PIs. The blue bars represent the PIs, and the red points denote the actual observations. For clarity, the center of each PI has been subtracted from both the observations and the interval bounds, centering the PIs at zero.}
	\label{fg:concrete_ordered}
\end{figure}

Figure~\ref{fg:concrete_ordered} compares the PIs produced by the GP model and unregularized QRF++. While the GP model yields PIs of nearly constant width, the widths of the QRF++ intervals vary considerably across samples.

\begin{figure}[tbh]
\begin{adjustwidth}{-.0\extralength}{0cm}
	\centering
	\includegraphics[width=.3\textwidth]{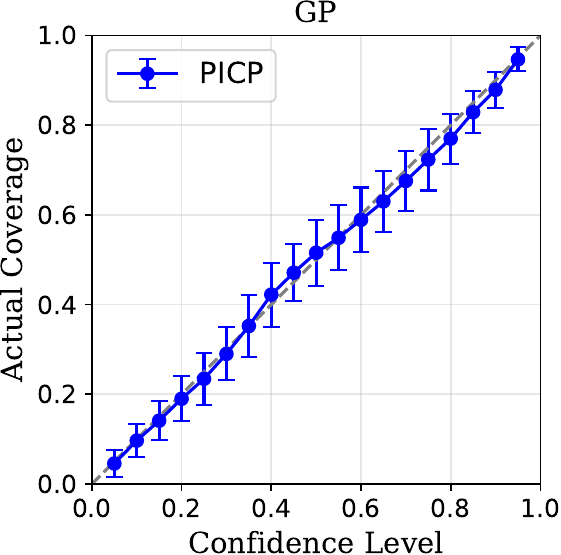}
	\includegraphics[width=.3\textwidth]{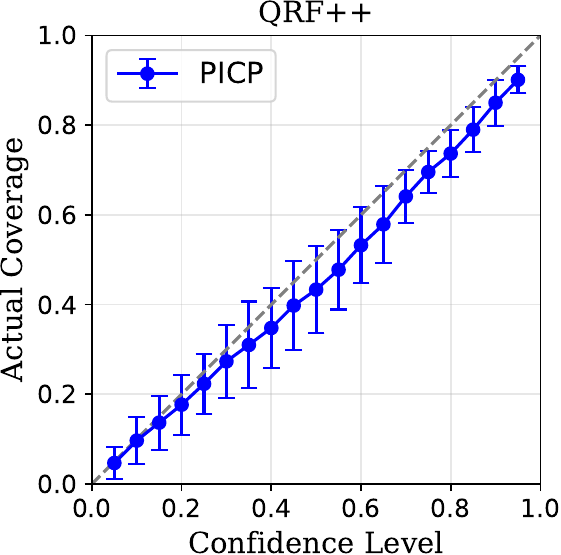}
	\includegraphics[width=.3\textwidth]{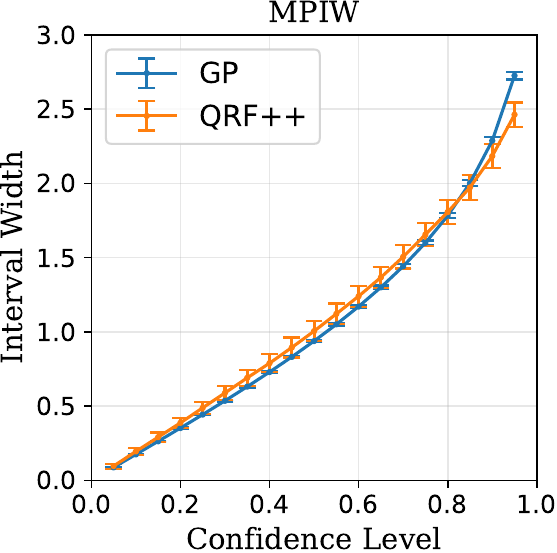}
	\caption{Same as Figure~\ref{fg:concrete_picp} but for the \texttt{\bf diabetes} dataset.}
	\label{fg:diabetes_picp}
\end{adjustwidth}
\end{figure}

Figure~\ref{fg:diabetes_picp} displays the results for the \texttt{\bf diabetes} dataset. In this case, the GP model is well-calibrated, whereas QRF++ appears to be slightly over-confident, as its actual coverage falls slightly below the nominal confidence level. Unlike the results for the \texttt{\bf concrete} dataset, the QRF++ curves did not exhibit a noticeable dependence on regularization.

\begin{figure}[tbh]
\begin{adjustwidth}{-.0\extralength}{0cm}
	\centering
	\includegraphics[width=.3\textwidth]{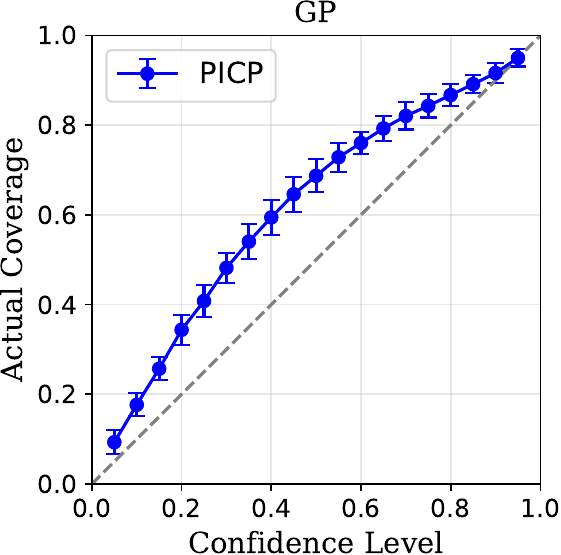}
	\includegraphics[width=.3\textwidth]{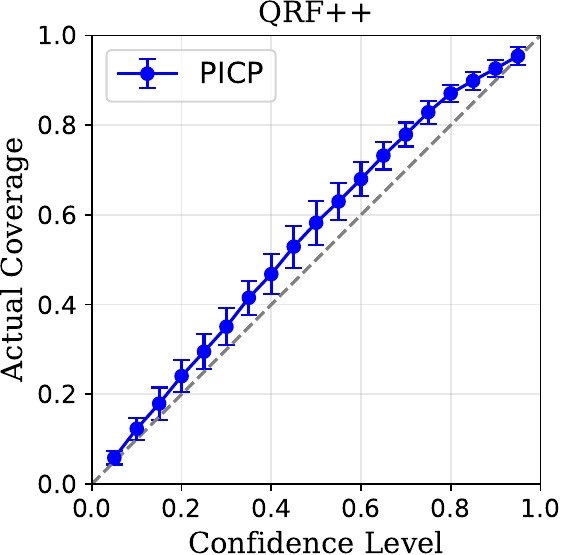}
	\includegraphics[width=.3\textwidth]{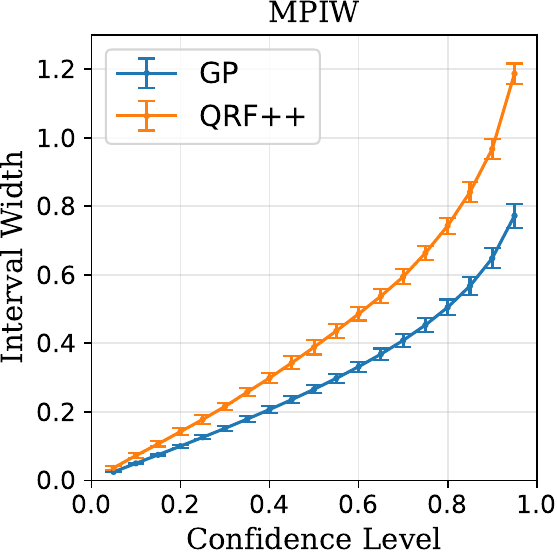}
	\caption{Same as Figure~\ref{fg:concrete_picp} but for the \texttt{\bf airfoil} dataset.}
	\label{fg:airfoil_picp}
\end{adjustwidth}
\end{figure}

\begin{figure}[tbh]
\vspace*{3mm}
	\centering
	\includegraphics[width=.4\textwidth]{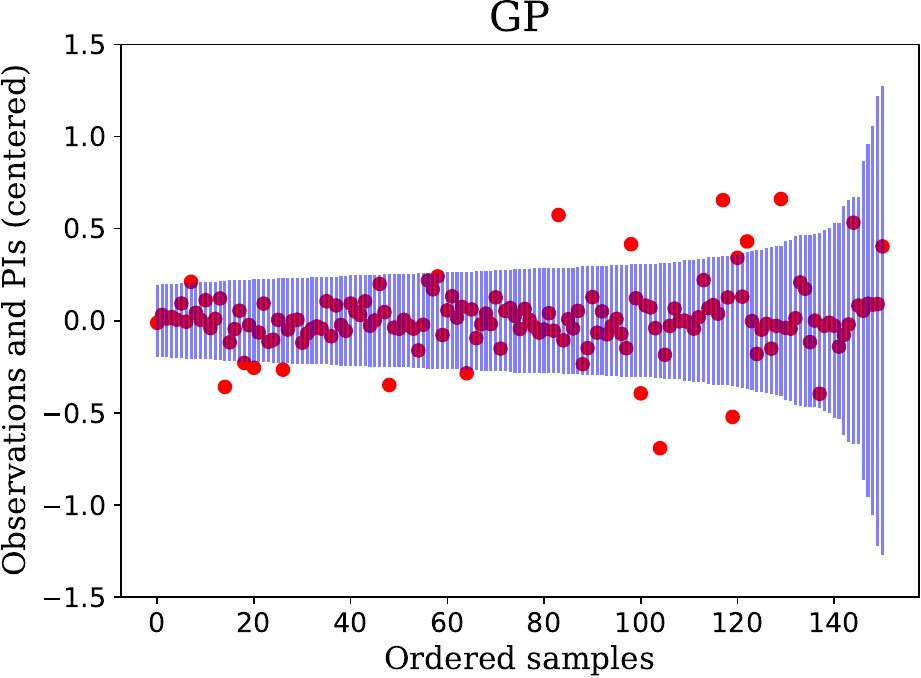}
	\qquad
	\includegraphics[width=.4\textwidth]{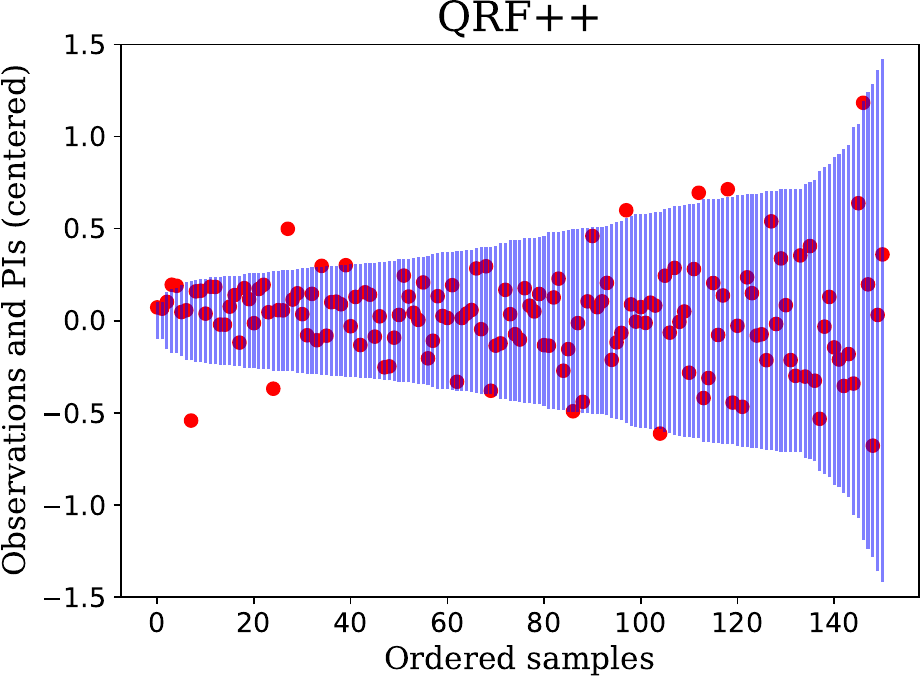}
	\caption{Same as Figure~\ref{fg:concrete_ordered} but for the \texttt{\bf airfoil} dataset.}
	\label{fg:airfoil_ordered}
\end{figure}

Figure~\ref{fg:airfoil_picp} displays the results for the \texttt{\bf airfoil} dataset. The coverage of the GP model shows a strong deviation from the nominal confidence level, while QRF++ is better calibrated. Interestingly, in this case, the PI widths produced by these models differ substantially (rightmost panel), with the GP model yielding sharper PIs. 
One plausible hypothesis for the GP model yielding a narrower MPIW and higher coverage is that the underlying target function is highly smooth, making it inherently well-suited for GP, whereas the piecewise-constant nature of QRF++ might be less compatible with such continuous characteristics. 
Finally, test samples ordered by PI widths are shown in Figure~\ref{fg:airfoil_ordered}.

In summary, QRF++ generally produces reliable PIs, and the deviation between the empirical coverage and the nominal confidence level ($|\text{PICP}(\lambda) - \lambda|$) is quantitatively small, if present. However, we find that the calibration quality and the trends in MPIW are highly dataset-dependent. Pursuing further improvements by employing methods such as conformal prediction \cite{Shafer2008,Fontana2023,Angelopoulos2023} merits future study, although it is beyond the scope of this work.

\section{Proposed Method}\label{sc:mainpart}

In this section, we define and elaborate on Tomographic Quantile Forests (TQF), a regression framework tailored for multivariate uncertainty quantification. The pseudocode is displayed in Algorithms~\ref{alg_1}, \ref{alg_2}, and \ref{alg_3} below. TQF is fully nonparametric, and leverages QRF++ as a backbone to mitigate the curse of dimensionality and suppress the impact of noisy features. Rather than outputting a full distribution directly, TQF employs a post-processing step that constructs an empirical measure of weighted points that approximates the conditional distribution $p(\yy | \xx)$.

Learning the distribution of Radon-transformed projections could be vulnerable to scale differences among different components of $\yy$. For example, a meteorological record $\yy$ might comprise a vector of $($solar radiation, precipitation, wind speed, temperature$)\in \RR^4$, which are measured in entirely different units. To mitigate this problem, we standardize each component of $\yy$ to zero mean and unit variance before training starts (line~\ref{eq:111111} in Algorithm~\ref{alg_1}).

\subsection{Stage I: Model fitting}

The core idea is to consider multiple direction vectors $\nn\in\SSS^{d-1}$ and, for each direction, model the conditional quantiles of the projected target $\nn ^{\top} \yy$ as a function of both the covariates $\xx$ and the direction $\nn$. At test time, the model predicts directional quantiles for multiple sampled directions and then reconstructs a multivariate conditional distribution by optimizing the locations and weights of a point cloud to approximately minimize the sliced Wasserstein distance (\eqref{eq:SW} in Appendix~\ref{sc:disc}) to the true distribution. Given a weighted point cloud that represents a distribution, a broad range of statistical analyses becomes straightforward. For example, one can compute the mean, median, covariance, and expectiles; estimate marginal distributions; and estimate the copula. One can apply kernel density estimation to find a smooth interpolation of the empirical sample density.

\begin{figure}[H]
	\centering
	\begin{algorithm}[H]
		\caption{\quad Tomographic Quantile Forests (TQF)}
		\label{alg_1}
		\begin{algorithmic}[1]
		\REQUIRE
			Training dataset $\D = \{(\xx_n, \yy_n)\}_{n=1}^{N}$, 
			target dimensionality $d= \dim(\yy_n)\in\NN$, 
			sample augmentation factor $G\in\NN$, 
			feature augmentation factor $\wt{G}\in\NN$,
			QRF++ model $\model_Q$ with a parameter $T\in\NN\cup\{0\}$, 
			quantile levels $(q_m)_{m=1}^{M}\in (0,1)^M$, 
			number of projection axes for prediction $K\in\NN$, 
			test sample $\xx_{\rm new}$
		\ENSURE 
			Weighted point cloud $\big\{ (w_j, \yy_j) \big\}_{j=1}^{J}$
			representing the conditional distribution 
			$p(\yy | \xx_{\rm new})$
		\vspace{2mm}
	   \item[] \hrule 
	   \item[] {\color{gray}\# Training}
		\STATE Normalize the target data:\label{eq:111111}
	   \[
	   	\wt{\D} \leftarrow \ckakko{\xx_n,\zz_n}_{n=1}^{N}
	   	:= \ckakko{\xx_n, {\texttt{scaler}}\mkakko{\yy_n}}_{n=1}^{N}
	   \]
	   \STATE Replicate each record in $\wt{\D}$ exactly $G$ times:\label{eq:34451f}
	   \[
		   \wt{\D}_{*}\leftarrow 
		   \ckakko{\mkakko{\hat\xx_\ell, \hat\zz_\ell}}_{\ell=1}^{GN}
		   := \Big\{
		   \scalebox{0.9}{$\underbrace{(\xx_1,\zz_1),\cdots,(\xx_1,\zz_1)}_G$}, 
		   \scalebox{0.9}{$\underbrace{(\xx_2,\zz_2),\cdots,(\xx_2,\zz_2)}_G,
		   (\xx_3,\zz_3),\cdots$} \Big\}
		   \quad 
	   \]
	   \STATE Draw unit vectors $\{\nn_\ell\}_{\ell=1}^{GN}$ randomly from $\SSS^{d-1}$
	   \STATE Draw $\wt{G}-1$ orthogonal matrices $O_1,O_2,\cdots,O_{\wt{G}-1}$ randomly from the Haar measure on the orthogonal group $\OOO(d)$
	   \STATE Perform projective augmentation of $\wt{\D}_*$:\label{ln:432wrefds}
	   \newline
	   \hspace*{6em}
	   $\displaystyle 
	   	\wt\D_{**} \leftarrow \left\{
	   		\mkakko{\hat\xx_\ell \,\|\, \nn_{\ell}
	   		\,\|\,O_1 \nn_{\ell}\,\|\cdots\|\,
	   		O_{\wt{G}-1}\nn_{\ell}, ~
	   		\nn_\ell^{\top} \hat\zz_\ell}
	   	\right\}_{\ell=1}^{GN}$
	   	\newline
	   	where $\|$ denotes concatenation of vectors
	   \STATE Train $\model_Q$ on $\wt\D_{**}$
	   \STATE Construct a new model $\model_Q^*$ based on $\model_Q$:\label{ln:23dsf88d}
	   \[
	   	\model_Q^*(\xx, \nn, q) := 
	   	\frac{1}{2}\big[
			\model_Q \big(\xx, \nn, q \big) - 
			\model_Q \big(\xx, -\nn, 1-q \big) \big]
	   \]
	   \item[] {\color{gray}\# Inference for a new predictor $\xx_{\rm new}$}
	   \STATE Draw unit vectors $\{\nn_k\}_{k=1}^{K}$ randomly from $\SSS^{d-1}$
	   \STATE $\D_{\rm slice} \leftarrow \emptyset$
	   \FOR{$k \leftarrow 1, \dots ,K$}
	   \FOR{$m \leftarrow 1, \dots ,M$}
	   		\STATE Predict the quantile as ~~ $\Q_{k,m} \leftarrow 
	   		\model_Q^{*}(\xx_{\rm new},\nn_k,q_m)$
	   		\STATE $\D_{\rm slice}\leftarrow 
	   		\D_{\rm slice} \bigcup \big\{
	       (\nn_k,\,q_m,\, \Q_{k,m})\big\}$ 
	   \ENDFOR
	   \ENDFOR
	   \STATE $\big\{(w_j,\zz_j)\big\}_{j=1}^{J} 
	   \leftarrow \text{QMEM} \mkakko{\D_{\rm slice}}$ 
	   \hfill \COMMENT{Algorithm~\ref{alg_2}}
	   \STATE Undo normalization:
	   $\big\{\yy_j\big\}_{j=1}^{J} 
			\leftarrow \big\{{\texttt{scaler}}^{-1}(\zz_j)\big\}_{j=1}^{J}$
	   \RETURN $\big\{(w_j, \yy_j)\big\}_{j=1}^{J}$
		\end{algorithmic}
	\end{algorithm}
\end{figure}

A naïve implementation would require fitting $K \times M$ separate quantile regressors for each of $K$ directions and $M$ quantile levels in $(0,1)$. That will inflate implementation effort and training time. Moreover, projected values $(\nn ^{\top} \yy)_{q}$ at nearby quantile levels $q$ and nearby directions $\nn$ are strongly correlated (i.e., $(\nn ^{\top} \yy)_{q}\simeq ({\nn'}^{\top} \yy)_{q'}$ if $\nn\simeq \nn'$ and $q\simeq q'$), making independent modeling statistically inefficient. We therefore adopt two design choices. First, we use QRF++ (introduced in Section~\ref{sc:qrfpp}) to learn multiple quantile levels simultaneously. Second, we \emph{embed the direction $\nn$ as part of the input}, concatenating $\xx$ and $\nn$ and feeding the joint vector $\xx \,\|\, \nn$ to the QRF++ model. It is worthwhile to mention an advantage of using QRF/QRF++: the monotonicity of estimated quantiles. Methods that fit independent models at different quantile levels are prone to quantile crossing due to broken monotonicity \cite{He1997,Zhao2000}; in contrast, QRF++ reads all quantiles from one estimated conditional CDF, which ensures monotonicity.

The resulting \emph{projective augmentation} proceeds as follows (lines~\ref{eq:34451f}--\ref{ln:432wrefds} in Algorithm~\ref{alg_1}). For each training pair $(\xx,\yy)$, sample $G>1$ unit vectors $\{\nn_g\}_{g=1}^{G}$ uniformly from the hypersphere $\SSS^{d-1}$. In theory, a concatenated vector $\xx \,\|\, \nn_g$ carries complete information on $\xx$ and $\nn_g$. However, in practice, this na\"{i}ve representation might be suboptimal, leading to RF's excessive partitions or slower convergence. For instance, consider curving out the interval $-1\leq x\leq 1$ from $\RR$. A decision tree would need two splits at $x=-1$ and $x=+1$. However, if $x^2$ is present as an extra feature, then only a single partition at $x^2=1$ will do the job. This way, feature enrichment can substantially help make trees simpler. Motivated by this observation, we propose to enrich the input $\xx\,\|\,\nn_g$ as follows. Let $\wt{G}\geq 1$ and draw $\wt{G}-1$ $d\times d$ orthogonal matrices $O_1, O_2,\cdots, O_{\wt{G}-1}$ randomly from the orthogonal group $\OOO(d)$. Then we rotate $\nn_g$ by $O_1, O_2,\cdots$ and append them as extra features, resulting in the long augmented vector $\xx \,\|\,\nn_g\,\|\,O_1 \nn_g\,\|\,O_2\nn_g\,\|\cdots\|\,O_{\wt{G}-1}\nn_g$. (If $\wt{G}=1$, no feature augmentation takes place.) This process is repeated for each $g=1,\cdots,G$, yielding $G$ augmented training pairs. Summarizing, the projective augmentation is a transformation
\al{
	(\xx, \yy) \mapsto 
	\left\{
		\big(\xx \,\|\, \nn_g \,\|\ O_1 \nn_g \,\| \cdots\|\,O_{\wt{G}-1}\nn_g,\ \nn_g ^{\top} \yy\big)
	\right\}_{g=1}^G\,.
}
Iterating this for all the training pairs $\{(\xx,\yy)\}$ will enlarge the dataset volume by a factor of $G$. Increasing $G$ would make it easier for the model to learn $\nn$-dependence, but the training cost would increase in proportion. Determining the optimal choice of $G$ and $\wt{G}$ on general grounds is difficult; we believe it would be sensitive to the dataset size and characteristics, the dimensionality of $\yy$, and the QRF++ hyperparameters. In Section~\ref{sc:45gfcqp} we will examine this numerically.

Because projections onto $\nn$ and $-\nn$ differ only by sign, one could restrict the sampling of directions to a hemisphere without loss of generality. However, it is known that decision-tree-based predictors are prone to boundary (``edge'') effects, with predictions flattening near the edge of the covariate domain~\cite{Athey2019}. To mitigate such boundary effects in QRF++, we sample directions over the entire sphere $\SSS^{d-1}$. 

By the definitions of projection and quantiles, one has $-(\nn ^{\top} \yy)_{q}=\big((-\nn) ^{\top} \yy\big)_{1-q}$. The predictions by QRF++, however, respect this symmetry only approximately. To enforce it, we symmetrize the predictions by taking the average of the two “partner” predictions (line \ref{ln:23dsf88d} in Algorithm~\ref{alg_1}). This symmetrization ensures that the symmetry holds exactly by construction.

Our approach outlined above is partly inspired by prior work on offline reinforcement learning, specifically \emph{Hindsight Experience Replay}~\cite{Andrychowicz2017}, where training of agents on sparse rewards is facilitated by augmenting state transition samples from a replay buffer with multiple goals and corresponding rewards, thereby reusing the same experience for diverse targets without additional environment interactions. It is also noteworthy that the method of constructing linear combinations of existing multiple targets has been proposed before \cite{Tsoumakas2014,Yamaguchi2024}, but their interest was not in uncertainty quantification.

\paragraph{\bf Computational Complexity} 
It is important to explicitly note the computational overhead introduced by the projective augmentation and target expansion in TQF. As detailed by Louppe \cite{Louppe2014}, training a standard random forest with $n_{trees}$ trees requires an average time complexity of roughly $\mathcal{O}(n_{trees} \cdot p \cdot N \log^2 N)$ for $N$ samples and $p$ features. In TQF, the effective sample size is multiplied by $G$, the number of features increases to $p + d\wt{G}$, and the multi-output split criterion in QRF++ evaluates $2T+1$ targets simultaneously. Assuming $N \gg G$, the theoretical average time complexity increases by a multiplicative factor of approximately $G(1 + d\wt{G}/p)(2T+1)$ compared to a standard univariate random forest. It is worth noting that this worst-case overhead can sometimes be partially offset in practice; as observed in Section~\ref{sc:yfdlqs}, expanding the target dimension may lead to stronger early partitions and shallower trees, effectively reducing the total number of node splits. Furthermore, because individual trees are built independently, the training process is embarrassingly parallel, allowing the actual runtime to be significantly reduced by leveraging multi-core processors. In any case, while TQF introduces a noticeable constant-factor overhead in memory and time, its overall asymptotic scaling with respect to the sample size $N$ remains $\mathcal{O}(N \log^2 N)$, ensuring that the method remains computationally tractable for typical tabular datasets.

\subsection{Stage II: Distribution reconstruction}

Once the QRF++ model is trained, we use their predictions at a new input $\xx_{\rm new}$ to assemble a weighted point cloud that approximates the conditional distribution $p(\yy | \xx_{\rm new})$. The procedure is outlined in Algorithms~\ref{alg_2} and \ref{alg_3}; we refer to it as the \emph{Quantile-Matching Empirical Measure (QMEM)}. We choose the point locations and weights by minimizing a loss that compares the model-predicted directional quantiles with those induced by the point cloud itself. Let $\Z$ denote a weighted point cloud and write its directional quantile as $\Q(\nn, q;\, \Z)$. If we had access to the true distribution $p$, the sliced 1-Wasserstein distance between $\Z$ and $p$ would be (cf.~\eqref{eq:SW} in Appendix~\ref{sc:disc})
\al{
	{\SW}_1(\Z, p) & = \int_{\SSS^{d-1}}\!\!\!\rmd \nn\;
	\W_1\big(\wh{R}_{\nn\sharp}\Z, \wh{R}_{\nn\sharp} p \big)
	\\
	& = \int_{\SSS^{d-1}}\!\!\!\rmd \nn \int_0^1 \rmd q ~
	\Big| \Q(\nn, q;\, \Z) - F^{-1}_{\wh{R}_{\nn\sharp} p}
	(q) \Big|
	\\
	& \simeq \frac{1}{K}\sum_{\nn_k} \frac{1}{M}\sum_{q_m}
	\Big| \Q(\nn_k, q_m;\, \Z) - F^{-1}_{\wh{R}_{\nn_k\sharp} p}
	(q_m) \Big|
	\\
	& \simeq \frac{1}{K}\sum_{\nn_k} \frac{1}{M}\sum_{q_m}
	\big| \Q(\nn_k, q_m;\, \Z) - \Q_{k,m} \big|\,.
}
In the last step, the inverse cumulative distribution function was replaced by the QRF++ predictions $\Q_{k,m}$. We adopt this discretized sliced Wasserstein objective as our loss and minimize it with respect to $\Z$. $\Q$ requires evaluating a sample quantile function, for which multiple conventions exist~\cite{Hyndman1996}. The choice is particularly consequential when the sample size is small. We adopt Hazen’s formula (Type~5 in Ref.~\cite{Hyndman1996}), which is the only definition that satisfies all desiderata listed therein.

Point-cloud optimization has been used to compute Wasserstein barycenters~\cite{Rabin2011,Bonneel2015,Claici2018}, where one seeks a distribution minimizing a weighted sum of squared Wasserstein distances to several input distributions. Our setting differs in that barycenter problems assume sample access to each input distribution, while here we only observe directional-quantile summaries $\Q_{k,m}$ of the target conditional distribution.

The point cloud $\Z=\{(w_j,\zz_j)\}_{j}$ consists of nonnegative weights $w_j$ and points $\zz_j$, subject to the normalization $\sum_{j} w_j=1$. As noted in \eqref{eq:64trfg} in Appendix~\ref{sc:disc}, the sliced Wasserstein distance $\SW_1$ is convex in its distributional argument; consequently, our loss (after discretizing the integral over $q\in[0,1]$) is \emph{approximately} convex. In particular, optimizing the weights with fixed support points is an approximately convex problem, which can be solved stably and efficiently.

By contrast, the loss viewed as a function of the locations $\{\zz_j\}$ is nonconvex; direct nonlinear optimization is expensive and prone to poor local minima. To mitigate this, we employ an alternating procedure that iterates \emph{point updates} and \emph{weight optimization}. In the point-update step, given the current weighted cloud, we fit a KDE and sample a refreshed set of support points from it. In the weight-optimization step, we reoptimize the weights for the new support so as to minimize the loss. Repeating these two steps drives the loss to a plateau.

To further improve robustness, we form an \emph{ensemble} of $E$ point clouds: we run $E$ parallel instances of the above alternation (sampling from the KDE and reoptimizing the weights), yielding $E$ weighted clouds whose losses are, by construction, \emph{approximately equal} (each run is stopped at a similar plateau). We then merge them by pooling supports and aggregating weights. Because the loss is (approximately) convex (cf.~\eqref{eq:64trfg} in Appendix~\ref{sc:disc}), Jensen’s inequality gives
\[
\LLL(\text{merged}) \;\le\; \frac{1}{E}\sum_{e=1}^E \LLL(\Z^{(e)}).
\]
Since the ensemble members have nearly the same loss, the average on the right is essentially that common level; hence the merged cloud achieves a loss that is \emph{no greater} (and in practice often \emph{strictly lower}) than that of any single member. The merged cloud may contain many negligible-weight points; we remove such points in the \emph{pruning} step (Algorithm~\ref{alg_3}). This completes the reconstruction phase.

The reconstruction steps described above is closely related to the cross-entropy method (CEM)~\cite{Rubinstein1999}. CEM is a global, gradient-free optimization scheme, widely used in evolutionary computation, that alternates between (i) generating a population of candidate solutions and evaluating their objective values and (ii) updating the sampling distribution so as to increase the probability of drawing \emph{elite} samples. In our QMEM procedure, the weights \(w_j\) play the role of performance-based importance assigned to candidates, analogous to fitness scores in CEM.

\paragraph{\bf Construction of Quantile Regions}
Although TQF outputs a discrete empirical measure rather than a continuous geometric contour, one can straightforwardly construct highly flexible probability regions from the predictions. By applying KDE to the output point cloud, one can extract the Highest Density Region (HDR) \cite{Hyndman1996HDR} covering a specified probability mass $\tau \in (0, 1)$. Because the TQF point cloud captures complex topologies (including multimodality and holes), the resulting HDRs naturally form nonconvex or even disjoint regions. In this sense, they serve as useful alternatives to standard geometric multivariate quantile regions, practically overcoming the inherent convexity limitation of DQR.

\begin{figure}[H]
\centering
\begin{adjustwidth}{-\extralength}{0cm}
\flushright
\begin{minipage}[t]{80mm}
	\begin{algorithm}[H]
		\caption{\quad Quantile-Matching Empirical Measure (QMEM)}
		\label{alg_2}
		\begin{algorithmic}[1]
		\REQUIRE
			Radon-transformed quantile values 
			$\D_{\rm slice} = \ckakko{\mkakko{\nn_k,q_m, \Q_{k,m}} \mid 
			k=1..K, m=1..M}$, loss function $\LLL(\cdot, \D_{\rm slice})$, 
			initial point cloud size $N_0\in\NN$, 
			regular point cloud size $N_1\in\NN$, ensemble size $E\in\NN$
	    \ENSURE 
	    	Weighted point cloud $\big\{ (w_j, \zz_j) \big\}_{j=1}^{J}$
	    \item[]
	    \STATE $w_j\leftarrow 1/N_0$~~for~~$j\in\{1,\dots,N_0\}$
	    \STATE $\displaystyle \{ \zz_j \}\leftarrow 
	    \underset{\zz'}{\mathrm{argmin}}~
	    \LLL\Big(\big\{(w_j, \zz'_j)\big\}_{j=1}^{N_0}, \D_{\rm slice}\Big)$
	    \STATE $\Z\leftarrow \big\{ (w_j, \zz_j) \big\}_{j=1}^{N_0}$
	    \STATE $\texttt{loss}\leftarrow \LLL\big(\Z, \D_{\rm slice}\big)$
	    \REPEAT
	    \STATE Fit a Gaussian KDE to $\Z$
	    \STATE Draw samples 
	    $\{\zz_j\}_{j=1}^{N_1}$ from the KDE
	    \STATE $\displaystyle \{ w_j \}\leftarrow 
	    \underset{w'}{\mathrm{argmin}}~
	    \LLL\Big(\big\{(w'_j, \zz_j)\big\}_{j=1}^{N_1}, \D_{\rm slice}\Big)$
	    \STATE $\Z \leftarrow \big\{ (w_j, \zz_j) \big\}_{j=1}^{N_1}$
	    \STATE $\texttt{loss} \leftarrow \LLL\big(\Z, \D_{\rm slice}\big)$
	    \UNTIL{\texttt{loss} stops decreasing}
	    \STATE Fit a Gaussian KDE to $\Z$
	    \STATE $\Z_* \leftarrow \emptyset$
	    \FOR{$e \leftarrow 1, \dots, E$}
	    \STATE Draw samples 
	    $\{\zz^*_j\}_{j=1}^{N_1}$ from the KDE
	    \STATE $\displaystyle \{ w^*_j \}\leftarrow 
	    \underset{w'}{\mathrm{argmin}}~
	    \LLL\Big(\big\{(w'_j, \zz^*_j)\big\}_{j=1}^{N_1}, \D_{\rm slice}\Big)$
	    \STATE $\Z^{(e)} \leftarrow 
	    		\big\{ (w^*_j/E, \zz^*_j) \big\}_{j=1}^{N_1}$
	    \STATE $\Z_* \leftarrow \Z_* \bigcup \Z^{(e)}$
	    \ENDFOR
	    \STATE $\Z_* \leftarrow \mathrm{Prune}\mkakko{\Z_*}$
	    \hfill\COMMENT{Algorithm~\ref{alg_3}}
	    \RETURN $\Z_*$
		\end{algorithmic}
	\end{algorithm}
	\end{minipage}
	\quad
	\begin{minipage}[t]{75mm}
	\begin{algorithm}[H]
		\caption{\quad Prune a Point Cloud}
		\label{alg_3}
		\begin{algorithmic}[1]
		\REQUIRE
			Loss function $\LLL(\cdot, \D_{\rm slice})$, 
			weighted point cloud $\big\{ (w_j, \zz_j) \big\}_{j=1}^J$
	    \ENSURE 
	    	Pruned point cloud $\big\{ (w'_j,\zz_j) \big\}$
	    \item[]
	    \STATE Sort the points in descending order by weight and relabel the indices so that $w_1\geq w_2\geq\cdots\geq w_J$%
	    \FOR{$\wt{j}\leftarrow 1,\dots, J$}
	    	\STATE $C \leftarrow \sum_{j=1}^{\wt{j}} w_{j}$
	    	\FOR{$j\leftarrow 1,\dots,\wt{j}$}
	    	\STATE $w'_j\leftarrow w_j/C$
	    	\ENDFOR{}
	    	\STATE $L_{\wt{j}}\leftarrow 
	    	\LLL\Big(\big\{(w'_j, \zz_j)\big\}_{j=1}^{\wt{j}},
	    	\D_{\rm slice}\Big)$
	    \ENDFOR
	    \STATE $j^* \leftarrow 
	    \mathrm{argmin}_{\wt{j}} \;\, L_{\wt{j}}$
	    \STATE $C\leftarrow \sum_{j=1}^{j^*} w_j$
	    \FOR{$j=1,\dots,j^*$}
	    	\STATE $w'_j \leftarrow w_j / C$
	    \ENDFOR
	    \RETURN $\big\{ (w'_j, \zz_j) \big\}_{j=1}^{j^*}$
		\end{algorithmic}
	\end{algorithm}
\end{minipage}
\end{adjustwidth}
\end{figure}

\section{Empirical Evaluation of the QMEM Algorithm}\label{sc:qmem}

We assess the reconstruction fidelity of QMEM by comparing the reconstructed conditional distributions with the ground truth on synthetic datasets. The discrepancy is measured by ED (\eqref{eq:ed} in Appendix~\ref{sc:disc}). The weight optimization for support points uses the SLSQP algorithm as implemented in the \texttt{scipy.optimize} package of Scipy \cite{scipy}.

\begin{figure}[H]
	\centering
	\includegraphics[width=.9\textwidth]{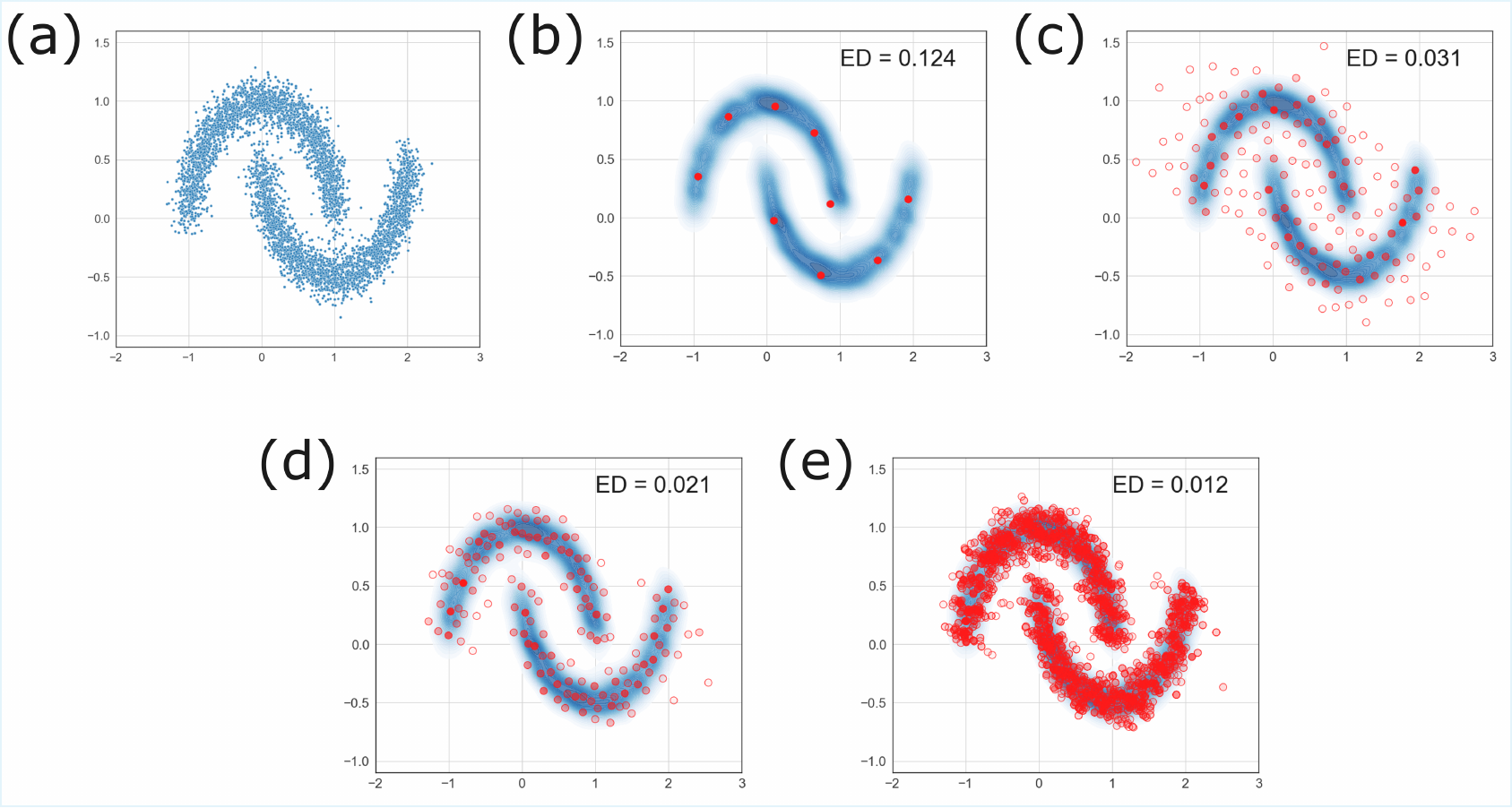}
	\caption{Numerical experiment of the QMEM algorithm. (a) The ``two moons'' dataset. (b) Best fit of 9 points to the data. (c) 150 points randomly sampled from the KDE on the support points in (b), with optimized weights; marker color indicates weight. (d) After a few iterations, the distribution gradually converges toward the true distribution. (e) Final point cloud obtained by pooling 20 clouds. After pruning, the population size is reduced from 3{,}000 to 2{,}456. The score ``ED'' in (b)--(e) stands for the Energy Distance measuring the discrepancy between the true and estimated distributions.}\label{fg:23443lk}
\end{figure}

In the first experiment, we use the 2D ``two moons'' dataset from scikit-learn~\cite{scikit-learn}, comprising 5{,}000 samples with Gaussian noise of standard deviation \(0.1\). The dataset is shown in Figure~\ref{fg:23443lk}(a). We apply QMEM with parameters \(K=25\), \(M=25\), \(N_0=9\), \(N_1=150\), and \(E=20\).%
\footnote{Modern multi-slice CT scanners acquire on the order of \(10^{7}\)–\(10^{8}\) measurement points per rotation \cite{Goldman2007}. By contrast, our experiments use only \(25\times 25=625\) points, i.e., a data volume smaller by \(10^{4}\)–\(10^{5}\) times. Consequently, reconstruction methods that are well suited to clinical CT (with extremely dense measurements) are not appropriate in our markedly sparse setting.} 
The computation took about 40 seconds on our machine (Intel Core i7-1165G7, 2.80\,GHz, with 16\,GB of RAM). No parallelization was used. 
The reconstructed points are overlaid on a smoothed rendering of the true distribution in Figure~\ref{fg:23443lk}(b)--(e), in the algorithmic order of progression; each panel reports the ED score (lower is better). The initial fit in (b) is poor, but accuracy improves as the iterations proceed, and the support points gradually concentrate around the support of the true distribution. Finally, merging the 20 clouds with pruning in (e) yields a marked improvement in accuracy, as is evident in the figure. The multimodal structure of the two moons is captured precisely. 

What is the oracle value range for the ED score? To estimate this, we drew 2{,}456 points with uniform weights from the ground-truth two-moons distribution, repeated over 20 random seeds, and computed the ED between each point set and the main dataset in Figure~\ref{fg:23443lk}(a). The average ED across trials was \(0.0103\) with standard deviation $7.2\times 10^{-4}$, reflecting the finite-sample bias of ED, which vanishes as the sample size tends to infinity. This indicates that the ED $0.0118$ achieved in Figure~\ref{fg:23443lk}(e) is close to the optimal value for this dataset.

\begin{figure}[H]
  \centering
  \begin{minipage}{.45\textwidth}
    \centering
    \includegraphics[width=.85\linewidth]{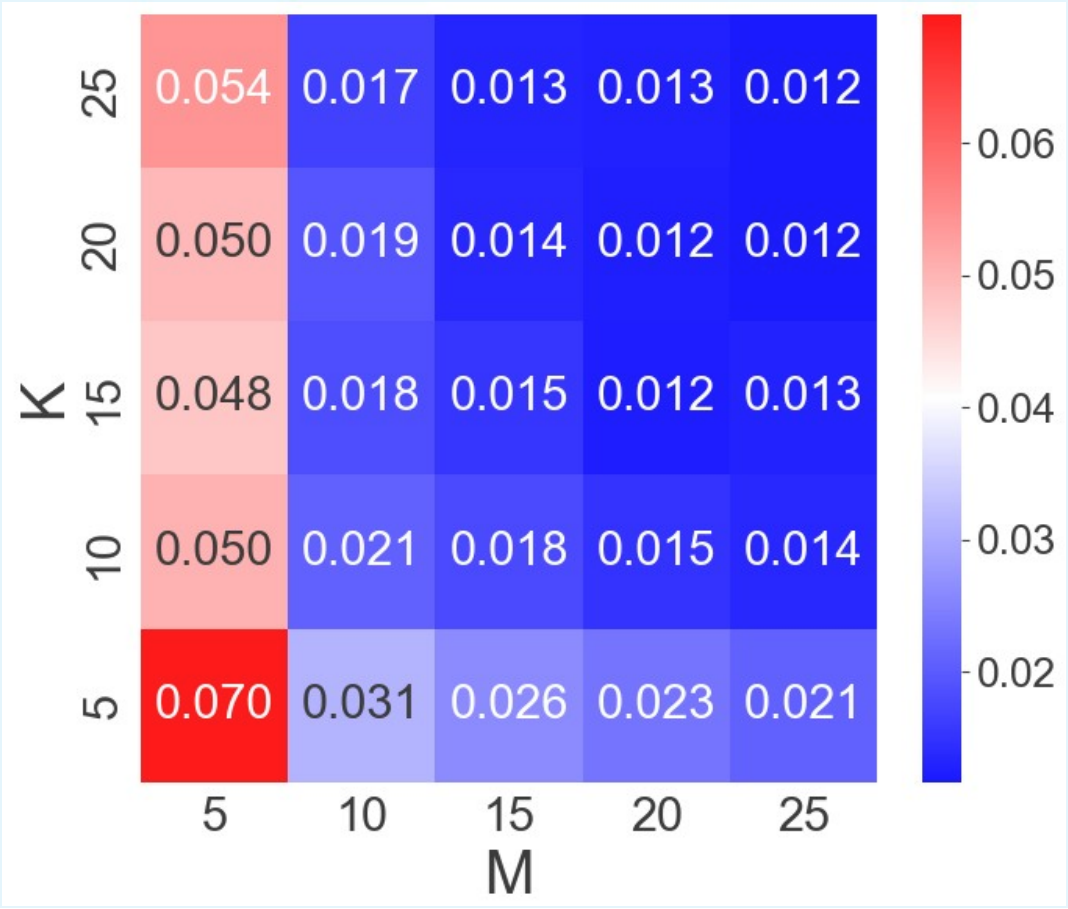}
    \caption{Reconstruction accuracy (ED score) for varying \(K\) and \(M\). Each score is the average over 3 trials with different random seeds.}\label{fg:km}
  \end{minipage}
  \qquad 
  \begin{minipage}{.4\textwidth}
    \centering
    \includegraphics[width=.9\linewidth]{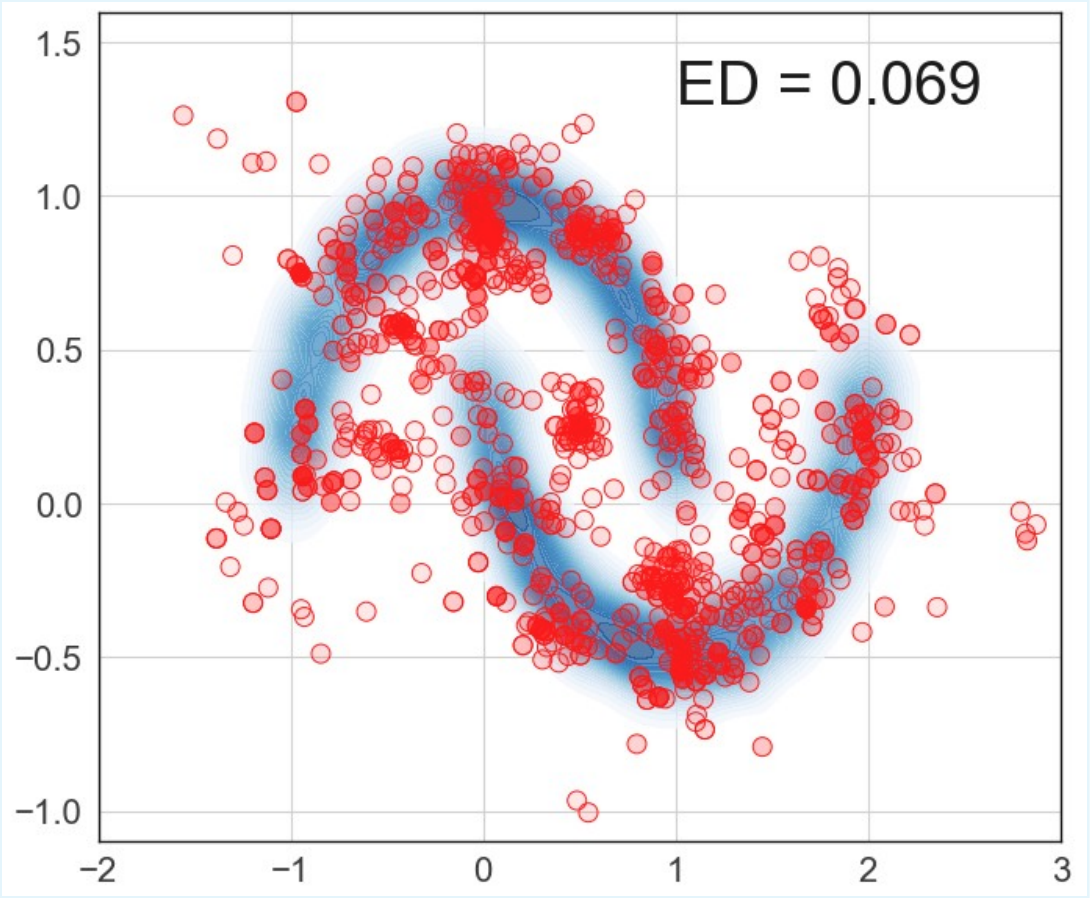}
    \caption{Reconstructed point cloud with \(K=M=5\). After pruning, the population size is reduced from 3,000 to 1,051.}\label{fg:55}
  \end{minipage}
\end{figure}
\begin{figure}[H]
  \centering
  \includegraphics[width=.45\textwidth]{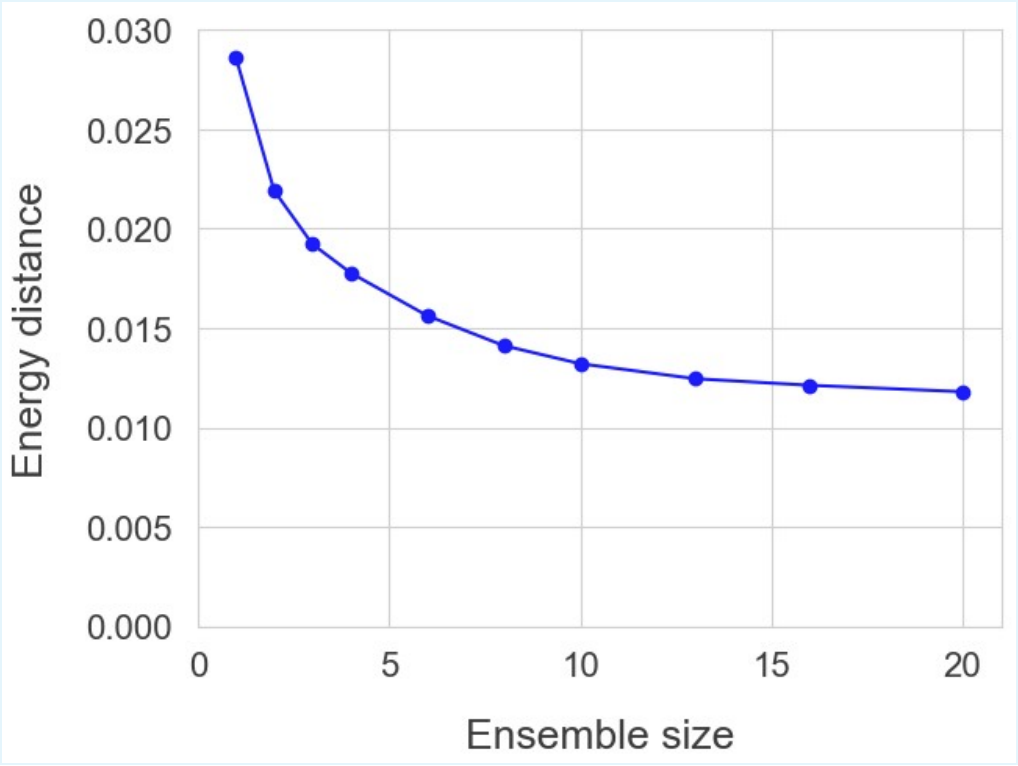}
  \caption{Dependence of reconstruction quality on the ensemble size \(E\). Plotted values are averages over 3 trials; the standard deviations (not shown) are comparable to the marker size.}\label{fg:ens}
\end{figure}

\paragraph{\bf Hyperparameter Dependence}

To examine sensitivity to the hyperparameters, we varied \(K\) and \(M\) and measured performance. (Recall that \(K\) is the number of projection axes and \(M\) is the number of quantiles to match.) As shown in Figure~\ref{fg:km}, smaller \(K\) and/or \(M\) degrade reconstruction quality, as expected. In particular, the impact of a small \(M\) appears stronger than that of a small \(K\). Figure~\ref{fg:55} shows the reconstruction for \(K=M=5\); fine-scale structure in the original dataset is missed, and ED is larger by a factor of \(5.7\) relative to Figure~\ref{fg:23443lk}(e). The dependence on the number of pooled clouds is reported in Figure~\ref{fg:ens}: larger \(E\) yields lower ED (i.e., higher quality) at higher computational cost, with performance appearing to saturate around \(E=15\text{--}20\) on this dataset.

\begin{figure}[H]
  \centering
  \begin{minipage}{.45\textwidth}
    \centering
    \includegraphics[width=\linewidth]{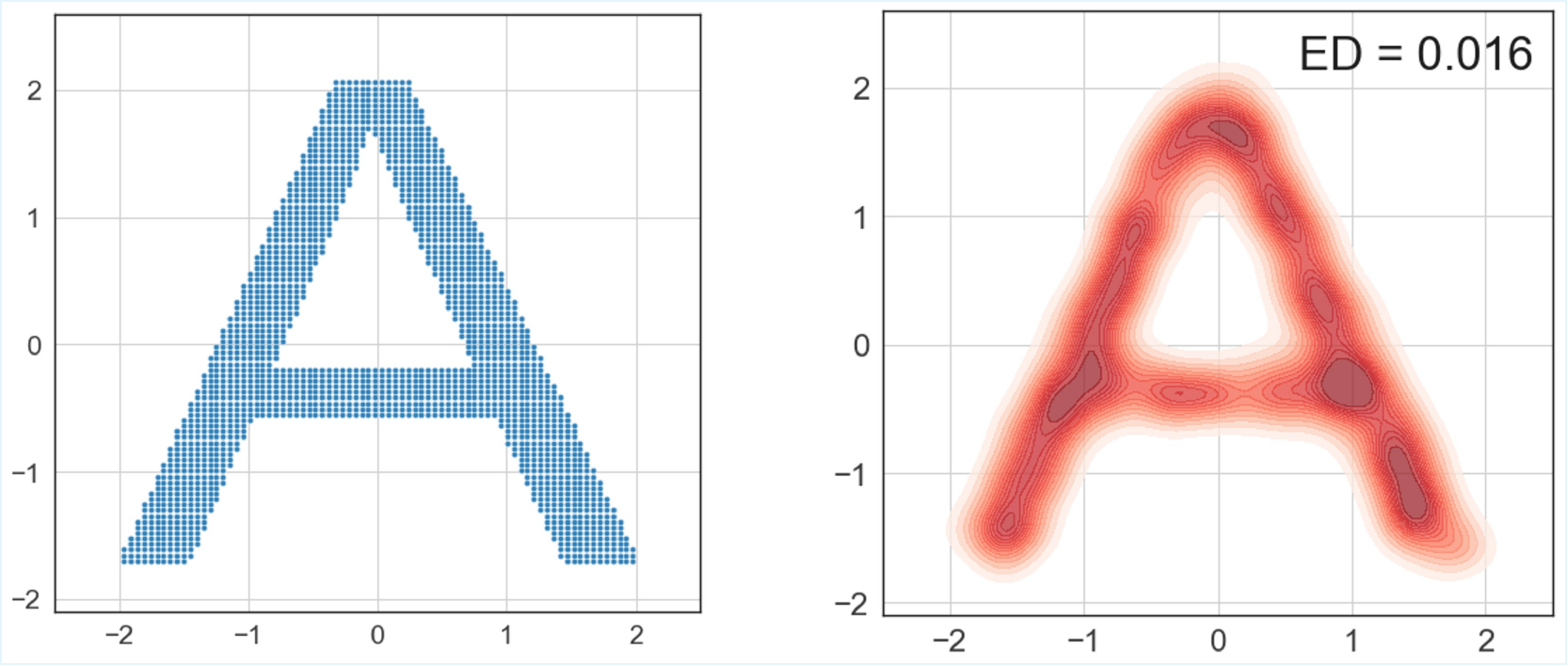}
    \vspace{-5pt}
    \caption{{\bf Left:} letter “A” dataset comprising 2,029 points. {\bf Right:} Distribution reconstructed from 25 projections with QMEM.}\label{fg:A}
  \end{minipage}
  \qquad 
  \begin{minipage}{.45\textwidth}
    \centering
    \includegraphics[width=\linewidth]{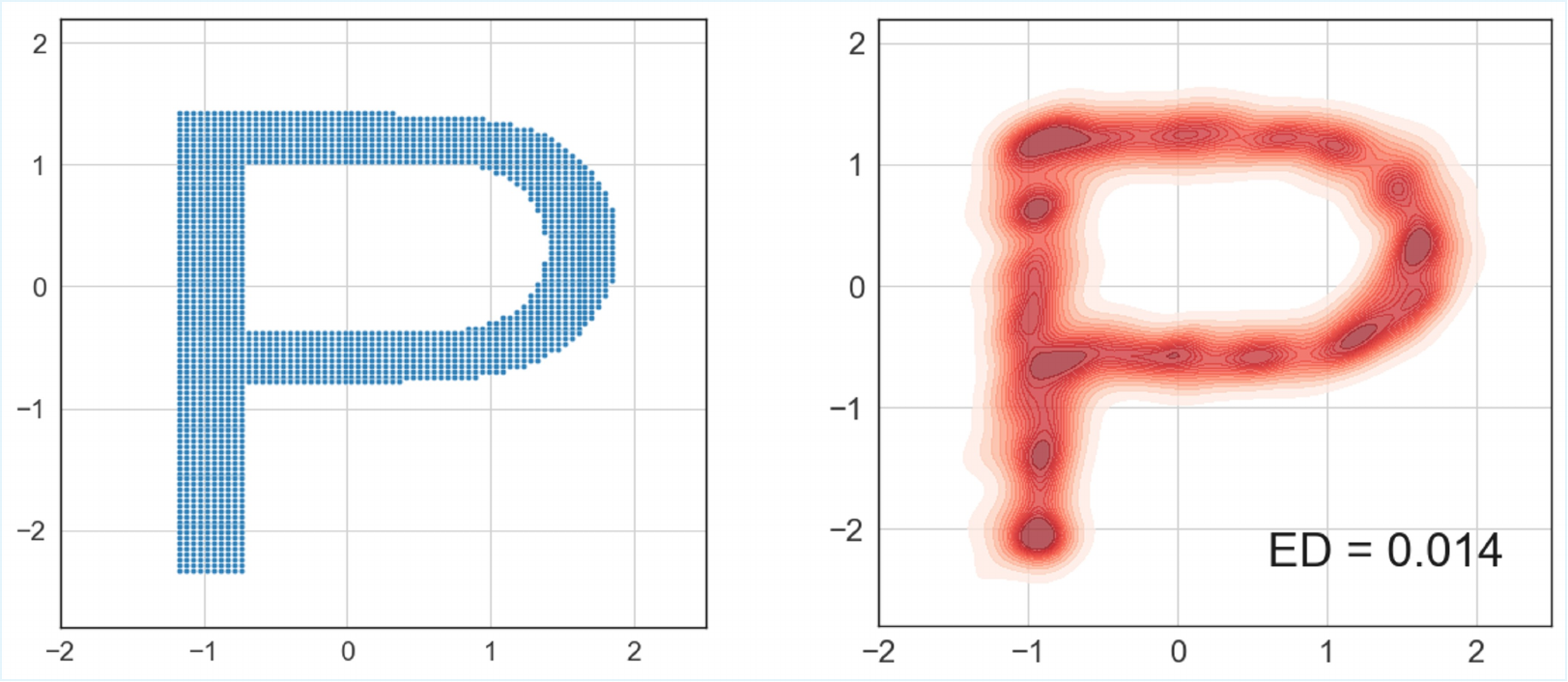}
    \vspace{-5pt}
    \caption{{\bf Left:} letter “P” dataset comprising 2,059 points. {\bf Right:} Distribution reconstructed from 25 projections with QMEM.}\label{fg:P}
  \end{minipage}
\end{figure}

In the second experiment, we used the 2D letter “A” dataset, consisting of 2{,}029 points with each coordinate standardized. The scatter plot is shown in Figure~\ref{fg:A} (left). Using the same parameters as before, we reconstructed the distribution, yielding 2{,}570 support points with optimized weights. A smoothed KDE of the reconstruction is shown in Figure~\ref{fg:A} (right).

In the third experiment, we used the 2D letter “P” dataset, consisting of 2{,}059 points with each coordinate standardized. The scatter plot is shown in Figure~\ref{fg:P} (left). Using the same parameters as before, we reconstructed the distribution, yielding 2{,}785 support points with optimized weights. A smoothed KDE of the reconstruction is shown in Figure~\ref{fg:P} (right). The reconstructions in Figures~\ref{fg:A} and \ref{fg:P} appear to be of good quality.

\section{Experimental Results for the TQF Algorithm}
\label{sc:tqfn}
\subsection{Evaluation on synthetic data I}\label{sc:45gfcqp}

In this subsection we evaluate TQF on a synthetic dataset. 
\paragraph{\bf Dataset}
 
\begin{figure}[H]
	\centering
	\includegraphics[width=.36\textwidth]{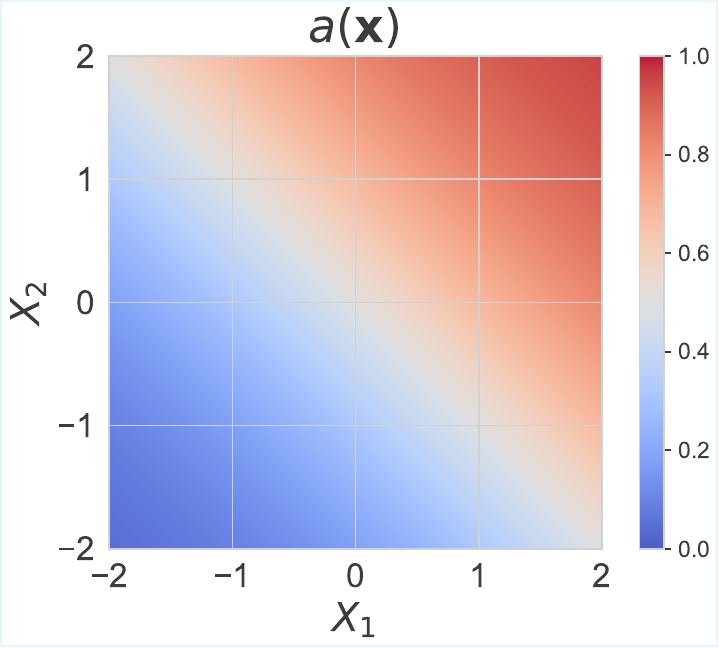}
	\caption{Illustration of the function $a(\xx)$ for $\xx \in [-2, 2]^2$.}\label{fg:a_x_color}
\end{figure}

\begin{figure}[tb]
\begin{adjustwidth}{-0\extralength}{0cm}
	\centering
	\includegraphics[width=.97\textwidth]{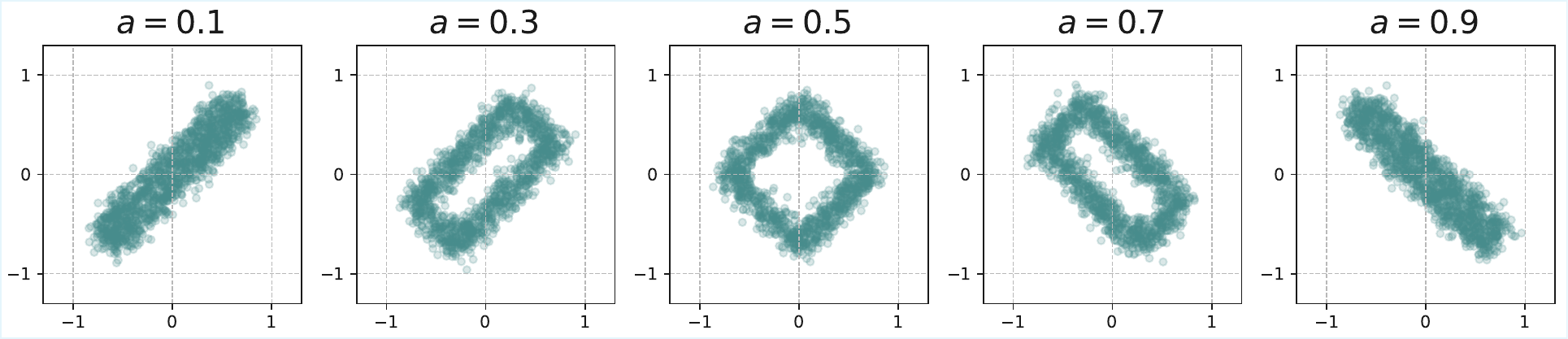}\hspace*{5mm}
	\caption{Conditional distribution $p(\yy\,|\,\xx)$ for varying values of $a(\xx)$ in our synthetic benchmark dataset. Each panel shows 1,500 points.}\label{fg:rdata}
	\vspace{1.2\baselineskip}
	\centering
	\includegraphics[width=.19\textwidth]{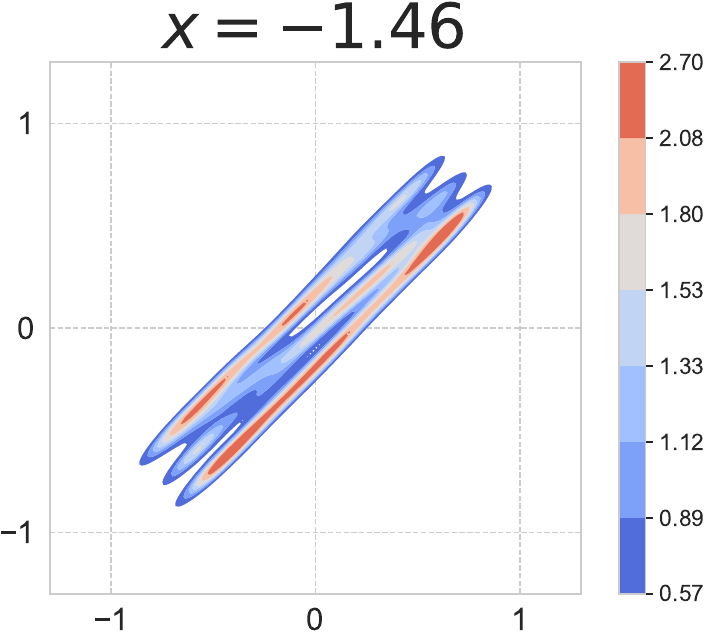}
	\includegraphics[width=.19\textwidth]{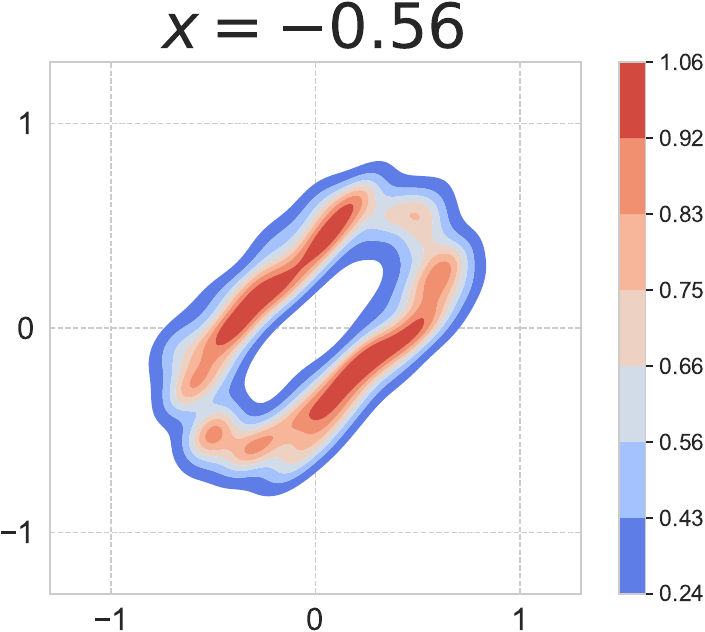}
	\includegraphics[width=.19\textwidth]{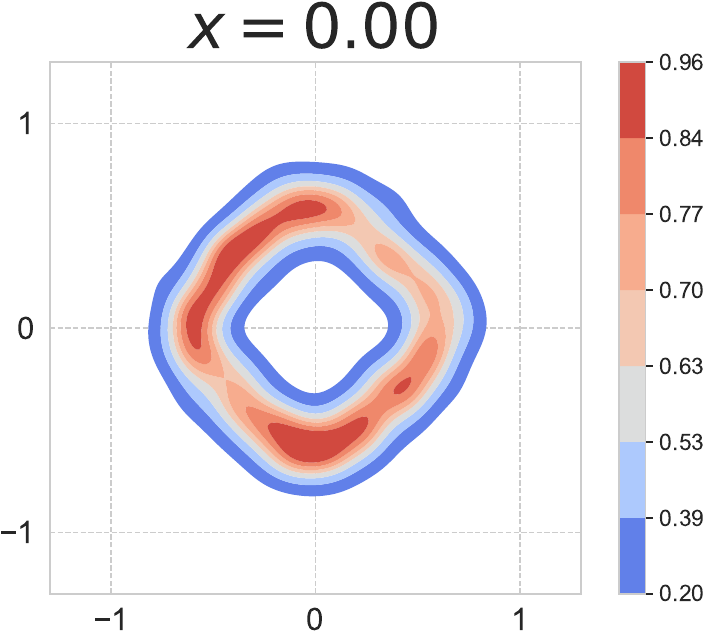}
	\includegraphics[width=.19\textwidth]{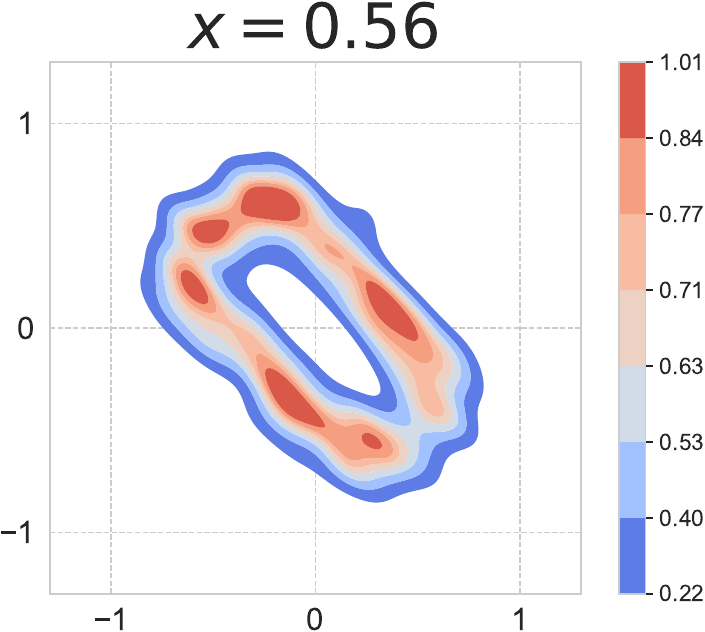}
	\includegraphics[width=.19\textwidth]{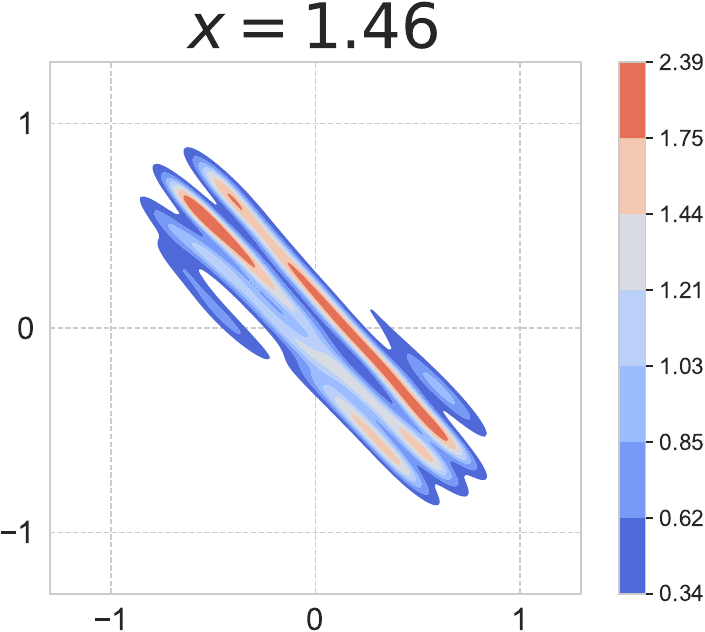}
	\caption{Predictive distributions $p(\yy\,|\, \xx)$ generated by TQF at $\xx=(x,x)$ with $x$ shown in the title of each panel, corresponding to $a(\xx)=0.1, 0.3, 0.5, 0.7$ and $0.9$, respectively, from left to right. Each panel shows KDE of $N$ points with $N=1475,2000, 1995, 1910$, and $1695$.}\label{fg:r_s}
\end{adjustwidth}
\end{figure}

We draw $\xx=(x^{(\ell)})_\ell\in\RR^p$ from a uniform distribution over $[-2,2]^p$, compute the coordinate average $\xx_{\rm avg}:=\frac{1}{p}\sum_{\ell=1}^{p} x^{(\ell)}$, and set $a(\xx):=\sigma(1.5\xx_{\rm avg})\in (0, 1)$, where $\sigma$ denotes the sigmoid function. The function $a$ is illustrated in Figure~\ref{fg:a_x_color}. Conditional on $a=a(\xx)$, the target variable $\yy=(y_1,y_2)$ is supported on a rectangle whose four vertices are $\mkakko{\frac{s_1(1-a)-s_2a}{\sqrt{2}}, \frac{s_1(1-a)+s_2a}{\sqrt{2}}}$ with $s_1,s_2\in\{\pm 1\}$, and is further perturbed by small isotropic noise $\mathcal{N}(\mathbf{0}, 0.08^2 \mathbbm{I}_2)$. Snapshots of $\yy$ are shown in Figure~\ref{fg:rdata}. The distribution is clearly non-Gaussian and highly non-unimodal, making it difficult to reconstruct from the marginals of $y_1$ and $y_2$ alone.
For the experiment, we generate $\{(\xx_i,\yy_i)\}_{i=1}^N$ with $p=2$ and $N=20{,}000$.

\paragraph{\bf Model} 
We configure TQF as follows. We compute all pairwise distances among the scalar target $y$ (i.e., projection $\nn^\top \yy$ of the original targets) and choose the $T$ parameters $\{w_t\}$ in \eqref{eq:comy} as the quantiles of this distance distribution at levels $(2t-1)/(2T)$ for $t=1,\dots,T$. In our setup we use $50$ trees, \verb|min_samples_leaf| $=300$, $T=5$, $G=15$, and $\wt{G}=10$. For QMEM we use $K=M=30$, $N_0=9$, $N_1=100$, and $E=20$. The training and output computation of TQF takes about 2 minutes on our machine.

\paragraph{\bf Evaluation Metric} 
ED is used to quantify the discrepancy between the predicted and the true distributions.

\paragraph{\bf Results}
Numerical results are shown in Figure~\ref{fg:r_s}. We observe that TQF’s distributional predictions closely match the oracle data in Figure~\ref{fg:rdata}. In particular, TQF accurately recovers the holes in the support of the conditional distribution, which are known to be difficult to capture with conventional directional-quantile methods \cite{Hallin2010,Paindaveine2011,Kong2012}.

\begin{figure}[tb]
	\vspace*{4mm}
	\centering
	\raisebox{11pt}{\includegraphics[height=.33\textwidth]{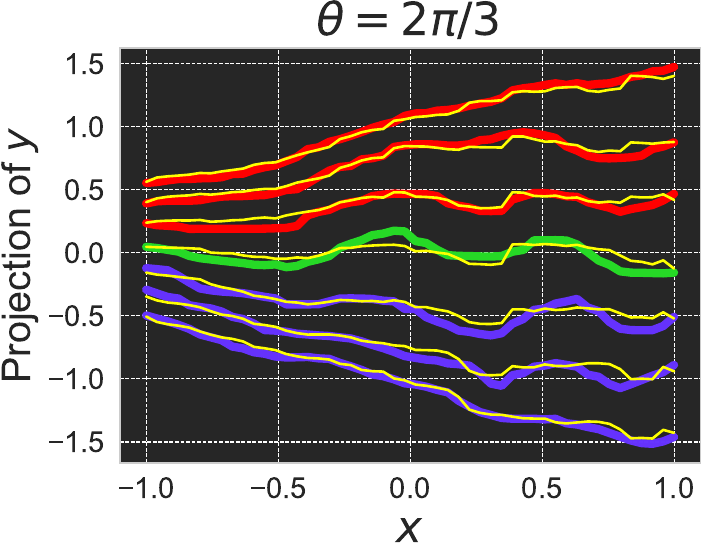}}
	\quad 
	\includegraphics[height=.35\textwidth]{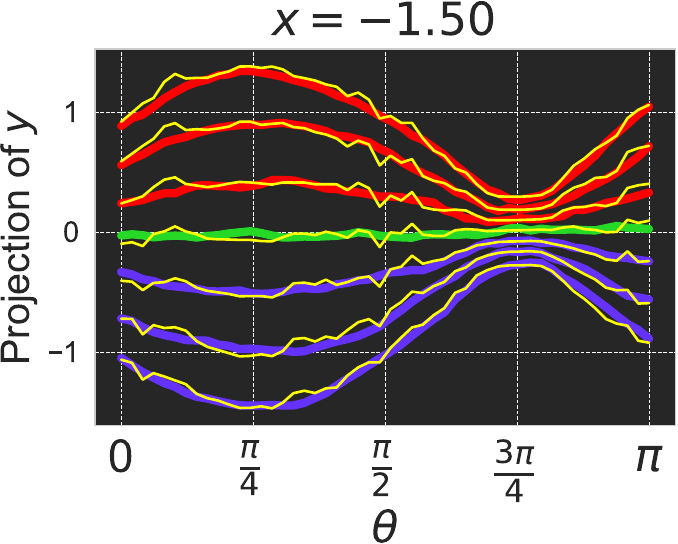}
	\caption{Quantile predictions for $\nn^\top \yy$ obtained by TQF for levels \{0.2, 0.3, 0.4, 0.5, 0.6, 0.7, 0.8\} denoted by thin yellow lines. The seven thick colored curves denote quantile predictions by KNN regression: \{0.2, 0.3, 0.4\} (blue), 0.5 (green), and \{0.6, 0.7, 0.8\} (red), respectively. {\bf Left:} predictions at $\xx=(x,x)$ and $\nn=(\cos \frac{2}{3}\pi, \sin\frac{2}{3}\pi)$. {\bf Right:} predictions at $\xx=(-1.5,-1.5)$ and $\nn=(\cos\theta,\sin\theta)$.}\label{fg:x_th_dep}
\end{figure}

To probe the internal behavior of TQF, we compare its predictions with those of a $k$-nearest-neighbor (KNN) regressor trained on the original dataset (i.e., without projective augmentation). For a test pair $(\xx,\nn)$, KNN assigns weights to training samples based on their distances to $\xx$, projects the corresponding responses via $\yy_i \mapsto \nn^\top \yy_i$, and then computes weighted quantiles of ${\nn^\top \yy_i}$. As shown in Figure~\ref{fg:x_th_dep}, the two methods produce closely matching estimates across varying $\xx$ and $\nn$, providing quantitative evidence that TQF behaves as intended.%
\footnote{Unlike forest-based regressors, KNN regression typically degrades rapidly in high-dimensional settings or in the presence of many uninformative noisy features \cite{Hastiebook}. The present setting is favorable to KNN due to highly informative features; accordingly, KNN may not serve as a reliable baseline in more general tabular problems.}

We further compare TQF with several popular baselines for multivariate distribution estimation; the results are reported in Table~\ref{tb:rrr}. We show only the cases with $a(\xx)\geq 0.5$ because the datasets for $a<0.5$ and $a>0.5$ are mirror images of each other (see Figure~\ref{fg:rdata}). In Table~\ref{tb:rrr}, “Point” denotes a non-probabilistic predictor that outputs only the mean of the ground-truth distribution. “Na\"{i}ve” approximates the joint distribution as the product of marginals, $p(y_1,y_2 |\xx)=p(y_1|\xx)p(y_2|\xx)$, thereby ignoring correlations. “GMM$_k$” fits a mixture of $k$ Gaussians. “Oracle” reports ED computed between two datasets of size 2{,}000 sampled from the ground-truth distribution at the same $\xx$ using different random seeds. We emphasize that the GMM$_1$/GMM$_2$/GMM$_3$ entries achieve \emph{lower} scores than standard Gaussian mixture regression because, in this experiment, the mixtures are fit directly to 2{,}000 samples from the \emph{ground-truth distribution} at fixed $\xx$, and thus incur no regression error.

In Table~\ref{tb:rrr} it is clear that TQF performs significantly better than the others, except for $a(\xx)=0.1$ for which the true distribution is a relatively simple rectangle. It is intriguing that ``Na\"ive'' works better than GMM$_1$ and GMM$_2$ at $a(\xx)=0.5$. Not surprisingly, the score of ``Point'' is always far worse than any other probabilistic method shown here.

\begin{figure}[tb]
\vspace*{0.5\baselineskip}
\begin{adjustwidth}{0\extralength}{0cm}
	\centering
	\includegraphics[width=.195\textwidth]{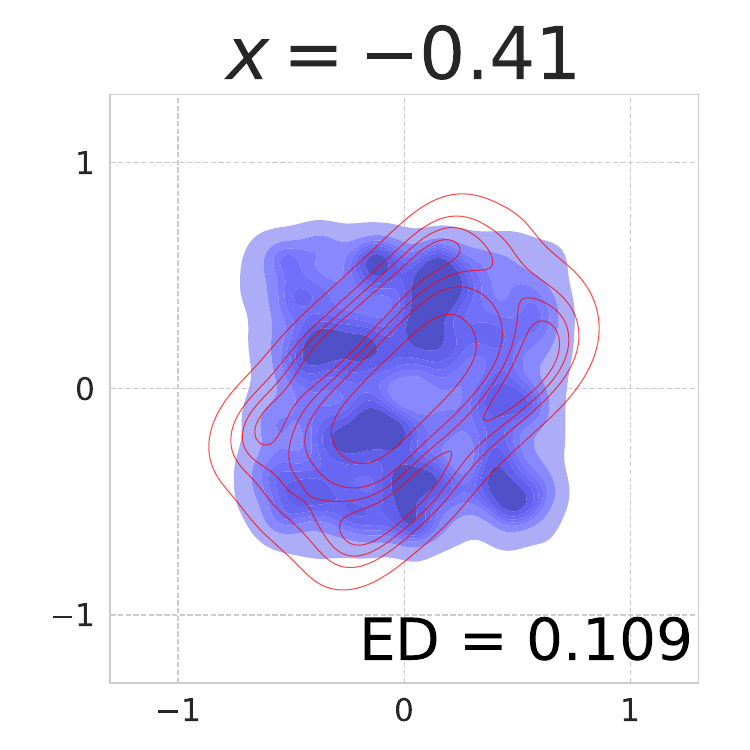}
	\includegraphics[width=.195\textwidth]{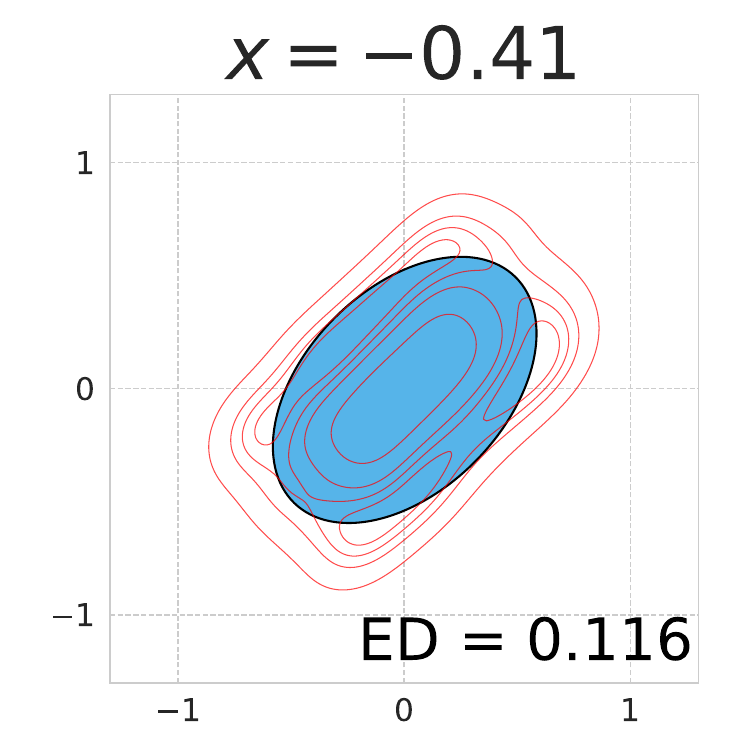}
	\includegraphics[width=.195\textwidth]{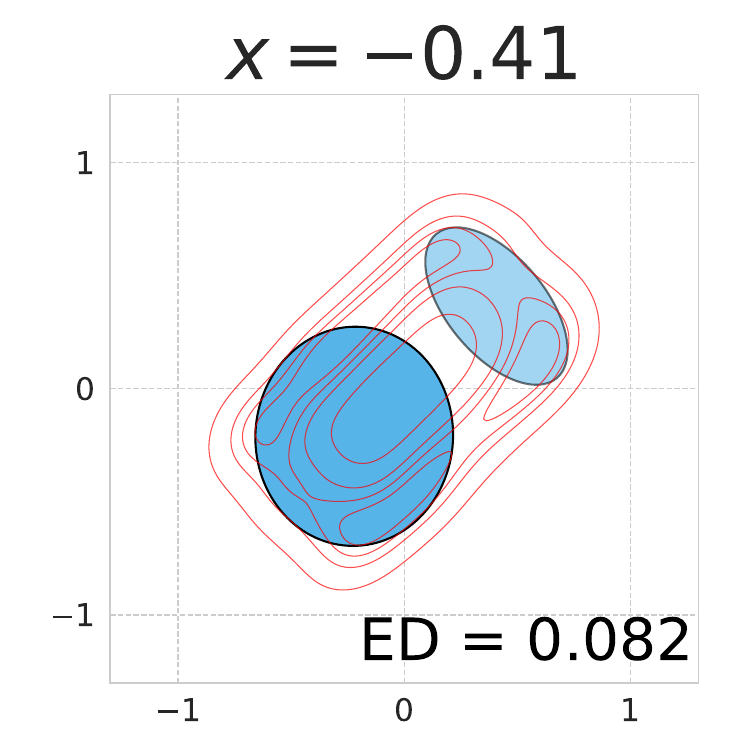}
	\includegraphics[width=.195\textwidth]{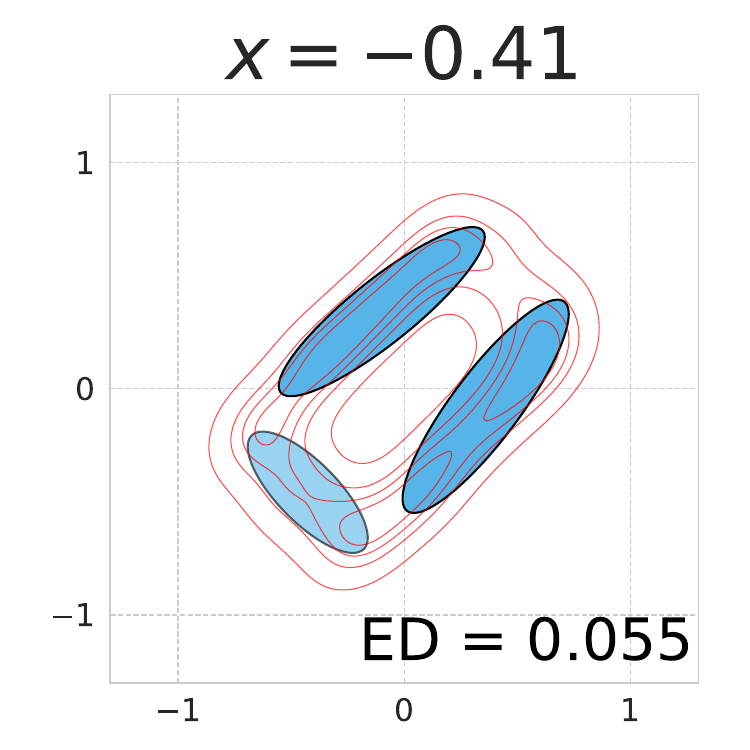}
	\includegraphics[width=.195\textwidth]{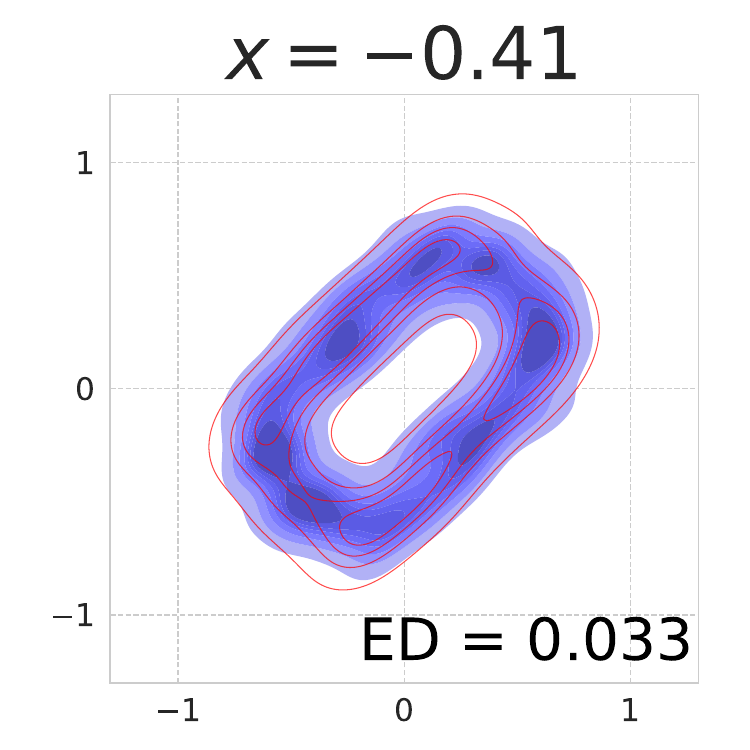}
	\caption{Illustration of the estimated conditional distribution $p(\yy|\xx)$ at $a(\xx)=0.35$ via the five methods from Table~\ref{tb:rrr}. From left to right: Na\"{i}ve, GMM$_1$, GMM$_2$, GMM$_3$, and TQF. The ED score is shown at the bottom right. Filled ellipses for GMM$_{1,2,3}$ indicate individual Gaussian components, covering about 63.2\% of the probability mass; the opacity of the ellipse reflects the relative weight of that component. Red contours represent the ground-truth distribution.}\label{fg:5r}
\end{adjustwidth}
\end{figure}

The actual fitted distributions from all the methods are juxtaposed in Figure~\ref{fg:5r}. We observe that the annulus shape of the true distribution is hard to approximate even with a superposition of multiple Gaussians, while TQF provides an accurate estimation. 

\begin{table}[tb]
\begin{adjustwidth}{0\extralength}{0cm}
	\centering
	\caption{Comparison of methods as measured by ED (lower is better), for the dataset of Figure~\ref{fg:rdata}. Scores are averaged over 10 random seeds for TQF, and 100 random seeds for the rest. Digits in parentheses denote one standard deviation. The best result in each row (except for Oracle), as determined using Welch's $t$ test at the 5\% significance level, is shown in boldface.}\label{tb:rrr}
	\setlength{\tabcolsep}{5pt}
	\begin{tabular}{llllllll}
		\toprule
		$a(\xx)$ & Point & Na\"{i}ve & GMM$_1$ & GMM$_2$ & GMM$_3$ & TQF & Oracle \\
		\midrule
		0.1 & 0.586(5) & 0.205(5) & 0.079(6) & 0.035(5) & {\bf\textsf{0.023(5)}} & 0.038(7) & 0.025(9) \\
		0.2 & 0.601(5) & 0.166(6) & 0.090(7) & 0.051(4) & 0.040(4) & {\bf\textsf{0.033(6)}} & 0.027(8) \\
		0.3 & 0.619(3) & 0.127(6) & 0.106(5) & 0.072(4) & 0.056(5) & {\bf\textsf{0.038(9)}} & 0.025(8) \\
		0.4 & 0.633(2) & 0.098(4) & 0.118(5) & 0.086(4) & 0.055(4) & {\bf\textsf{0.039(6)}} & 0.025(8) \\
		0.5 & 0.638(1) & 0.086(3) & 0.121(4) & 0.094(4) & 0.054(3) & {\bf\textsf{0.043(12)}} & 0.026(7) \\
		\bottomrule
	\end{tabular}
\end{adjustwidth}
\end{table}

\begin{table}[tb]
	\begin{adjustwidth}{0\extralength}{0cm}
	\caption{Hyperparameter dependence of TQF on the dataset of Figure~\ref{fg:rdata}, quantified by ED. Digits in parentheses denote one standard deviation over 9 trials with different random seeds. The best and the second-best results in each row are shown in boldface.}\label{tb:abl}
	\setlength{\tabcolsep}{5pt}
	\centering
	\begin{tabular}{lllllllll}
		\toprule
		& \multicolumn{8}{c}{\((G,\wt G)\)} 
		\\
		\cmidrule(lr){2-9}
		$a(\xx)$
		& (15,1) & (15,5) & (15,10) & (15,15)
		& (1,10) & (5,10) & (10,10) & (20,10) 
		\\
		\midrule
		0.1 & 0.067(23) & 0.042(16)
		& {\bf\textsf{0.036(14)}} & 0.039(18) & 0.050(13) & 0.046(22) 
		& {\bf\textsf{0.037(13)}} & 0.053(26)
		\\
		0.2 & 0.055(10) & 0.037(5)
		& {\bf\textsf{0.035(5)}} & 0.038(4) & 0.046(7) & {\bf\textsf{0.035(6)}} 
		& 0.038(6) & 0.036(5)
		\\
		0.3 & 0.045(11) & 0.046(11)
		& {\bf\textsf{0.040(13)}} & {\bf\textsf{0.043(11)}} & 0.060(16) & {\bf\textsf{0.043(10)}} 
		& {\bf\textsf{0.043(7)}} & 0.045(14)
		\\
		0.4 & 0.054(12) & 0.045(10)
		& {\bf\textsf{0.043(10)}} & 0.045(10) & 0.055(9) & 0.045(8) 
		& 0.045(7) & {\bf\textsf{0.041(7)}}
		\\
		0.5 & 0.061(17) & {\bf\textsf{0.048(17)}}
		& 0.051(22) & 0.051(18) & 0.059(16) & 0.053(16) 
		& {\bf\textsf{0.050(17)}} & 0.051(17)
		\\
		\bottomrule
	\end{tabular}
	\end{adjustwidth}
\end{table}

We systematically investigated the effects of sample augmentation and feature augmentation by varying the hyperparameters $G$ and $\wt{G}$. Varying $G$ changes the effective training-set size for QRF++, and we therefore scale \verb|min_samples_leaf| proportionally with $G$ (e.g., \verb|min_samples_leaf|$=20$ when $G=1$). The experimental results are reported in Table~\ref{tb:abl}. The configurations $(G,\wt{G})=(15,1)$ and $(1,10)$ yield noticeably worse scores than the other settings, suggesting that using both $G>1$ and $\wt{G}>1$ is beneficial for TQF. Moreover, performance appears largely insensitive to the specific values of $G$ and $\wt{G}$ provided they exceed $1$. Overall, Table~\ref{tb:abl} indicates that both sample augmentation (via $G$) and feature augmentation (via $\wt{G}$) are important for TQF to achieve strong performance.

\subsection{Evaluation on synthetic data II}\label{sc:cxovu8}

Next, we test TQF on another synthetic dataset.

\begin{figure}[tb]
	\centering
	\includegraphics[height=.27\textwidth]{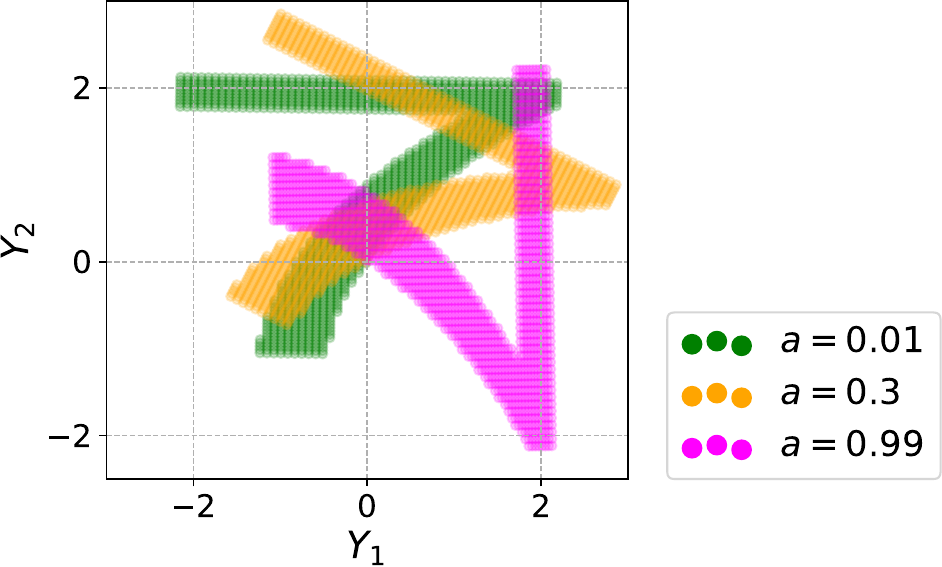}
	\quad 
	\includegraphics[height=.27\textwidth]{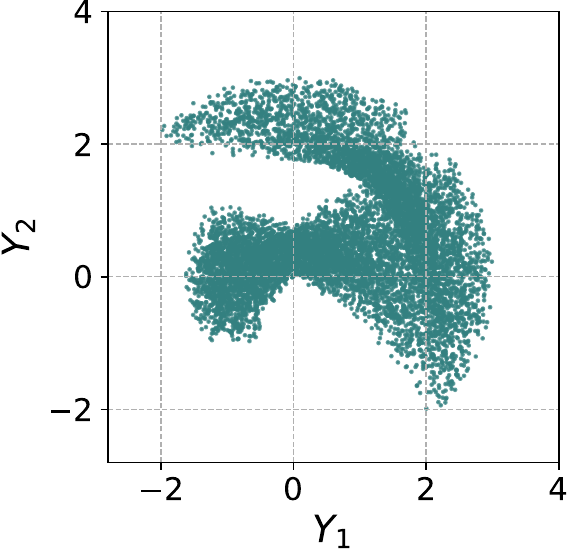}
	\caption{{\bf Left:}~Distribution $p(\yy|\xx)=p\big(\yy | a(\xx) \big)$ for three values of $a(\xx)\in[0,1]$. 
	{\bf Right:}~Scatter plot of 10,000 points generated by $\yy\sim p\big(\yy|\xx\big)$ with $\xx\sim \text{Unif}\mkakko{[-2,2]^5}$.}\label{fg:cdata}
\end{figure}

\paragraph{\bf Dataset} 
We draw $\xx\in\RR^{p_1+p_2}$ uniformly from $[-2,2]^{p_1+p_2}$ and map it to the unit interval via $a(\xx):=\sigma(1.5\xx'_{\rm avg})\in(0,1)$ (cf.~Figure~\ref{fg:a_x_color}), where $\xx'_{\rm avg}:=\frac{1}{p_1}\sum_{\ell=1}^{p_1}x^{(\ell)}$. The remaining components $(x^{(p_1+1)},\ldots,x^{(p_1+p_2)})\in\RR^{p_2}$ are unused and serve as noise features. The conditional distribution $p(\yy|\xx)$ forms a thick “7”-shaped pattern that is rotated about the origin by an angle proportional to $a(\xx)$. Concretely, $\yy\in\RR^2$ is sampled uniformly from a fixed 7-shaped point cloud consisting of 1{,}312 points, and the entire cloud is then rotated clockwise by $\frac{\pi}{2}a(\xx)$ radians. The resulting data are visualized in Figure~\ref{fg:cdata}. In our experiments, we set $(p_1,p_2)=(2,3)$ and generate ${(\xx_i,\yy_i)}_{i=1}^N$ with $N=20{,}000$. Unlike the dataset in Section~\ref{sc:45gfcqp}, whose conditional mean is fixed at $\yy=\mathbf{0}$, the conditional mean in the present dataset varies with $\xx$.

\paragraph{\bf Model}
We configure TQF as follows. We use 120 trees with \verb|min_samples_leaf| $=100$, $G=20$, $\wt{G}=10$, and $T=2$. We found TQF to be sensitive to the choice of the parameters $\{w_t\}$ in \eqref{eq:comy}; after some trial and error, we set $(w_1,w_2)=(0.3,0.4)$. For QMEM, we use the same hyperparameters as in Section~\ref{sc:45gfcqp}. Training and output computation of TQF takes 9 minutes on our machine.

\begin{figure}[tbh]
	\centering
	\includegraphics[height=.14\textheight]{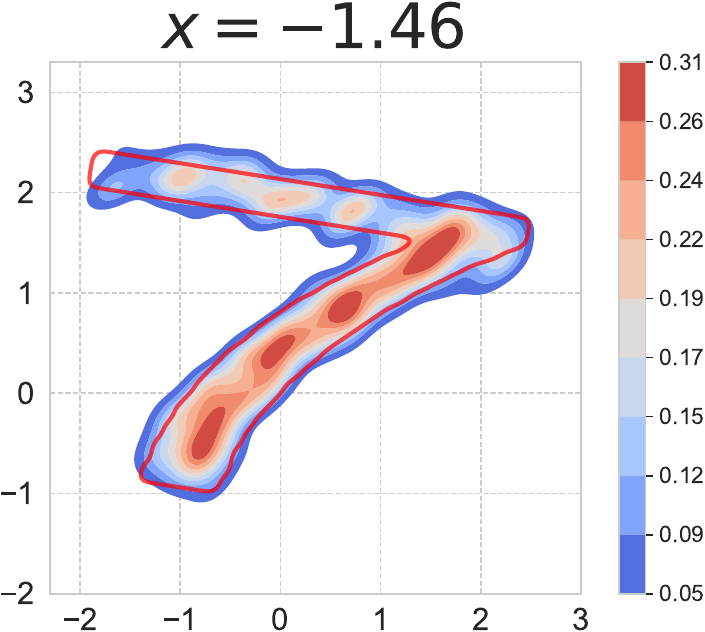} \qquad 
	\includegraphics[height=.14\textheight]{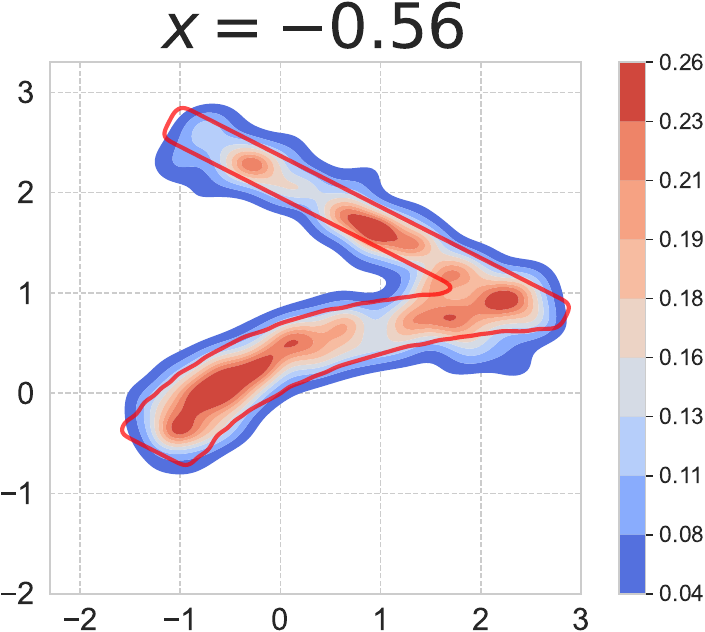} \qquad 
	\includegraphics[height=.14\textheight]{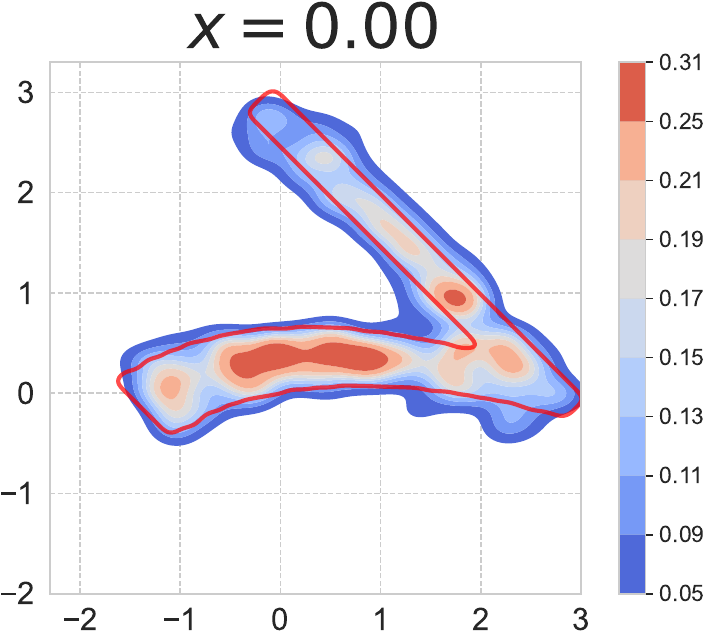}
	\vspace{5mm}
	\\
	\includegraphics[height=.14\textheight]{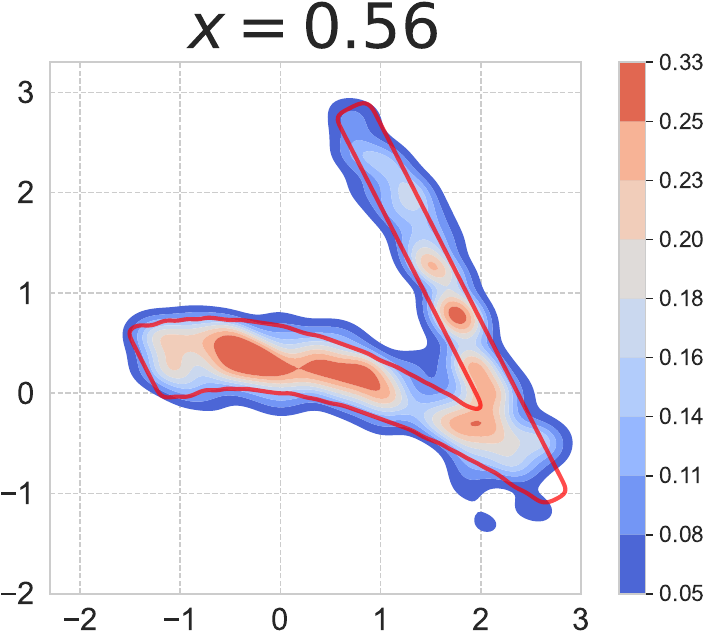} \qquad 
	\includegraphics[height=.14\textheight]{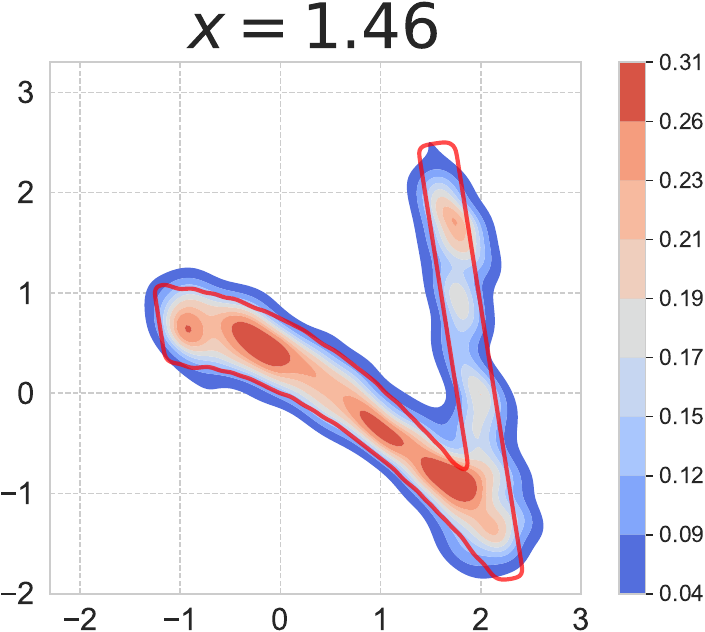}
	\vspace{2mm}
	\caption{Predictive distributions $p(\yy|\xx)$ generated by TQF at $\xx=(x,\cdots,x)\in \RR^5$ with $x$ in the title of each panel, corresponding to $a(\xx)=0.1, 0.3, 0.5, 0.7$ and $0.9$ from top left to bottom right, respectively. Each panel shows a KDE based on $N$ weighted points, with $N = 1740, 1605, 1895, 1805$ and $1745$, respectively. Red contours indicate the boundary of the true support.}\label{fg:c_s}
\end{figure}

\paragraph{\bf Results}
Numerical results are shown in Figure~\ref{fg:c_s}. As $\xx$ varies, even in the presence of noise features, TQF accurately captures the rotating, multimodal support of the conditional distribution; qualitatively, the overlap between the predictions and the ground truth appears satisfactory.

\begin{figure}[t]
	\centering
	\raisebox{4pt}{\includegraphics[height=.22\textheight]{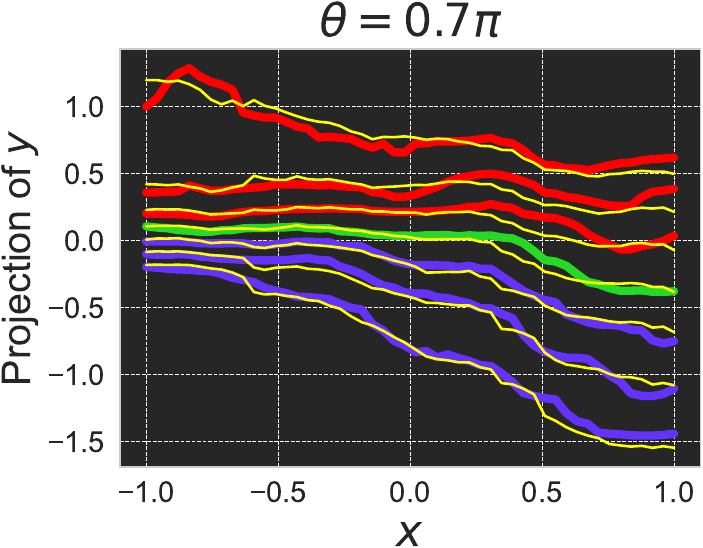}}
	\quad 
	\includegraphics[height=.225\textheight]{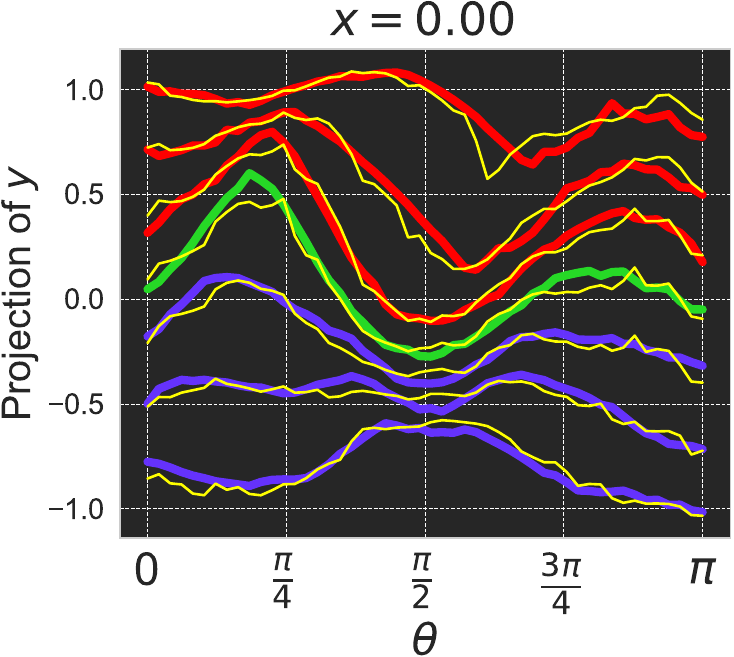}
	\caption{Same as Figure~\ref{fg:x_th_dep} but for the dataset of Figure~\ref{fg:cdata}. {\bf Left:} predictions at $\xx=(x,x,x,x,x)$ and $\nn=(\cos 0.7\pi, \sin 0.7\pi)$. {\bf Right:} predictions at $\xx=(0, 0, 0, 0, 0)$ and $\nn=(\cos\theta, \sin\theta)$.}\label{fg:c_th_dep}
\end{figure}

In Figure~\ref{fg:c_th_dep}, we compare TQF with a KNN quantile regressor. For a fairer comparison, the KNN regressor uses only the first two components of $\xx$ (i.e., the informative features) to compute neighbor distances. We observe good quantitative agreement between the two across a wide range of $x$ and $\theta$.

Table~\ref{tb:ctb} compares TQF with several baseline methods. For the “Oracle” column, we draw two bootstrap samples of size 1{,}312 from the underlying 7-shaped point cloud (with replacement) and report ED between them. We note that the scores for “Point,” “GMM$_k$,” and “Oracle” are independent of $a(\xx)$ because the true distribution is identical up to rotation. Overall, TQF performs comparably to GMM$_3$.

\begin{table}[bht]
	\centering
	\caption{Same as Table~\ref{tb:rrr} but for the dataset of Figure~\ref{fg:cdata}. The best results in each row (except for Oracle), as determined using Welch's $t$ test at the 5\% significance level, are shown in boldface.}\label{tb:ctb}
	\setlength{\tabcolsep}{5pt}
	\begin{tabular}{llllllll}
		\toprule
		$a(\xx)$ & Point & Na\"{i}ve & GMM$_1$ & GMM$_2$ & GMM$_3$ & TQF & Oracle \\
		\midrule
		0.1 & 0.920(8) & 0.208(6) & 0.196(7) & 0.110(25) & {\bf\textsf{0.058(9) }}& {\bf\textsf{0.056(12)}} & 0.035(11) \\
		0.2 & 0.920(8) & 0.207(6) & 0.196(7) & 0.110(25) & {\bf\textsf{0.058(9)}} & {\bf\textsf{0.050(14)}} & 0.035(11) \\
		0.3 & 0.920(8) & 0.192(5) & 0.196(7) & 0.110(25) & {\bf\textsf{0.058(9)}} & {\bf\textsf{0.060(19)}} & 0.035(11) \\
		0.4 & 0.920(8) & 0.164(4) & 0.196(7) & 0.110(25) & {\bf\textsf{0.058(9)}} & {\bf\textsf{0.063(16)}} & 0.035(11) \\
		0.5 & 0.920(8) & 0.136(4) & 0.196(7) & 0.110(25) & {\bf\textsf{0.058(9)}} & {\bf\textsf{0.065(20)}} & 0.035(11) \\
		\bottomrule
	\end{tabular}
\end{table}

\begin{figure}[bht]
	\vspace{.2\baselineskip}
	\centering
	\includegraphics[width=.195\textwidth]{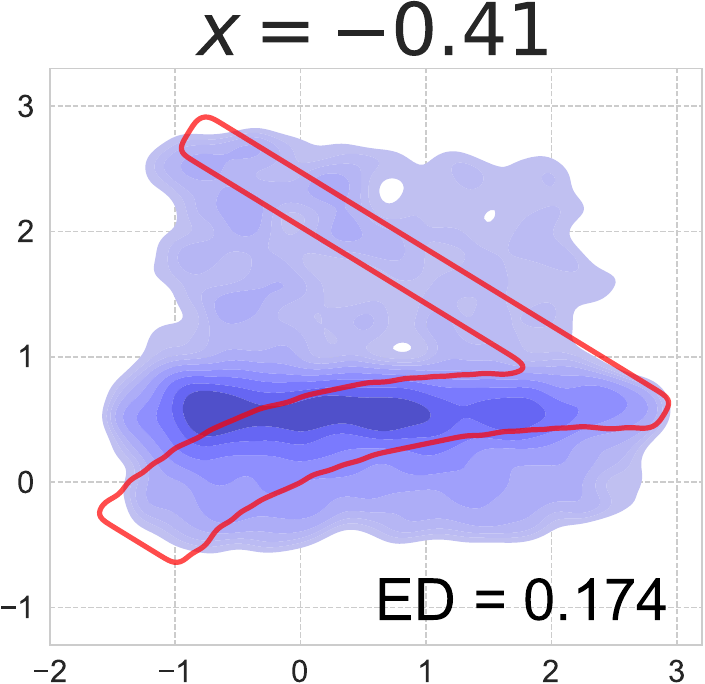}
	\includegraphics[width=.195\textwidth]{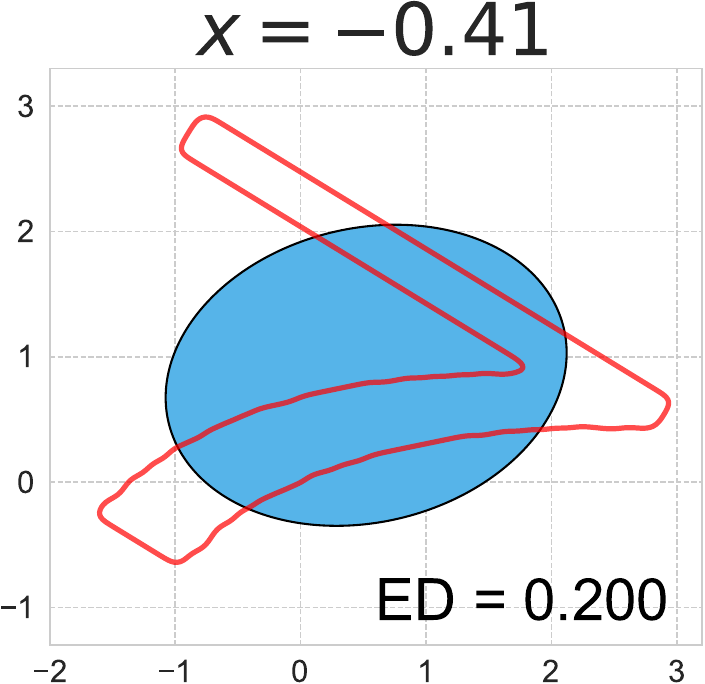}
	\includegraphics[width=.195\textwidth]{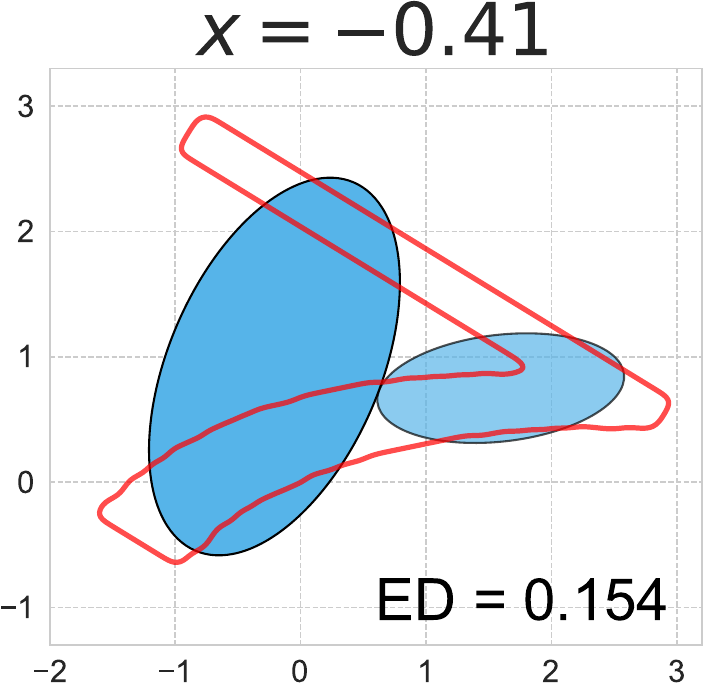}
	\includegraphics[width=.195\textwidth]{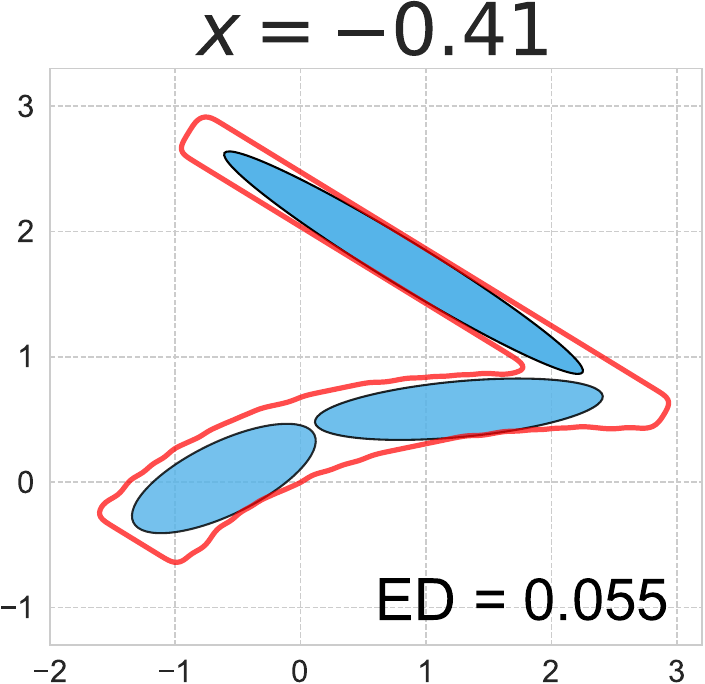}
	\includegraphics[width=.195\textwidth]{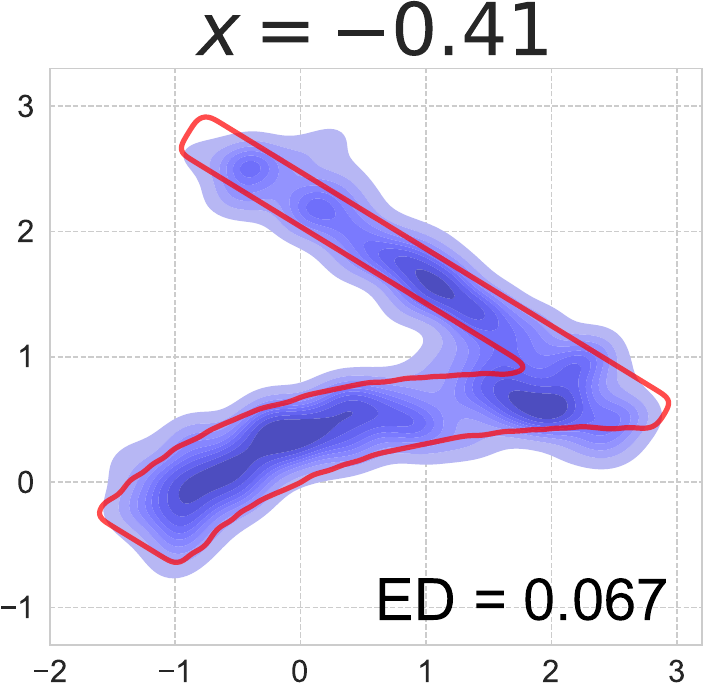}
	\vspace{-0.5\baselineskip}
	\caption{Same as Figure~\ref{fg:5r} but for the dataset of Figure~\ref{fg:cdata}. From left to right: Na\"{i}ve, GMM$_1$, GMM$_2$, GMM$_3$, and TQF. Red contours indicate the boundary of the true support.}\label{fg:5cc}
\end{figure}

Figure~\ref{fg:5cc} visualizes the predictive distributions produced by the probabilistic methods in Table~\ref{tb:ctb}, “Na\"{i}ve,” GMM$_1$, GMM$_2$, GMM$_3$, and TQF (from left to right), together with the corresponding ED score. We see that the nonparametric TQF approach achieves performance comparable to GMM$_3$, which benefits from a carefully specified parametric form and sufficient mixture complexity. In contrast, “Na\"{i}ve” and GMM$_{1,2}$ fail to capture finer geometric features of the true conditional distribution.

\subsection{Comparing TQF and DRF on small data}\label{sc:small}

As both TQF and DRF are tree-based models for distributional prediction, we conduct a numerical experiment in this subsection to highlight their differences. Recall that DRF is consistent under certain conditions \cite[Theorem~2]{Cevid2022}, suggesting that with sufficiently large training data its predictions converge to the optimal (or near-optimal) conditional distribution. From the perspective of model comparison, it is therefore particularly informative to study the scarce-data regime. How do TQF and DRF perform when the training set is extremely small?

\begin{figure}[h]
	\centering
	\includegraphics[width=.3\textwidth]{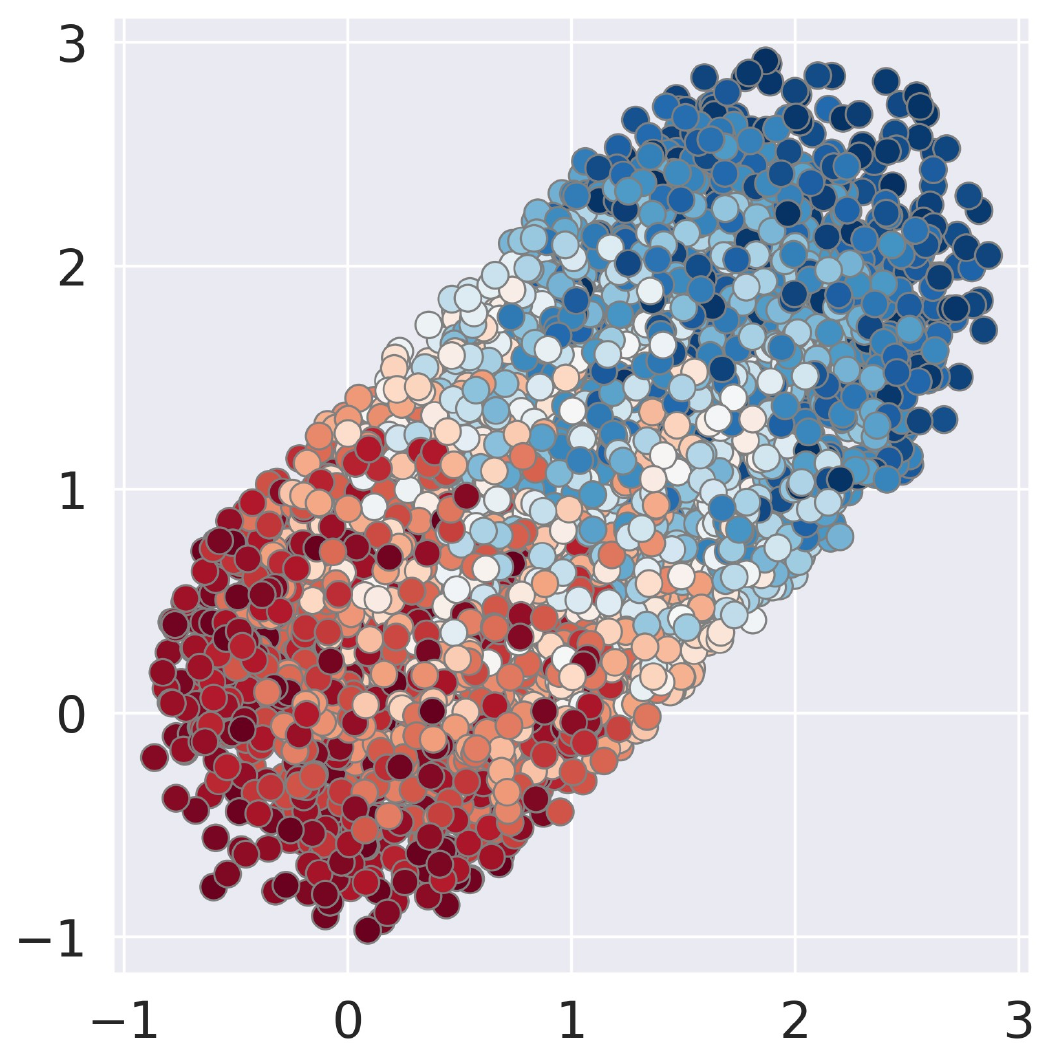}
	\caption{	Target-value distribution for the sliding-disk dataset. Colors indicate the input value $x\in\RR$. Although 4{,}000 points are shown for illustration, only 30 points are used to train the models.}\label{fg:sliding_circle}
\end{figure}

\paragraph{\bf Dataset}
We construct a dataset ${(x_i,\yy_i)}_{i=1}^{N}$ with only $N=30$ samples. Each $x_i\in\RR$ is drawn uniformly from $[0,1]$. Conditional on $x_i$, we sample $\yy_i\in\RR^2$ uniformly from the unit disk centered at $(2x_i,2x_i)$. A scatter plot of the resulting $\yy_i$ is shown in Figure~\ref{fg:sliding_circle}.

\paragraph{\bf Model}

To tune the hyperparameters of TQF and DRF, we generated 30 independent pairs of a training set (30 samples) and a validation set (1{,}000 samples). We then averaged ES (\eqref{eq:2qpsod} in Appendix~\ref{sc:459ufdkjlsf}) over the 30 runs to assess performance. For TQF, we used 10 trees, fixed $G=20$, $\wt{G}=10$, $T=0$, $K=20$, $M=20$, and $E=10$, and varied \verb|min_samples_leaf| over the grid $\{ 5,10,15,20,25,30 \}$. We found \verb|min_samples_leaf|$=15$ to perform best. For DRF, we used 10 trees and searched over the following hyperparameter grid:
\bi
	\item \verb|honesty|: \{True, False\}
	\item \verb|min_node_size|: $\{1,2,\ldots,15\}$
	\item \verb|sample_fraction|: $\{0.3,0.5,0.7,0.9\}$
\ei
We found DRF to be relatively stable with respect to these hyperparameters, consistent with the numerical results reported in Ref.~\cite{Biewen2025}. The best configuration in our search was \verb|honesty| $=\text{True}$, \verb|min_node_size| $=4$, and \verb|sample_fraction| $=0.7$. For both TQF and DRF, increasing the number of trees did not yield a noticeable improvement in performance.

\paragraph{\bf Evaluation Metric}

We generated 300 independent training datasets and trained TQF and DRF on each of them. For each trained model, we drew $x_{\rm test}\sim \mathrm{Unif}([0,1])$ and produced a predictive distribution $\wt{p}(\yy|x_{\rm test})$. Prediction accuracy was quantified using ED and NLL, evaluated on 50{,}000 and 10{,}000 test samples, respectively, drawn from the true conditional distribution $p_{\rm true}(\yy|x_{\rm test})$. To compute NLL numerically, we estimated the predictive density using \verb|gaussian_kde| in SciPy \cite{scipy} with default parameters. In addition, ES in \eqref{eq:2qpsod} was computed on a test set $\{(x_k,\yy_k)\}_{k=1}^{1000}$.

\begin{table}[tbh]
	\centering
	\caption{Mean and standard deviation of performance metrics for 300 runs with different random seeds for the sliding-disk dataset. The bottom row shows $p$-values of the Wilcoxon signed-rank test. Best values are shown in boldface.}\label{tb:2models_scores}
	\begin{tabular}{lrrr}
		\toprule
		& ED$(\downarrow)$ & ES$(\downarrow)$ & \quad NLL$(\downarrow)$
		\\\midrule
		TQF & {\bf\textsf{0.290(133)}} & {\bf\textsf{0.538(18)}} & 2.01(69)
		\\
		DRF & 0.345(109) & 0.544(21) & {\bf\textsf{1.86(26)}}
		\\
		$p$-value & \quad$1.32 \times 10^{-13}$ & $1.46 \times 10^{-6}$ & 0.238
		\\\bottomrule
	\end{tabular}
\end{table}

\paragraph{\bf Results} 

\begin{figure}[tb]
	\vspace*{3mm}
	\begin{adjustwidth}{0\extralength}{0cm}
	\centering
	\includegraphics[width=.47\textwidth]{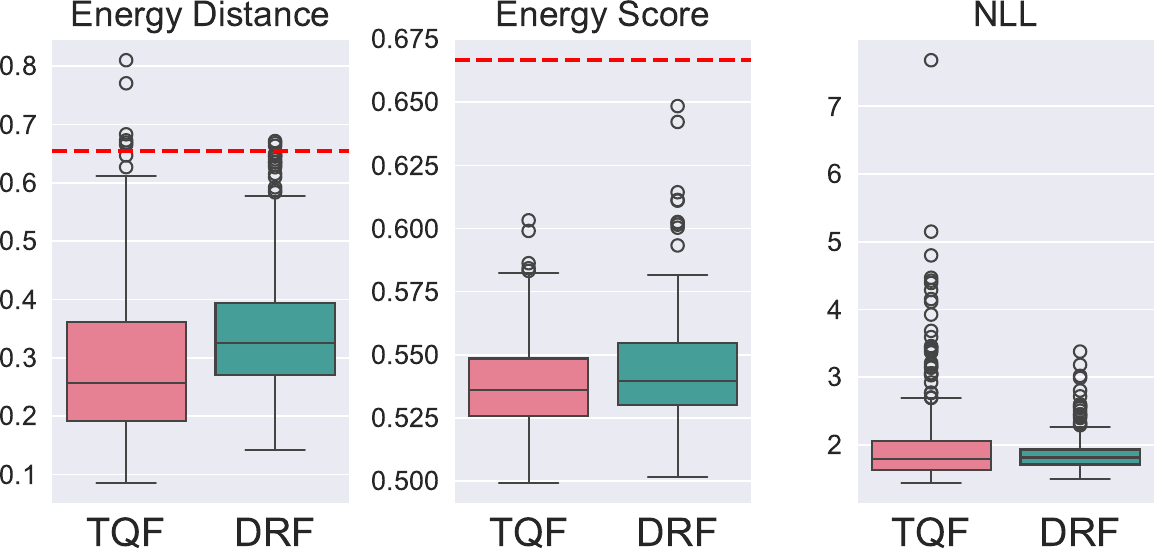}
	\quad 
	\includegraphics[height=.2\textwidth]{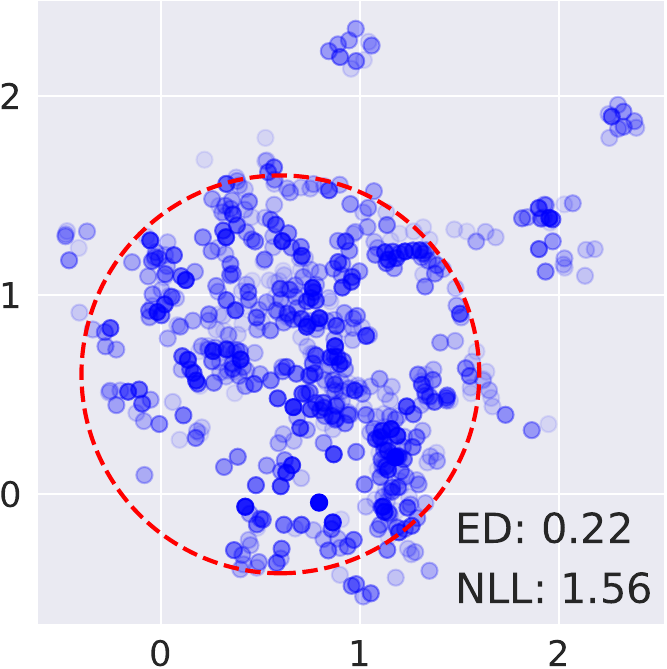}
	\quad 
	\includegraphics[height=.2\textwidth]{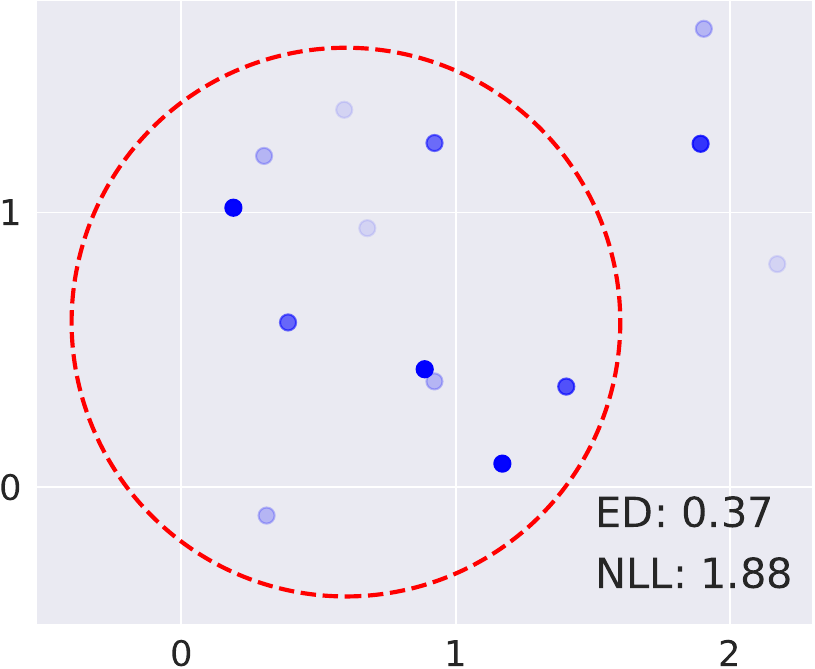}
	\put(-449, 94){\Large (a)}
	\put(-213, 94){\Large (b)}
	\put(-111, 94){\Large (c)}
	\vspace{.5\baselineskip}
	\caption{Numerical results for the sliding-disk data. (a) Boxplots for the scores of TQF and DRF for 300 independent runs. For reference, we show by red dashed lines the scores of a point prediction: $\yy_{\rm pred}|x = (2x,2x)$. (b) Example of a prediction for $p(\yy|x=0.3)$ by TQF, consisting of 705 points. Opacity of points is proportional to their weights. (c) Example of a prediction for $p(\yy|x=0.3)$ by DRF. The two models in (b) and (c) were trained on the exactly same training data. ED and NLL are shown at the bottom right. Red dashed circles indicate the boundary of the true conditional distribution.}\label{fg:compa_simple}
	\end{adjustwidth}
\end{figure}

The results are reported in Table~\ref{tb:2models_scores} and Figure~\ref{fg:compa_simple}. Table~\ref{tb:2models_scores} shows that TQF achieves better ED and ES scores, as supported by the very small $p$-values from the Wilcoxon signed-rank test, which rejects the null hypothesis that the median scores of the two methods are equal. For NLL, DRF appears slightly better than TQF, although the difference is not statistically significant. NLL values should be interpreted with caution because they depend on the choice of KDE settings, such as the bandwidth.

Figure~\ref{fg:compa_simple}(a) visualizes the score distributions across the 300 runs. A number of outliers are present, especially for NLL, consistent with the general observation that NLL penalizes poor predictions more strongly than ES. Panels (b) and (c) compare representative predictive distributions produced by the two models trained on the same dataset. Because DRF can only reweight training samples and cannot generate new support points, its distributional predictions may deteriorate when the training data are scarce. In contrast, TQF produces denser predictions, since the QMEM reconstruction step is designed to generate support points flexibly beyond the observed samples.

\subsection{Evaluation on real-world data}

Next, we train the models on a real-world dataset and examine the characteristics of their distributional predictions.

\paragraph{\bf Dataset}
\begin{figure}[h]
	\centering
	\includegraphics[width=.44\textwidth]{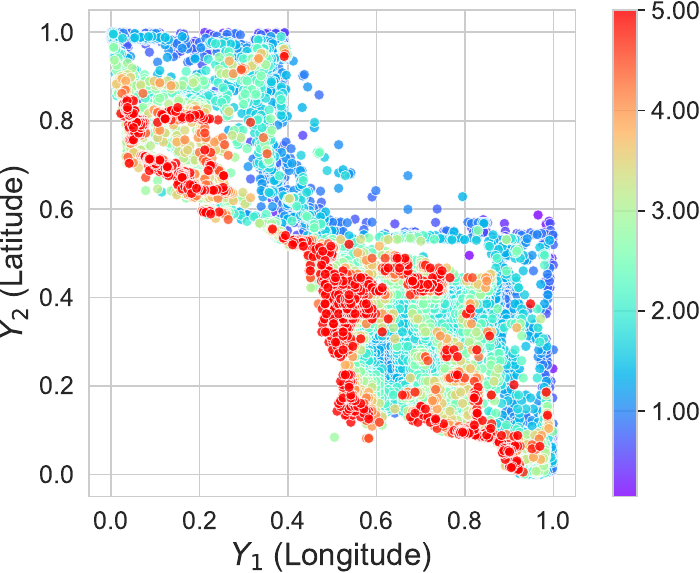}
	\caption{Spatial distribution of records in the California Housing dataset. Colors indicate the median house value (the \texttt{MedHouseVal} column).}\label{fg:housing}
\end{figure}
We use the California Housing dataset \cite{KelleyPace1997}\footnote{\url{https://scikit-learn.org/stable/modules/generated/sklearn.datasets.fetch_california_housing.html}}, which is based on the 1990 California census. It contains 20{,}640 records with eight covariates, and is commonly used for univariate regression with the median house value (\texttt{MedHouseVal}) as the prediction target and the remaining attributes (\texttt{MedInc, HouseAge, AveRooms, AveBedrms, Population, AveOccup, Latitude, Longitude}) as features. In contrast, we use \{\texttt{Latitude}, \texttt{Longitude}\} as the prediction targets and treat the other seven variables (including \texttt{MedHouseVal}) as features. Although \texttt{HouseAge} is capped at 52 years and \texttt{MedHouseVal} at 5.00001 (in units of \$100{,}000), we keep these values unchanged, since any additional treatment of censored observations could introduce bias into the subsequent analysis.

Although inherently spatial, we treat this dataset purely as a generic multi-output tabular regression task to demonstrate TQF's general applicability, rather than competing with specialized spatial statistics methods.

The dataset is spatially imbalanced, with records concentrated around major metropolitan areas such as Los Angeles and the San Francisco Bay Area, which can distort the learning problem. To mitigate this effect, we apply a nonlinear quantile transformation (scikit-learn's \texttt{QuantileTransformer}) to \texttt{Longitude} and \texttt{Latitude} so that their \emph{marginal} distributions become approximately uniform on $[0,1]$. The transformed distribution is shown in Figure~\ref{fg:housing}. Overall, higher house values tend to be observed closer to the coast.

\paragraph{\bf Models}

We compare the following models: ``Simple'', RF$^*$, LightGBM$^*$, KNN, GP, NGBoost, and TQF. Details are given below. Models marked with $^*$ output point predictions.
\bi
	\item ``Simple'': A trivial baseline that returns the entire training set as the predictive distribution for every test input.
	\item RF$^*$: A random-forest regressor with 200 trees. We tune \verb|min_samples_leaf| by grid search over $\{1,5,10,20,50,100,200,300\}$ using the $\ell_2$ loss, and select \verb|min_samples_leaf|\,$=10$.
	\item LightGBM$^*$: A gradient-boosted decision-tree model \cite{Ke2017}. We tune the hyperparameters over the grids below, using the $\ell_2$ loss.
	\begin{itemize}
		\item \verb|learning_rate|: $\{0.005,0.01,0.02,0.03,0.05,0.1\}$ \quad $\to$ select $0.05$
		\item \verb|min_data_in_leaf|: $\{1,5,10,20,50,100,200,300\}$ \quad $\to$ select $50$
	\end{itemize}
	The number of boosting rounds is chosen automatically via early stopping. Since LightGBM does not natively support multivariate targets, we train two independent models, one per target component.
	\item KNN: As we will see below, RF feature importances suggest that \texttt{MedHouseVal} and \texttt{AveOccup} are the two most informative features. We standardize these two variables and train a KNN regressor using them. For a test input, we return the responses of the $k$ nearest training points with uniform weights as the predictive distribution. Grid search selects $k=100$.
	\item GP: A Gaussian-process-based baseline. Because exact GP inference is prohibitively expensive for $N>10^4$, we use a scalable approximation: we generate 1{,}000 random Fourier features using scikit-learn's \texttt{RBFSampler} and fit a Bayesian linear model in the resulting feature space. To facilitate training, we first apply \texttt{QuantileTransformer} so that each feature is approximately standard normal. To obtain bivariate predictions, we fit two independent models (one per target component), each outputting a predictive mean and standard deviation. This independent modeling ignores cross-correlation between the two target components.

	\item NGBoost: Natural Gradient Boosting \cite{ngboost2020,OMally2021}.\footnote{\url{https://stanfordmlgroup.github.io/projects/ngboost/}}
	It models predictive uncertainty by fitting a parametric distribution and optimizing its parameters using natural gradients. In our experiments, we use a bivariate normal distribution. We tune the hyperparameters over the grids below, using the $\ell_2$ loss. We do not constrain \verb|max_depth|.
	\begin{itemize}
		\item \verb|learning_rate|: $\{0.001,0.005,0.01,0.02,0.05,0.1,0.2,0.3,0.4,0.5\}$ \quad $\to$ select $0.1$
		\item \verb|min_samples_leaf|: $\{5,10,20,50,100,200,300\}$ \quad $\to$ select $100$
	\end{itemize}
	The number of boosting rounds is determined via early stopping.
	\item TQF: We use 50 trees. We tune the hyperparameters over the grids below.
	\begin{itemize}
		\item $G$: $\{5,10\}$ \quad $\to$ select $10$
		\item $\wt{G}$: $\{1,5,10\}$ \quad $\to$ select $1$
		\item $T$: $\{0,3,6\}$ \quad $\to$ select $3$
		\item \verb|min_samples_leaf|: $\{1,2,3,4,5,10,25,50,100,200,300,500\}$ \quad $\to$ select $5$
	\end{itemize}
\ei

\paragraph{\bf Evaluation Metric} 

We use $\mathrm{R}^2$ and ES to quantify mean-prediction accuracy and distributional-prediction accuracy, respectively. We perform 10-fold cross-validation (CV) and report average scores. During hyperparameter tuning, each model is trained on $8/9$ of the training data, with the remaining $1/9$ used as a validation set.

\paragraph{\bf Results}

\begin{figure}[htb]
	\centering
	\includegraphics[width=.45\textwidth]{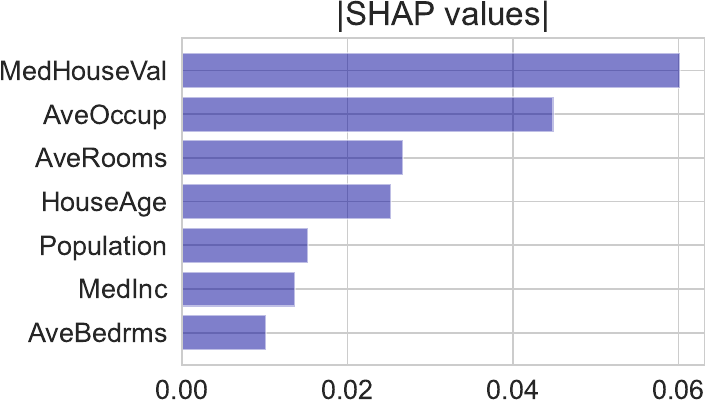}
	\caption{Feature importances (global absolute SHAP values) computed with RF.}\label{fg:housing_rf_shap}
\end{figure}

Table~\ref{tb:housing_scores} summarizes the results. (For RF and LightGBM, ES is computed using the fact that, for point predictions, ES reduces to the mean absolute error.) Overall, this regression task is challenging: no method attains $\mathrm{R}^2$ above 0.3. The best-performing models in terms of $\mathrm{R}^2$ are RF and TQF, and their difference is not statistically significant ($p=0.22$; paired $t$-test). We also compute SHAP\footnote{\url{https://shap.readthedocs.io/en/latest/index.html}} values \cite{Lundberg2017} for RF (Figure~\ref{fg:housing_rf_shap}). The most important feature is \texttt{MedHouseVal}, which is consistent with the spatial pattern in Figure~\ref{fg:housing}, followed by \texttt{AveOccup} (average occupancy per household).

\begin{table}[t]
	\begin{adjustwidth}{0\extralength}{0cm}
	\caption{Prediction performance on the California housing dataset with bivariate targets \texttt{(Longitude, Latitude)}. Parentheses report one standard deviation across the 10 CV folds for $\mathrm{R}^2$, and across all 20{,}640 test predictions aggregated over 10-fold CV for ES. Best scores are shown in boldface.}
	\label{tb:housing_scores}
	\centering
	\begin{tabular}{lrrrrrrr}\toprule[0.6mm]
		& Simple & RF$^*$ & LightGBM$^*$ & KNN & GP & NGBoost & TQF
		\\\midrule
		$\mathrm{R}^2$ $(\uparrow)$ & $0.0(0)$ & {\bf\textsf{0.202(7)}} & $0.196(9)$ & $0.094(9)$ & $0.191(10)$ & $0.191(7)$ & {\bf\textsf{0.204(8)}}
		\\
		$\ES$ $(\downarrow)$ & $0.244(90)$ & $0.311(190)$ & $0.314(189)$ & $0.221(117)$ & $0.229(122)$ & {\bf\textsf{0.215(122)}} & {\bf\textsf{0.216(128)}}
		\\\bottomrule[0.6mm]
	\end{tabular}
	\end{adjustwidth}
\end{table}

For ES, the trivial baseline (``Simple'') attains 0.244, which serves as a natural reference level: any informative distributional predictor should achieve ES below 0.244. The best-performing models are NGBoost and TQF, with ES around 0.215--0.216, and their difference is not statistically significant ($p=0.44$; Wilcoxon signed-rank test).

\begin{figure}[bth]
\begin{adjustwidth}{-0\extralength}{0cm}
	\centering
	\includegraphics[width=.24\textwidth]{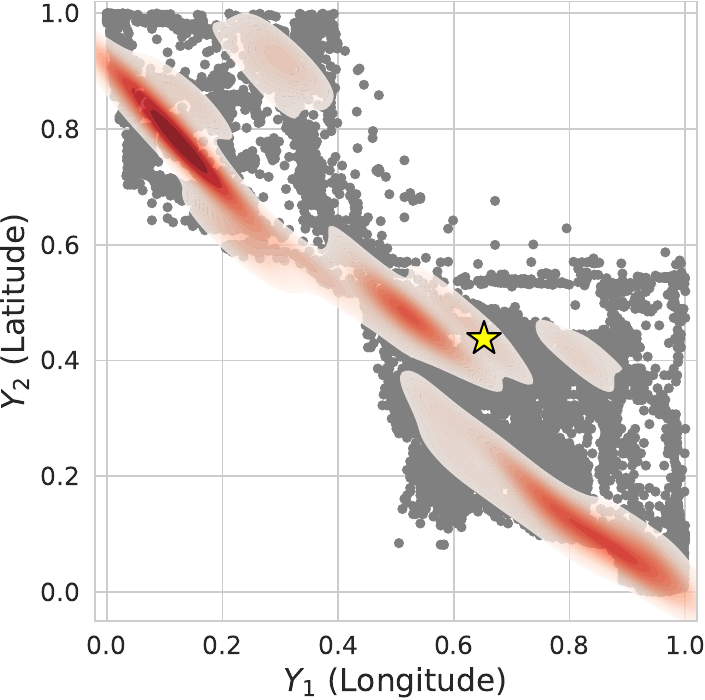}
	\includegraphics[width=.24\textwidth]{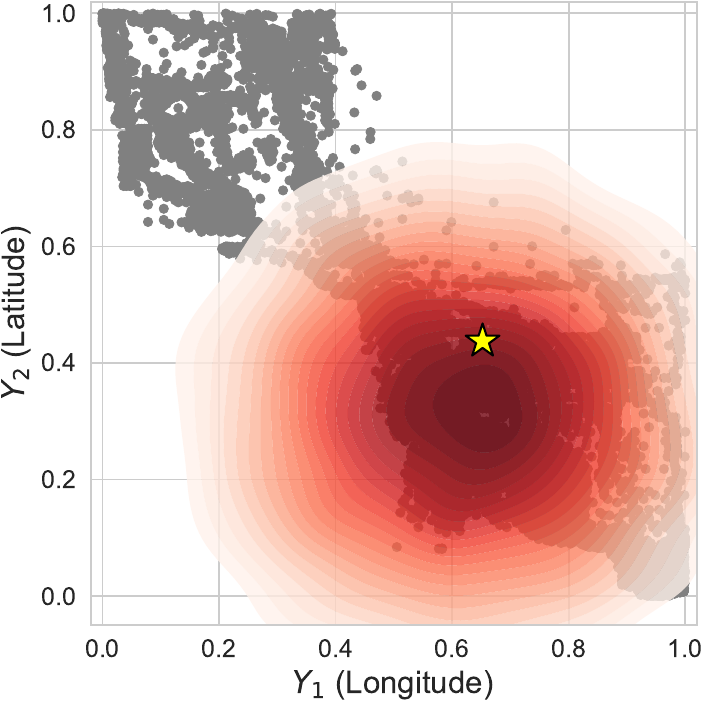}
	\includegraphics[width=.24\textwidth]{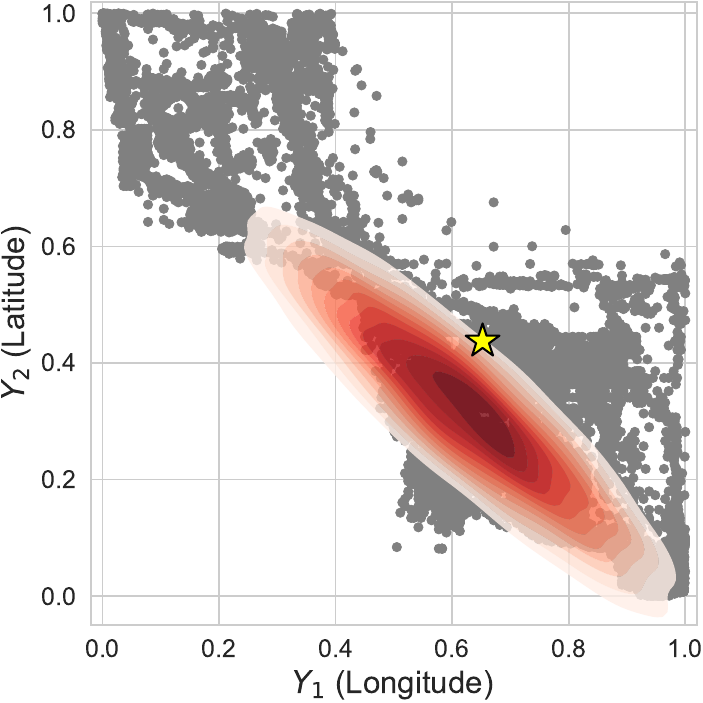}
	\includegraphics[width=.24\textwidth]{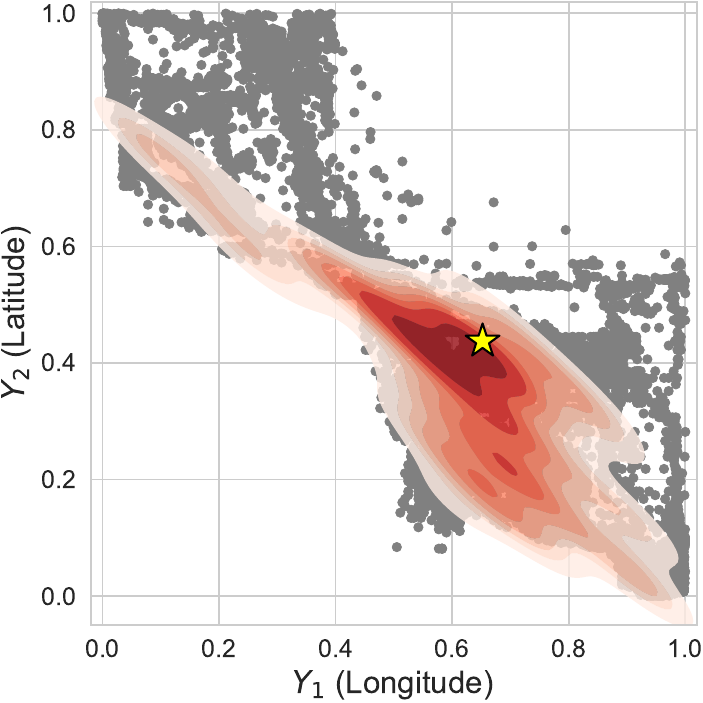}
	\put(-384, 109){\large \sf KNN}
	\put(-270, 109){\large \sf GP}
	\put(-177, 109){\large \sf NGBoost}
	\put(-60, 109){\large \sf TQF}
	\caption{Predictive distributions for the same test sample, provided by KNN, GP, NGBoost, and TQF from left to right, respectively. All models were trained on the same training data. A small yellow star in each panel is the ground truth.}\label{fg:housing_dist_images}
\end{adjustwidth}
\end{figure}

Figure~\ref{fg:housing_dist_images} illustrates how distributional predictions differ across the models. NGBoost, which fits a bivariate Gaussian, produces a narrow, tilted ellipse, highlighting the importance of \emph{joint} modeling of \texttt{Longitude} and \texttt{Latitude}. By contrast, the GP baseline fits the two components independently and therefore yields a broader, axis-aligned predictive distribution that ignores cross-correlation between the targets. For this particular test input, TQF produces a distribution with a shape qualitatively similar to that of NGBoost. The fact that the two methods achieve comparable ES on average, despite the nonparametric nature of TQF, may suggest that a unimodal Gaussian approximation is already adequate for capturing uncertainty in this dataset. Nevertheless, we emphasize that TQF attains the highest $\mathrm{R}^2$ among the tested models, indicating that its uncertainty estimation does not come at the expense of point-prediction accuracy.

\section{Conclusions}\label{sc:concl}

In this paper, we presented Tomographic Quantile Forests (TQF), a new random-forest-based regression model for multivariate targets that provides nonparametric predictive uncertainty estimates. To the best of our knowledge, this is the first work to implement CT-principle-based distributional prediction using tree-based models. Whereas conventional directional-quantile methods typically require training a separate model for each direction in the target space, TQF covers all directions within a single forest model. We hope that this work complements the rapidly growing literature on deep-learning-based uncertainty-aware modeling and encourages further research on more sophisticated non-neural methods tailored to tabular data.

It should be emphasized that TQF is positioned as a multivariate extension of tree-based regressors (e.g., QRF) specifically tailored for tabular data, rather than a full generative model.

TQF involves several hyperparameters, and tuning them can be time-consuming in practical applications; thus, making the method more robust to hyperparameter choices is an important direction for future research. Furthermore, while we relied heavily on synthetic data in this study to rigorously evaluate the recovery of ground-truth distributions, extensive benchmarking on diverse real-world datasets is a necessary next step. In this context, we plan to explore applications to complex, large-scale spatiotemporal data, such as those arising in weather forecasting, where comparisons with domain-specific spatial methods will be highly relevant.

\appendixtitles{yes} 
\appendixstart
\appendix
\numberwithin{equation}{section} 

\section{Technical Background}\label{sc:prel}

\subsection{Discrepancy measures for distributions}\label{sc:disc}

Below we provide definitions and formulas for several discrepancy metrics between probability distributions. For further details, we refer the reader to Refs.~\cite{CIT-004,COTFNT,Bischoff2024review}. Throughout this section, $\mu$ and $\nu$ denote probability measures on $\RR^d$.

\paragraph{\bf Wasserstein Distance} 

The \emph{$p$-Wasserstein distance} \cite{Villani2003book,Villani2009book,COTFNT,Montesuma2025,Peyre2025} is defined as \al{
	\W_p(\mu,\nu) \equiv \mkakko{
		\underset{\pi\in\Pi(\mu,\nu)}{\inf}\int \rmd \pi(\xx, \yy)\;
		\| \xx - \yy \|_2^p 
	}^{1/p},
}
where $\Pi(\mu,\nu)$ is the set of all couplings between $\mu$ and $\nu$ and $p\geq 1$. $\W_p$ satisfies all the axioms of distance and $\W_p(\mu,\nu)\geq 0$ with the equality if and only if $\mu=\nu$. The case $p=1$ is especially well known as the earth mover's distance. If $d=1$ (namely, if $\mu$ and $\nu$ are measures on $\RR$), there is a convenient representation
\al{
	\W_p(\mu,\nu) = \mkakko{\int_{0}^{1}\rmd q ~\left|
	F^{-1}_\mu(q) - F^{-1}_\nu(q)
	\right|^{p}}^{1/p},
	\label{eq:Was}
}
where $F^{-1}_{\mu,\nu}$ denote the inverse CDF of $\mu$ and $\nu$.

\paragraph{\bf Sliced Wasserstein Distance}

Computation of $\W_p$ is expensive when $d\geq 2$. To bypass this problem, one can use the \emph{sliced $p$-Wasserstein distance} \cite{Rabin2011,Bonneel2015} defined as
\al{
	{\SW}_p(\mu,\nu)\equiv \mkakko{\int_{\SSS^{d-1}}\!\!\!\rmd \nn\;
	\W_p \big(
		\wh{R}_{\nn\sharp}\mu, \wh{R}_{\nn\sharp}\nu \big)^p
	}^{1/p},
	\label{eq:SW}
}
where $\SSS^{d-1}:=\{\xx\in \RR^d \mid \| \xx \|_2 = 1 \}$ is a unit hypersphere, $\int \rmd \nn$ is a uniform measure on $\SSS^{d-1}$ normalized to $1$, and $\wh{R}_{\nn\sharp}$ is a pushforward operator induced by projection with $\nn\in\SSS^{d-1}$. Since $\W_p\big(\wh{R}_{\nn\sharp}\mu, \wh{R}_{\nn\sharp}\nu \big)$ is a Wasserstein distance in one dimension, it can be computed fast. The integral over $\SSS^{d-1}$ is typically replaced with a Monte Carlo average with finite samples $\{\nn_k\}$.

\paragraph{\bf Energy Distance}

Energy distance (ED) measures the discrepancy of two distributions using a Euclidean norm \cite{Szabo2002,Szabo2003,Baringhaus2004,Szekely2004,Klebanov2005,
Baringhaus2010,Szekely2013,Szekely2016}. ED is defined by
\al{
	\ED(\mu, \nu) & = \Big(
		2 \, \EE_{\xx \sim \mu, \yy\sim\nu}
		\| \xx - \yy \|_2 
		- \EE_{\xx, \xx'\sim\mu}
		\| \xx - \xx' \|_2 
		- \EE_{\yy, \yy'\sim \nu}
		\| \yy - \yy' \|_2 
	\Big)^{1/2} ,
	\label{eq:ed}
}
which vanishes if and only if $\mu=\nu$. ED offers a more robust alternative to divergence measures such as Kullback–Leibler and total variation for finite-sample analysis. Unlike these density-dependent metrics, ED bypasses the need for density estimation, allowing for direct computation from empirical data.

\paragraph{\bf Convexity}

All the metrics above share the convexity property. Collectively denoting $\W_p(\mu,\nu)^p$, $\SW_p(\mu,\nu)^p$ and $\ED(\mu,\nu)$ by ${\cal D}(\mu,\nu)$, one has the inequality
\al{
	{\cal D}((1-\lambda)\mu_1+\lambda \mu_2,\nu)\leq 
	(1-\lambda){\cal D}(\mu_1,\nu)+\lambda {\cal D}(\mu_2,\nu)
	\label{eq:64trfg}
}
for $0\leq \lambda \leq 1$ and arbitrary probability measures $\mu_1,\mu_2$ and $\nu$.

\subsection{Radon transform}

Given a unit vector $\nn\in \SSS^{d-1}$, any $\xx\in\RR^d$ admits an orthogonal decomposition $\xx=s\nn+\xx_\perp$ where $s\in\RR$ and $\nn^\top\xx_{\perp}=0$. The Radon transform \cite{Deans1983book,Helgason1999book} of an integrable function $f(\xx)$ over $\RR^d$ is defined by
\al{
	\mathcal{R}f(s,\nn) \equiv 
	\int_{\RR^d}\rmd \xx\;\delta(s-\nn^\top \xx)f(\xx)
	= \int_{\RR^{d-1}}\!\!\!\rmd\xx_\perp\; f(s\nn+\xx_\perp)\,.
}
If the data of $\mathcal{R}f$ for all $\nn$ are available, the original function $f$ can be recovered through the inverse Radon transform. This is the theoretical basis of modern computed tomography \cite{Hsieh2015book,Willemink2019}. In practice, however, only noisy data for a finite number of projections can be acquired, which leads to an ill-posed problem. Various numerical methods have been developed to address this. Among the most classical and widely used is the Filtered Back-Projection (FBP) \cite{Bracewell1967,Ramachandran1971,Logan1975}. FBP is numerically efficient, but when the number of projections is insufficient, FBP-reconstructed images are of poor quality \cite{Boas2012,Hsieh2015book,Purisha2019}.

\subsection{Proper scoring rules}\label{sc:459ufdkjlsf}

In general, the true distribution $p(\yy)$ is unknown and we only observe samples drawn from it. Evaluating the quality of a probabilistic prediction $\wt{p}(\yy)$ from an observation $\yy \sim p(\yy)$ is therefore nontrivial. As argued in Ref.~\cite{Gneiting2007}, a scoring rule (loss function) for probabilistic forecasts should be \emph{strictly proper}: the expected score is minimized if and only if the predicted distribution $\wt{p}$ coincides with the data-generating distribution $p$. Pedagogical reviews on proper scoring rules can be found in Refs.~\cite{Gneiting2014,Bjerregaard2021,Alexander2022,Pic2025,Waghmare2025}. The negative log-likelihood (NLL), $-\log \wt{p}(\yy)$, is a canonical example of a strictly proper scoring rule. Another widely used example in one dimension is the \emph{continuous ranked probability score} (CRPS) \cite{Matheson1976}, defined as%
\footnote{Sometimes the negative of this definition is used for CRPS in the literature. In our convention, \emph{lower} CRPS indicates a \emph{better} prediction.}
\al{
	\CRPS(y,\wt{p}) = \EE_{\wt{y}\sim \wt{p}} |y-\wt{y}|	
	- \frac{1}{2}\,\EE_{\wt{y},\wt{y}' \sim \wt{p}} |\wt{y}-\wt{y}'|
}
where $y,\wt{y}$ and $\wt{y}'\in\RR$. CRPS has been widely used in meteorology. A multivariate generalization of CRPS is the \emph{energy score} (ES) \cite{Gneiting2007}, defined in $\RR^d (d>1)$ as
\al{
	\ES(\yy, \wt{p}) = \EE_{\wt{\yy}\sim \wt{p}}\|\yy-\wt{\yy}\|
	- \frac{1}{2}\,\EE_{\wt{\yy},\wt{\yy}'\sim \wt{p}} \|\wt{\yy}-\wt{\yy}'\|\,.
}
There is an intriguing link between CRPS and ES. If we project both the sample and the distribution onto a one-dimensional axis, compute their CRPS, and average the result over all directions, then the outcome coincides with ES up to a multiplicative constant \cite[Theorem~4.1]{Korotin2021}. Namely,
\al{
	\ES(\yy, \wt{p}) & = \frac{d-1}{2}\frac{\text{Vol}(\SSS^{d-1})}{\text{Vol}(\SSS^{d-2})}
	\int_{\SSS^{d-1}} \!\!\! \rmd \nn~
	\CRPS \big( \nn^\top \yy, \wh{R}_{\nn\sharp}\wt{p} \big)
}
where $\int \rmd\nn$ is the uniform measure on $\SSS^{d-1}$ normalized to unity, and $\wh{R}_{\nn\sharp}$ is the pushforward operator from Appendix~\ref{sc:disc}. Now we substitute $\text{Vol}(\SSS^{d-1})=2\pi^{d/2}/\Gamma(d/2)$ and make use of the fact that CRPS is twice the average of the mean pinball loss, to obtain
\al{
	\ES(\yy, \wt{p}) & = (d-1)\sqrt{\pi}
	\frac{\Gamma(\frac{d-1}{2})}{\Gamma(\frac{d}{2})}
	\int_{\SSS^{d-1}} \!\!\! \rmd \nn~
	\int_0^1\rmd q~\mkakko{ \mathbbm{1}_{\nn^\top \yy \leq F^{-1}_{\wh{R}_{\nn\sharp}\wt{p}}(q)} - q }\mkakko{ F^{-1}_{\wh{R}_{\nn\sharp}\wt{p}}(q)-\nn^\top \yy }
	\label{eq:2qpsod}
}
where $F^{-1}_{\wh{R}_{\nn\sharp}\wt{p}}$ is the inverse CDF of $\wh{R}_{\nn\sharp}\wt{p}$. These integrals can be approximated by a discrete sum efficiently.

\addtocontents{toc}{\protect\setcounter{tocdepth}{0}}
\reftitle{References}
\bibliography{mdpi_v3.bbl}

\funding{This research received no external funding.}

\dataavailability{The original data presented in the study are openly available at \url{https://github.com/TaTKSM/TQF}.} 

\conflictsofinterest{The author declares no conflicts of interest.} 
\end{document}